\documentclass[10pt,twoside,openright,english,italian]{book}

\usepackage[twoside=true]{geometry}

\usepackage{amsmath,amsfonts,amssymb,epsfig,verbatim,color,amsthm}
\usepackage{graphicx}
\usepackage{multirow}
\usepackage{rotating}
\usepackage{booktabs}
\usepackage{pdfpages}
\usepackage{subfig}
\usepackage{algorithm}
\usepackage{algpseudocode}
\usepackage{url}
\usepackage{bm}
\usepackage{array}
\usepackage{xcolor,colortbl}
\usepackage{textcomp}
\usepackage{multirow}

\newcommand{\mat}[1]{\mathtt #1}

\newcommand{\ind}[1]{\mathbb 1_{#1}}
\newcommand{\vct}[1]{\mathbf #1}

\newcommand{\argmin}{\operatornamewithlimits{\arg\,\min}}
\newcommand{\x}{{\bf x}}

\newcommand{\et}{\textit{et al. }}

\newtheorem{proposition}{Proposition}
\newtheorem{theorem}{Theorem}

\usepackage{url}
\usepackage{lineno}
\usepackage{times}
\usepackage{textcomp}

\usepackage{phdthesis}

\usepackage{fancyhdr}
\usepackage{color}
\usepackage{array}
\usepackage{mdwmath}
\usepackage{mdwtab}
\usepackage{amsmath,amssymb}
\usepackage{cite}
\usepackage{graphicx}
\usepackage{glossaries}
\usepackage{listings}
\usepackage{subfig}
\usepackage{booktabs}
\usepackage{latexsym}
\usepackage{color}
\usepackage{url}
\usepackage{bnf}
\usepackage{rotating}
\usepackage{multirow}
\usepackage{phdtitle}
\usepackage{paralist}
\usepackage{bibentry}
\usepackage[bookmarks=true,pdftex=false,bookmarksopen=true,pdfborder={0,0,0}]{hyperref}
\usepackage{lscape}

\usepackage{algpseudocode}

\usepackage{algorithm}
\usepackage{longtable}
\usepackage[T1]{fontenc} 
\usepackage[latin1]{inputenc}

\newcommand{\hide}[1]{}

\DeclareMathAlphabet{\mathpzc}{OT1}{pzc}{m}{it}

\usepackage[greek,english,italian]{babel}

\usepackage{ulem}
\hypersetup{pdftitle={Title}, pdfauthor={Author}}

\nobibliography*

\department {Dipartimento di Scienze Ambientali, Informatica e Statistica}
\phdprogram{Dottorato di Ricerca in Informatica, XXXII ciclo}

\author{Alemu Leulseged Tesfaye}
\title{Constrained Dominant sets and Its applications in computer vision}

\supervisor{Marcello Pelillo}

\chair{Ricardo Focardi}
\titleimage{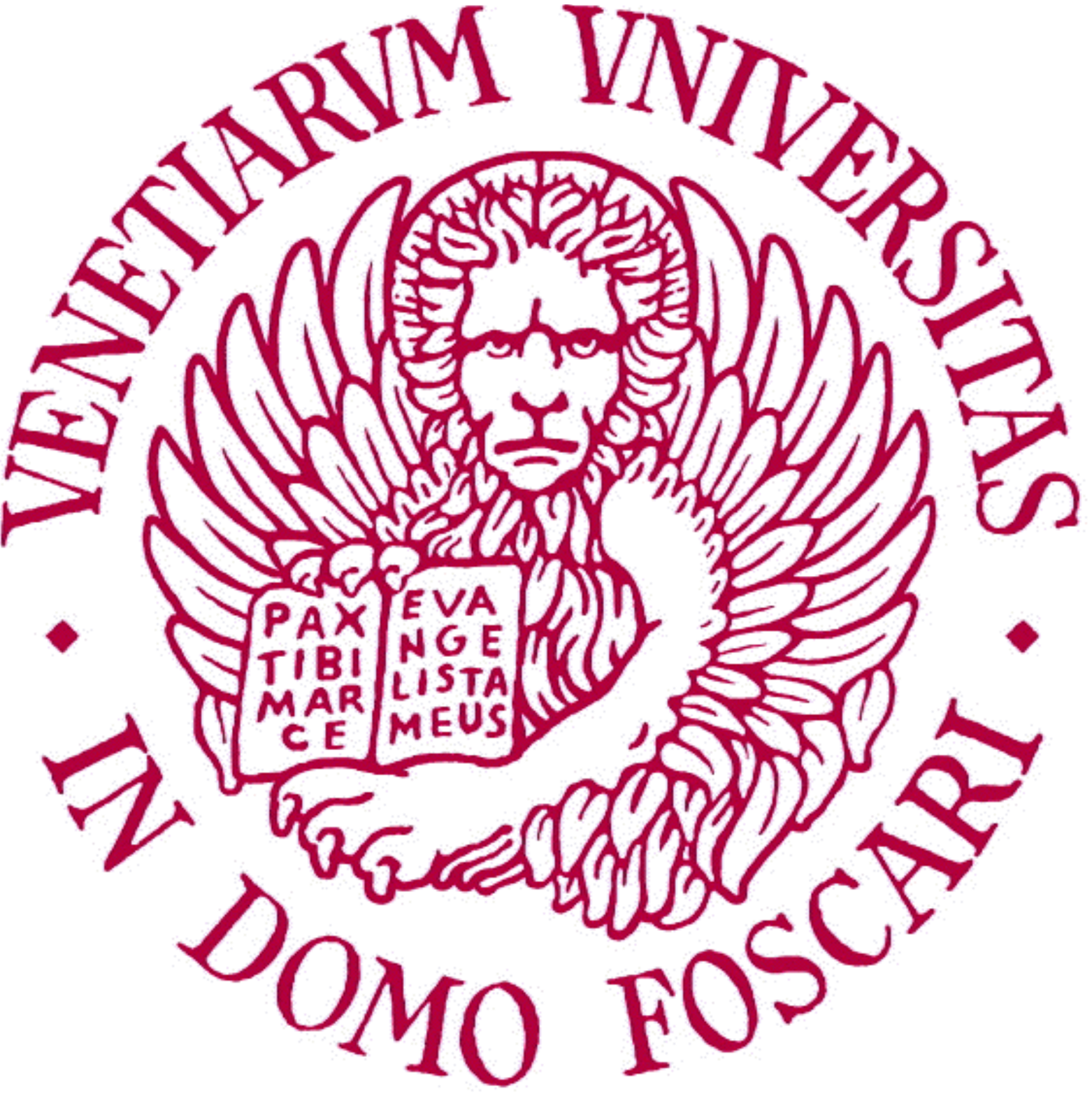}
\phdcycle{}

\begin{document}
\selectlanguage{english}

\maketitle

\pagestyle{empty}

\cleardoublepage


\cleardoublepage

\pagestyle{fancy}
\setcounter{page}{1}
\pagenumbering{Roman}

\chapter*{Abstract}
\lettrine{I}n this thesis, we present new schemes which leverage a constrained clustering method to solve several computer vision tasks ranging from image retrieval, image segmentation and co-segmentation, to person re-identification. In the last decades clustering methods have played a vital role in computer vision applications; herein, we focus on the extension, reformulation, and integration of a well-known graph and game theoretic clustering method known as Dominant Sets. Thus, we have demonstrated the validity of the proposed methods with extensive experiments which are conducted on several benchmark datasets.

We first discuss `Dominant Sets for "Constrained" Image Segmentation,' DSCIS. In DSCIS, we present a unified model to tackle image segmentation and co-segmentation problem in both an interactive and unsupervised fashion; whereby,  the proposed algorithm can deal naturally with several types of constraints and input modality, including scribbles, sloppy contours, and bounding boxes, and is able to robustly handle noisy annotations on the part of the user. Our method is based on some properties of a family of quadratic optimization problems related to dominant sets, a well-known graph-theoretic notion of a cluster which generalizes the concept of a maximal clique to edge-weighted graphs. In particular, we show that by properly controlling a regularization parameter which determines the structure and the scale of the underlying problem, we are in a position to extract groups of dominant-set clusters that are constrained to contain predefined elements. 

Following, we present novel schemes for content-based image retrieval (CBIR)  using constrained dominant sets (CDS). We present two different CBIR methods. The first method, `Multi-feature Fusion for Image Retrieval Using Constrained Dominant Sets,' MfFIR, fuse several hand-crafted and deep features to endow a representative similarity which better define the closeness of given query and gallery images; whereas, the second one, `Constrained Dominant Sets for Image Retrieval,' CDSIR, exploit a constrained diffusion process to produce a robust similarity between query and gallery images.  In MfFIR, we propose a computationally efficient approach to fuse several hand-crafted and deep features, based on the probabilistic distribution of a given membership score of a constrained cluster in an unsupervised manner. Towards this end, we first introduce an incremental nearest neighbor (NN) selection method, whereby we dynamically select k-NN to the query. Next, we build several graphs from the obtained NN sets and employ constrained dominant sets (CDS) on each graph G to assign edge weights which consider the intrinsic manifold structure of the graph, and detect false matches to the query. Finally, we compute the positive-impact weight (PIW) based on the dispersive degree of the characteristics vector. As a result, we exploit the entropy of a cluster membership-score distribution. In addition, the final NN set bypasses a heuristic voting scheme. Our approach presents two main advantages. Firstly, compared to the state of the art methods, it can robustly quantify the effectiveness of features for a specific query, without any supervision. Secondly, by diffusing the pairwise similarity between the nearest neighbors, our model can easily avoid the inclusion of false-positive matches in the final shortlist. On the other hand, in CDSIR, we leverage constrained dominant sets to dynamically constrain a similarity diffusion process to provide context-sensitive similarities.

Finally, we present a Deep Constrained Dominant Sets (DCDS); in which, we are able to optimize the constrained-clustering process in an end-to-end manner and leverage contextual information in the learning process. In this work, we integrate the well-known graph and game-theoretic method called CDS into a deep model and tackle the challenging computer vision problem of person re-identification. Furthermore, we reformulate the problem of person re-identification as a constrained-clustering problem and build a model that is end-to-end trainable.

\chapter *{Acknowledgments}

I first thank the almighty God for everything. I then must express my deepest gratitude to my supervisor professor Marcello Pelillo, for his scientific guidance, encouragement and fruitful discussion during my Ph.D. journey.

I also would love to thank my co-authors: Eyasu Zemene (Josh) for being not only a co-author but also a good friend and brother; and Dr. Mubarak Shah for granting me a visiting scholar position at CRCV, University of Central Florida, whereby I solidify my computer vision knowledge further.

And, special thanks go to Yoni, Teddy, Sure, Seyum, Tinsu, Ili, Felix, Carmelo, and Ayoub; please accept my sincere gratitude for being a good friend; moreover, for the unforgettable moments, we had in Campo and Baum (coffee area).  
I also would like to thank the Department of Computer Science of Ca' Foscari University of Venice for financing my Ph.D. study; I thank Prof. Ricardo Focardi and Nicola Miotello for co-ordinating the Ph.D. program and for the kind assistance they have provided me without hesitation.  

I thank my external reviewers: Prof. Lamberto Ballan, and Prof. Xu Yongchao for the time they spent on carefully reading the thesis and
for their insightful comments and suggestions.

Last but not least, I thank my family for the unconditional love and support, especially my mother (etetye) and aunt (eneye).  
\selectlanguage{english}
\chapter*{Preface}

\lettrine{T}his dissertation is submitted in fulfillment of the requirements for the degree of doctor of philosophy at Ca' Foscari University of Venice. The thesis presents novel methods which inherit several techniques from game theory, graph theory, and deep learning. The first chapter introduces the graph and game theoretic clustering methods called Dominant Sets and its constrained variant Constrained Dominant Sets. The second chapter presents the application of constrained dominant sets to tackle the problem of interactive image segmentation and co-segmentation (in both supervised and unsupervised fashion) \cite{ZemAP19}; it has been appeared in Transactions on Pattern Analysis and Machine Intelligence (TPAMI). The third chapter discusses two distinct methods which attack the same problem of content-based image retrieval. The first method, called Constrained Dominant Sets for Image Retrieval \cite{ZemAlePalICPR2016}, has been presented in International Conference on Pattern Recognition (ICPR); whereas, the second one called Multi-feature Fusion using Constrained Dominant Sets for Image retrieval \cite{AlemuPelillo} has been appeared in a journal known as Image and Vision Computing (IVC). Finally, the last chapter presents a very interesting work which integrates constrained dominant sets in a deep neural network model. It is a collaboration work with Dr. Mubarak Shah; which has been done while I was a visiting scholar at the Center for Research in Computer Vision (CRCV). This work \cite{AlemuPel2019} has been published in International Conference on Computer Vision (ICCV). 

\chapter*{Introduction}

Manually labeling a large amount of data, that arise due to a deluge of information from several automations and data collections, is costly. Due to this reason, unsupervised clustering has attracted a considerable attention of pattern recognition community. Clustering deals with defining classes from the data without a prior knowledge of the class labels. Cluster analysis has been applied to solve several real world problems such as anomaly detection,  image segmentation, natural language processing, document grouping and recommendation systems.
	
Clustering methods can be roughly divided into two main categories such as partitioning and hierarchical algorithm.  Partitioning based clustering methods split the dataset into k <or = n groups, where n is the number of objects in the dataset, whereas hierarchical algorithms gradually form clusters through either agglomerations or divisions. Furthermore, there has been a resurgence of interest around graph based (pairwise) methods \cite{GdalyahuWW01,Aks,ShiM00}, that cast the data to be clustered (pixels, super-pixel, edge elements, etc) as a vertices of a similairity (edge-weighted) graph, where the edges reflect neighborhood relations, and the weights denote the similarity between data. Indeed, it is natural to map the data to be clustered to the nodes of a weighted graph (the so-called similarity graph), with edge weights representing similarity relations. In cluster analysis, graph clustering is defined as a process of searching for groups of related vertices in a graph. Graph-theoretic clustering methods are of significant interest since they cast clustering as pure graph-theoretic problems for which a solid theory and powerful algorithms have been developed. As can be observed from \cite{DudaHS01}, these methods can produce highly intricate clusters, but they rarely optimize an easily specified global cost function. Graph-theoretic clustering methods basically comprises searching for certain combinatorial structures in the similarity graph, such as a minimum cut \cite{GdalyahuWW01,ShiM00,WuL93} or a minimum spanning tree \cite{Zahn71}.
	
In this thesis, we present different approaches that leverage techniques which are based on some properties of a family of quadratic optimization problems related to dominant sets, a well-known graph-theoretic notion of a cluster which generalizes the concept of a maximal clique to edge-weighted graphs.
Moreover, we exploit the constrained version of dominant sets clustering known as constrained dominant sets clustering \cite{ZemeneP16} (CDS); which is based on some properties of a family of quadratic optimization problems related to dominant sets. In particular, by properly controlling a regularization parameter which determines the structure and the scale of the underlying problem, one can extract a dominant set cluster which is constrained to contain user-provided constraints.
Thus, we able to tackle several computer vision problems such as image segmentation and co-segmentation, image retrieval, and person re-identification problems.  
		
We first discuss our novel and multi-modal scheme, which is formulated in such a way that tackles the problem of image segmentation and co-segmentation in both unsupervised and interactive manner.
Image segmentation is arguably one of the oldest and best-studied problems in computer vision, being a fundamental step in a 
variety of real-world applications, and yet remains a challenging task \cite{RichardSzeliski11} \cite{ForsPonse11}.
Besides the standard, purely bottom-up formulation, which involves partitioning an input image into coherent regions, in the past few years several variants have been proposed which are attracting increasing attention within the community.
Most of them usually take the form of a ``constrained'' version of the original problem, whereby the segmentation process is guided by some external source of information.
For example, user-assisted (or ``interactive'') segmentation has become quite popular nowadays, especially because of its potential applications in problems such as image and video editing, medical image analysis, etc. \cite{GrabCutRotherKB04,iccvLempitsky09,MilCutCVPR14,BaiSapIJCV2009,LiSunTanShuACM2004,ProSapIP2007,BoyJolICCV2001,MorBarIP1998}.
Given an input image and some information provided by a user, usually in the form of a scribble or of a bounding box,
the goal is to provide as output a foreground object in such a way as to best reflect the user's intent.
By exploiting high-level, semantic knowledge on the part of the user, which is typically difficult to formalize, we are therefore able to effectively solve segmentation problems which would be otherwise too complex to be tackled using fully automatic segmentation algorithms.
Another example of a  ``constrained'' segmentation problem is image co-segmentation. Given a set of images, the goal here is to jointly segment same or similar foreground objects. The problem was first introduced by Rother \et \cite{CarThoAndVlaCVPR2006} who used histogram matching to simultaneously segment the foreground object out from a given pair of images. 
Recently, several techniques have been proposed which try to co-segment groups containing more than two images, even in the presence of similar backgrounds. Joulin \et \cite{ArmFraJeaCVPR2010}, for example, proposed a discriminative clustering framework, combining normalized cut and kernel methods and the framework has recently been extended in an attempt to handle multiple classes and a significantly larger number of images \cite{ArmFraJeaCVPR2012}. In this work (which is an extended version of \cite{ZemeneP16}), we propose a unified approach to address this kind of problems which can deal naturally with various input modalities, or constraints, and is able to robustly handle noisy annotations on the part of the external source. In particular, we shall focus on interactive segmentation and co-segmentation (in both the unsupervised and the interactive versions).
	
Next, we present our works on CBIR. Image retrieval (CBIR) has recently attracted considerable attention within the computer vision community, especially because of its potential applications such as database retrieval, web and mobile image search. The goal of semantic image search, or content-based image retrieval (CBIR), is to search for a query image from a given image dataset. This is done by computing image similarities based on low-level image features, such as color, texture, shape and spatial relationship of images. Variation of images in illumination, rotation, and orientation has remained a major challenge for CBIR.  Recently, locally constraining the diffusion process has shown its effectiveness on learning the intrinsic manifold structure of a given data. However, existing constrained-diffusion based retrieval methods have several shortcomings. For instance, manual choice of optimal local neighborhood size, do not allow for intrinsic relation among the neighbors, fix initialization vector to extract dense neighbor; which negatively affect the affinity propagation. In CDSIR, leveraging the constrained dominant sets we tackle the above issues. On the other hand, we develop a feature-fusion based image retrieval method known as  Multi-feature Fusion for Image Retrieval Using Constrained Dominant Sets (MfFIR) . Multi-feature based CBIR attacks the CBIR problem by introducing an approach which utilizes multiple low-level visual features of an image. Intuitively, if the to-be-fused feature works well by itself, it is expected that its aggregation with other features will improve the accuracy of the retrieval. Nevertheless, it is quite hard to learn in advance the effectiveness of the to-be-fused features for a specific query image. In MfFIR, we propose a computationally efficient approach to fuse several hand-crafted and deep features, based on the probabilistic distribution of a given membership score of a constrained cluster in an unsupervised manner.
	
We finally discuss our work on the challenging computer vision problem of person re-identification. In this work, for the very first time, we integrate the well known graph and game theoretic clustering method called dominant sets in end-to-end manner. Thereby, we do the optimization of constrained-clustering process in the context of deep learning. Person re-identification aims at retrieving the most similar images to the probe image, from a large-scale gallery set captured by camera networks. Among the challenges which hinder person re-id tasks, include background clutter, pose, viewpoint and illumination variation can be mentioned. Person re-id can be considered as a person retrieval problem based on the ranked similarity score, which is obtained from the pairwise affinities between the probe and the dataset images. However, relying solely on the pairwise affinities of probe-gallery images, ignoring the underlying contextual information between the gallery images often leads to  an undesirable similarity ranking.  To overcome this, we propose an intriguing scheme which treats person-image retrieval problem as a {\em constrained clustering optimization} problem, called deep constrained dominant sets (DCDS). Given a probe and gallery images, we re-formulate person re-id problem as finding a constrained cluster, where the probe image is taken as a constraint (seed) and each cluster corresponds to a set of images corresponding to the same person. By optimizing the constrained clustering  in an end-to-end manner, we naturally leverage the contextual knowledge of a set of images corresponding to the given person-images. We further enhance the performance by integrating an auxiliary net alongside DCDS, which employs a multi-scale ResNet.
To summarize, the main contributions of this thesis are:
\begin{itemize}
	\item It leverages the constrained dominant sets to attack several computer vision problems in both classical and deep flavors.
	\item The proposed DSCIS has a number of interesting features which distinguishes it from existing approaches. Specifically: 1) it solves both image segmentation and co-segmentation in an interactive and unsupervised manner.
	 2) in the case of noiseless scribble inputs, it asks the user to provide {\em only} foreground pixels; 3) it turns out to be {\em robust} in the presence of input noise, allowing the user to draw, e.g., imperfect scribbles.
	\item The proposed Image retrieval methods come with several advantages. In particular, the proposed CDSIR: 1) it constrains the diffusion process by locally extracting dense neighbors whose local neighborhood size (K) is fixed automatically; means that different neighbors can have a different value of K. 2) it has no initialization step; the dynamics, to extract the dense neighbors, can start at any point in the standard simplex 3) it turns out to be {\em robust} to noisy affinity matrices.
	\item On other hand, through MfFIR, we contribute a generic approach which can be applied not only to image retrieval but also to other computer vision problems, such as object detection and person re-identification. Furthermore, unlike existing feature-fusion methods, we propose a simple but efficient entropy-based feature effectiveness weighting system.
	\item Finally, for the very first time, we integrate the well-known clustering method, dominant sets, in a deep neural network (DNN) model. Moreover, we establish a one-to-one correspondence between person re-identification and constrained clustering problem.
\end{itemize}
	
\clearpage
\selectlanguage{english}

\tableofcontents

\cleardoublepage
\listoffigures
\listoftables

\cleardoublepage
\newpage

\setcounter{page}{1}
\pagenumbering{arabic}

\cleardoublepage
\chapter{Dominant Sets and Quadratic Optimization}

\label{chap:intro}

Clustering or partitioning a given data based on the similarity among the data points is a fundamental task in many fields of study such as Machine Learning, Computer Vision, and Statistics. In this chapter, we discuss the well-known graph and game-theoretic pairwise data clustering scheme called Dominant Sets and its constrained variant Constrained Dominant Sets.

In the dominant set framework, the data to be clustered are represented as an undirected edge-weighted graph with no self-loops $G = (V, E,w)$, where $V = \{1, . . . , n\}$ is the vertex set, $E \subseteq V \times V$ is the edge set, and $w : E \rightarrow R_+^*$ is the (positive) weight function. Vertices in $G$ correspond to data points, edges represent neighborhood relationships, and edge-weights reflect similarity between pairs of linked vertices. As customary, we represent the graph $G$ with the corresponding weighted adjacency (or similarity) matrix, which is the $n \times n$ nonnegative, symmetric matrix $A = (a_{ij})$ defined as $a_{ij} = w(i, j)$, if $(i, j) \in E$, and $a_{ij} = 0$ otherwise. Since in $G$ there are no self-loops, note that all entries on the main diagonal of $A$ are zero.

For a non-empty subset $S \subseteq V$, $i \in S$, and $j \notin S$, define
\begin{equation}
\label{eq1}
\phi_S(i,j)=a_{ij}-\frac{1}{|S|} \sum_{k \in S} a_{ik}~.
\end{equation}
This quantity measures the (relative) similarity between nodes $j$ and $i$, with respect to the average similarity between node $i$ and its neighbors in $S$. Note that $\phi_S(i,j)$ can be either positive or negative. Next, to each vertex $i \in S$ we assign a weight defined (recursively) as follows:
\begin{equation}
w_S(i)=
\begin{cases}
1,&\text{if\quad $|S|=1$},\\
\sum_{j \in S \setminus \{i\}} \phi_{S \setminus \{i\}}(j,i)w_{S \setminus \{i\}}(j),&\text{otherwise}~.
\end{cases}
\end{equation}
Intuitively, $w_S(i)$ gives us a measure of the overall similarity between vertex $i$ and the vertices of $S\setminus \{i\}$
with respect to the overall similarity among the vertices in $S\setminus \{i\}$. Therefore, a positive $w_S(i)$ indicates that adding $i$ into its neighbors in $S$ will increase the internal coherence of the set, whereas in the presence of a negative value we expect the overall coherence to be decreased. 

A non-empty subset of vertices $S \subseteq V$ such that $W(T) > 0$ for any non-empty $T \subseteq S$, is said to be a {\em dominant set} if:
\begin{enumerate}
\item $w_S(i)>0$, for all $i \in S$,
\item $w_{S \cup \{i\}}(i)<0$, for all $i \notin S$.
\end{enumerate}
It is evident from the definition that a dominant set satisfies the two basic properties of a cluster: internal coherence and external incoherence. Condition 1 indicates that a dominant set is internally coherent, while condition 2 implies that
this coherence will be destroyed by the addition of any vertex from outside. In other words, a dominant set is a maximally coherent data set.

Now, consider the following linearly-constrained quadratic optimization problem:
\begin{equation}
\label{eq2}
\begin{array}{ll}
   \text{maximize }  &  f(\x) = \x' A \x \\
   \text{subject to} &  \mathbf{x} \in \Delta
\end{array}
\end{equation}
where a prime denotes transposition and  
$$
\Delta=\left\{ \x \in R^n~:~ \sum_{i=1}^n x_i = 1, \text{ and } x_i \geq 0 \text{ for all } i=1 \ldots n \right\}
$$ 
is the standard simplex of $R^n$.
In \cite{PavPelCVPR2003,PavPel07} a connection is established between dominant sets and the local solutions of \eqref{eq2}. In particular, it is shown that if $S$ is a dominant set then its ``weighted characteristic vector,'' which is the vector $ {\bf x}^S \in \Delta$ defined as
\begin{displaymath}
x^S_i=
\begin{cases} \frac{w_S(i)}{\sum_{j \in S}w_S(j)},&\text{if\quad $i \in S$},\\ 0,&\text{otherwise}
\end{cases}
\end{displaymath}
is a strict local solution of \eqref{eq2}. Conversely, under mild conditions, it turns out that if $\x$ is a (strict) local solution of program \eqref{eq2} then its ``support''
$$
\sigma(\x) = \{i \in V~:~x_i > 0\}
$$ 
is a dominant set.
By virtue of this result, we can find a dominant set by first localizing a solution of program \eqref{eq2} with an appropriate continuous optimization technique, and then picking up the support set of the solution found. In this sense, we indirectly perform combinatorial optimization via continuous optimization. A generalization of these ideas to hypergraphs has recently been developed in \cite{RotPelPAMI2013}.

\begin{figure}
	\centering
	\includegraphics[width=0.95\linewidth,height=0.3\linewidth]{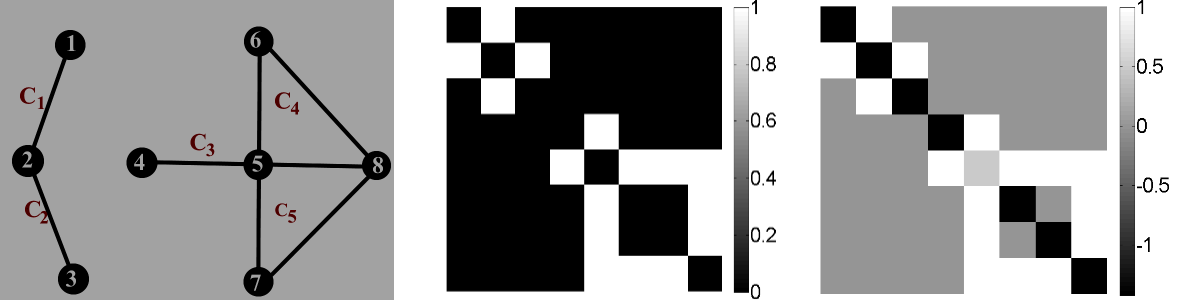}
	\caption{\small An example graph (left), corresponding affinity matrix (middle), and scaled affinity matrix 
		built considering vertex 5 as a user constraint (right). Notation $C_i$ refers to the $i^{th}$ maximal clique.}
	\label{fig:ExamplarGraphAndAffinity}
\end{figure}

Note that, by construction, dominant sets capture compact structures. To deal with arbitrarily shaped clusters, path-based similarity measures can profitably be used \cite{ZemPel15}. In the work reported in this thesis, however, we did not make use of this notion.

\section{Constrained dominant sets}

Let $G=(V,E,w)$ be an edge-weighted graph with $n$ vertices and let $A$ denote as usual its (weighted) adjacency matrix. Given a subset of vertices $S \subseteq V$ and a parameter $ \alpha > 0$, define
the following parameterized family of quadratic programs:
\begin{equation}
\label{eqn:parQP}
\begin{array}{ll}
   \text{maximize }  &  f_S^\alpha(\x) = \x' (A - \alpha \hat I_S) \x \\
   \text{subject to} &  \mathbf{x} \in \Delta
\end{array}
\end{equation}
where $\hat I_S$ is the $n \times n$ diagonal matrix whose diagonal elements are 
set to 1 in correspondence to the vertices contained in $V \setminus S$ and to zero otherwise, and the 0's
represent null square matrices of appropriate dimensions.
In other words, assuming for simplicity that $S$ contains, say, the first $k$ vertices of $V$, we have:

$$
\hat I_S = 
\begin{pmatrix} 
  ~~0~~ & ~~0~~ \\ 
  ~~0~~ & I_{n-k}  
\end{pmatrix}
$$
where $I_{n-k}$ denotes the $(n-k) \times (n-k) $ principal submatrix of the $n \times n$ identity matrix $I$ 
indexed by the elements of $V \setminus S$.
Accordingly, the function $f_S^\alpha$ can also be written as follows:
$$
f_S^\alpha(\x) = \x' A \x - \alpha \x'_{\bar{S}} \x_{\bar{S}}
$$
$\x_{\bar{S}}$ being the $(n-k)$-dimensional vector obtained from
$\x$ by dropping all the components in $S$.
Basically, the function $f_S^\alpha$ is obtained from $f$ by inserting in the
affinity matrix $A$ the value of the parameter $\alpha$ in the main diagonal positions
corresponding to the elements of $V \setminus S$.

Notice that this differs markedly, and indeed
generalizes, the formulation proposed in \cite{PavPel03} for obtaining a hierarchical clustering 
in that here, only a subset of elements in the main diagonal
is allowed to take the $\alpha$ parameter, the other ones being set to zero.
We note in fact that the original (non-regularized) dominant-set formulation (\ref{eq2}) \cite{PavPel07} 
as well as its regularized counterpart described in \cite{PavPel03}
can be considered as degenerate version of ours, corresponding to the cases $S=V$ and $S=\emptyset$, respectively.
It is precisely this increased flexibility which allows us to use
this idea for finding groups of ``constrained'' dominant-set clusters.

We now derive the Karush-Kuhn-Tucker (KKT) conditions for program (\ref{eqn:parQP}),
namely the first-order necessary conditions for local optimality (see, e.g., \cite{Lue84}).
For a point $\x \in \Delta$ to be a KKT-point there should exist $n$
nonnegative real constants $\mu_1 , \ldots , \mu_n$ and an additional real number $\lambda$
such that 
\begin{displaymath}
[(A - \alpha \hat I_S) \x]_i - \lambda + \mu_i = 0
\end{displaymath}
for all $i=1 \ldots n$, and
\begin{displaymath}
\sum_{i=1}^n x_i \mu_i = 0~.
\end{displaymath}
Since both the $x_i$'s and the $\mu_i$'s are nonnegative, the
latter condition is equivalent to saying that $i \in \sigma(\x)$ implies
$\mu_i= 0$, from which we obtain:
\begin{displaymath}
[(A - \alpha \hat I_S) \x]_i ~
\begin{cases} 
~ = ~   \lambda, ~ \mbox{ if } i \in \sigma(\x) \\ 
~ \le ~ \lambda, ~ \mbox{ if } i \notin \sigma(\x)  
\end{cases}
\end{displaymath}
for some constant $\lambda$.
Noting that $\lambda = \x'A\x - \alpha \x'_{\bar{S}} \x_{\bar{S}}$ and recalling the definition of $\hat I_S$,
the KKT conditions can be explicitly rewritten as:
\begin{equation}
\label{eqn:KKT}
\left\{
\begin{array}{llll}
(A\x)_i - \alpha x_i  & =   & \x'A\x - \alpha \x'_{\bar{S}} \x_{\bar{S}}, & \mbox{ if } i \in \sigma(\x) \mbox{ and } i \notin S \\
(A\x)_i               & =   & \x'A\x - \alpha \x'_{\bar{S}} \x_{\bar{S}}, & \mbox{ if } i \in \sigma(\x) \mbox{ and } i \in S \\
(A\x)_i               & \le & \x'A\x - \alpha \x'_{\bar{S}} \x_{\bar{S}}, & \mbox{ if } i \notin \sigma(\x)
\end{array}
\right.
\end{equation}

We are now in a position to discuss the main results which motivate the algorithm presented in this thesis.
Note that, in the sequel, given a subset of vertices $S\subseteq V$, the face of $\Delta$ corresponding to $S$ is given by: $\Delta_{S}=\{x\in \Delta : \sigma (x)\subseteq S\}$.

\begin{proposition}
\label{prop:gamma}
Let $S \subseteq V$, with $S \neq \emptyset$.
Define
\begin{equation}
\label{eqn:defgamma}
\gamma_S = \max_{ \x \in \Delta_{V \setminus S}} 
\min_{i \in S} ~ \frac{\x' A \x - (A\x)_i}{\x' \x}
\end{equation}
and let $\alpha > \gamma_S$.
If $\x$ is a local maximizer of $f_S^\alpha$ in $\Delta$, then
$\sigma(\x) \cap S \neq \emptyset$.
\end{proposition}

\proof
Let $\x$ be a local maximizer of $f_S^\alpha$ in $\Delta$, and suppose
by contradiction that no element of $\sigma(\x)$ belongs to $S$ or, in other words,
that $\x \in \Delta_{V \setminus S}$.
By letting 
$$
i = \argmin_{j \in S} ~ \frac{\x' A \x - (A\x)_j}{\x' \x}
$$
and observing that $\sigma(\x) \subseteq V \setminus S$ implies $\x' \x = \x'_{\bar{S}} \x_{\bar{S}}$,
we have: 
$$
\alpha >  \gamma_S \ge 
\frac{\x' A \x - (A\x)_i}{\x' \x} =
\frac{\x' A \x - (A\x)_i}{\x'_{\bar{S}} \x_{\bar{S}}}~.
$$
Hence, $(A\x)_i > \x' A \x - \alpha \x'_{\bar{S}} \x_{\bar{S}}$ for $i \notin \sigma(\x)$, 
but this violates the KKT conditions (\ref{eqn:KKT}), thereby proving the proposition.
\endproof

The following proposition provides a useful and easy-to-compute upper bound for $\gamma_S$. 

\begin{proposition}
\label{prop:bound}
Let $S \subseteq V$, with $S \neq \emptyset$. Then, 
\begin{equation}
\label{eqn:bound}
\gamma_S \le \lambda_{\max}(A_{V \setminus S})
\end{equation}
where $\lambda_{\max}(A_{V \setminus S})$ is the largest eigenvalue of the principal submatrix of $A$
indexed by the elements of $V \setminus S$. 
\end{proposition}
\proof
Let $\x$ be a point in $\Delta_{V \setminus S}$ which attains the maximum $\gamma_S$
as defined in (\ref{eqn:defgamma}).
Using the Rayleigh-Ritz theorem \cite{HorJon85} and the fact that $\sigma(\x)\subseteq V \setminus S$, we obtain:
$$
\lambda_{\max}(A_{V \setminus S}) \ge \frac{\x'_{\bar{S}} A_{V \setminus S} \x_{\bar{S}}}{\x'_{\bar{S}} \x_{\bar{S}}}
= \frac{\x' A \x}{\x' \x}~.
$$
Now, define $\gamma_S(\x) = \max \{ (A \x)_i~:~i \in S \}$.
Since $A$ is nonnegative so is $\gamma_S(\x)$, and recalling the definition of $\gamma_S$ we get:
$$
\frac{\x' A \x}{\x' \x} \ge \frac{\x' A \x - \gamma_S(\x)}{\x' \x} = \gamma_S
$$
which concludes the proof.
\endproof

The two previous propositions provide us with a simple technique to determine dominant-set clusters containing user-selected vertices. Indeed, if $S$ is the set of vertices selected by the user, by setting 
\begin{equation}
\label{alphabound}
\alpha > \lambda_{\max}(A_{V \setminus S})
\end{equation}
we are guaranteed that all local solutions of (\ref{eqn:parQP}) will have a support 
that necessarily contains elements of $S$.
%
%
Note that this does not necessarily imply that the (support of the) solution found corresponds to a dominant-set cluster of the original affinity matrix $A$, as adding the parameter $-\alpha$ on a portion of the main diagonal intrinsically changes the scale of the underlying problem. However, we have obtained extensive empirical evidence which supports a conjecture which turns out to be very useful for our interactive image segmentation application.

To illustrate the idea, let us consider the case where edge-weights are binary, which basically
means that the input graph is unweighted. In this case, it is known that dominant sets correspond to maximal cliques \cite{PavPel07}. Let $G=(V,E)$ be our unweighted graph and let $S$ be a subset of its vertices.
For the sake of simplicity, we distinguish three different situations of increasing generality.

\noindent
{\bf Case 1.} The set $S$ is a singleton, say $S = \{u\}$. In this case, we know from 
Proposition \ref{prop:bound} that all solutions $\x$ of 
$f_\alpha^S$ over $\Delta$ will have a support which contains $u$, that is $u \in \sigma(\x)$.
Indeed, we conjecture that there will be a unique local (and hence global) solution here whose support
coincides with the {\em union} of all maximal cliques of $G$ which contain $u$. 

\noindent
{\bf Case 2.} The set $S$ is a clique, not necessarily maximal. In this case, 
Proposition \ref{prop:bound} predicts that all solutions $\x$ of (\ref{eqn:parQP})
will contain at least one vertex from $S$.
Here, we claim that indeed the support of local solutions is the union of the maximal cliques that contain $S$.

\noindent
{\bf Case 3.} The set $S$ is not a clique, but it can be decomposed as a collection of (possibly overlapping)
maximal cliques $C_1, C_2, ..., C_k$ (maximal with respect to the subgraph induced by $S$).
In this case, we claim that if $\x$ is a local solution, then its support can be obtained by taking the union of
all maximal cliques of $G$ containing one of the cliques $C_i$ in $S$.

To make our discussion clearer, consider the graph shown in Fig. \ref{fig:ExamplarGraphAndAffinity}. 
In order to test whether our claims hold, we used as the set $S$ different combinations of vertices, and
enumerated all local solutions of (\ref{eqn:parQP}) by multi-start replicator dynamics (see Section \ref{sec:DynamicsToExtractCDS}).
Some results are shown below, where on the left-hand side we indicate the set $S$, while on
the right hand-side we show the supports provided as output by the different runs of the algorithm. 
\begin{table}[h]
\begin{tabular}
{p{2.2cm}  p{0.5cm}  p{6cm}}
1.  ~~$S = \emptyset$    &   $\Rightarrow$     & $\sigma(\x_1) = \{ 5, 6, 8 \}$, ~$\sigma(\x_2) = \{ 5, 7, 8 \}$\\
2.  ~~$S = \{ 2 \}$       &   $\Rightarrow$     & $\sigma(\x) = \{ 1, 2, 3 \}$ \\
3.  ~~$S = \{ 5 \}$       &   $\Rightarrow$     & $\sigma(\x) = \{ 4, 5, 6, 7, 8 \}$ \\
4.  ~~$S = \{ 4, 5 \}$    &   $\Rightarrow$     & $\sigma(\x) = \{ 4, 5 \}$ \\
5.  ~~$S = \{ 5, 8 \}$    &   $\Rightarrow$     & $\sigma(\x) = \{ 5, 6, 7, 8 \}$ \\
6.  ~~$S = \{ 1, 4 \}$    &   $\Rightarrow$     & $\sigma(\x_1) = \{ 1, 2 \}$, ~$\sigma(\x_2) = \{ 4, 5 \}$\\
7.  ~~$S = \{ 2, 5, 8 \}$ &   $\Rightarrow$     & $\sigma(\x_1) = \{ 1, 2, 3 \}$, ~$\sigma(\x_2) = \{ 5, 6, 7, 8 \}$
\end{tabular}
\end{table}

Notice that in the unconstrained case ($S=\emptyset$), the algorithm returns the two largest cliques, depending on 
the starting point. We refer the reader to \cite{PelTor06} (and references therein) for a thorough analysis of the use of replicator and similar dynamics for the (unconstrained) maximum clique problem.

The previous observations can be summarized in the following general statement which does comprise all three cases. 
Let $S = C_1 \cup C_2 \cup \ldots \cup C_k$ ($k \ge 1$) be a subset of vertices of $G$, 
consisting of a collection of cliques $C_i$ ($i=1 \ldots k$).
Suppose that condition (\ref{alphabound}) holds, and let 
$\x$ be a local solution of (\ref{eqn:parQP}). Then, $\sigma(\x)$ consists of the union of 
all maximal cliques containing some clique $C_i$ of $S$.


\section{Finding constrained dominant sets using replicator dynamics}
\label{sec:DynamicsToExtractCDS}

Given an arbitrary real-valued $n \times n$ matrix $W= (w_{ij})$, consider the following continuous-time dynamical system
\begin{equation}
	\label{ContinuousReplicator}
	\dot{x} = x_i\left( (W\vct{x})_i - \vct{x}'W\vct{x} \right)
\end{equation}
for $i = 1 \ldots n$, where a dot signifies derivative w.r.t. time, and its discrete-time counterpart:
\begin{equation}
	\label{eqn:Replicator}
	x_i{(t+1)} = x_i{(t)} \frac{C + (W\x{(t)})_i}{C + \x{(t)}'W\x{(t)}}
\end{equation}
where $C$ is a proper constant to avoid negative values in the numerator (and denominator).  
These are known as {\em replicator dynamics} in evolutionary game theory \cite{Wei95,HofSig98} and it turns out that, for a large constant $C$, (\ref{eqn:Replicator}) is essentially an Euler discretization of (\ref{ContinuousReplicator}).

It is readily seen that the standard simplex $\Delta$ is invariant under these dynamics, which means that every trajectory starting 
in $\Delta$ will remain in $\Delta$ for all future times. 
Moreover, their stationary points, i.e., the points satisfying 
$\dot{x}= 0$ for (\ref{ContinuousReplicator}) and $x_i{(t+1)}=x_i{(t)}$ for (\ref{eqn:Replicator}), 
coincide and are the solutions of the equations:
$$
x_i\left( (W\vct{x})_i - \vct{x}'W\vct{x} \right) = 0~.
$$
A stationary point $\vct x$ is said to be {\em asymptotically stable} if every
trajectory which starts close enough to $\vct x$ will eventually converge to it.

The following result, known as the Fundamental Theorem of Natural Selection \cite{HofSig98},
provides us useful information concerning the convergence properties of replicator dynamics.

\begin{theorem}
If $W$ is symmetric ($W=W'$), then the function $\x{(t)}'W\x{(t)}$ 
is strictly increasing along any nonstationary trajectory, 
under both continuous-time (\ref{ContinuousReplicator}) and discrete-time (\ref{eqn:Replicator})
replicator dynamics. Furthermore, any such trajectory converges to a stationary point. Finally, a vector $\vct x \in \Delta$ is
asymptotically stable under (\ref{ContinuousReplicator}) and (\ref{eqn:Replicator}) if and only if $\vct x$ is a 
strict local maximizer of $\vct x ' W \vct x$ on $\Delta$.
\end{theorem}

Thanks to these properties, replicator dynamics naturally suggest themselves as a simple heuristics for finding (constrained) dominant sets \cite{PavPel07}.
In our case, the matrix $W$ is given by

$$
W = A - \alpha \hat I_S
$$
which, in the discrete-time case, yields:
\begin{equation}
	\label{eqn:Replicator2}
x_i{(t+1)} ~=~
\begin{cases} 
x_i{(t)} \frac{C + (A\x{(t)})_i}{C + \x{(t)}'(A-\alpha \hat I_S)\x{(t)}} ~ \mbox{ if } i \in S  \\
\\
x_i{(t)} \frac{C+ (A\x{(t)})_i  - \alpha x_i{(t)}}{C+ \x{(t)}'(A-\alpha \hat I_S)\x{(t)}} ~ \mbox{ if } i \notin S 
\end{cases}
\end{equation}



Since the process cannot leave the boundary of $\Delta$, it is customary to start the dynamics from some 
interior point, a common choice being the barycenter of $\Delta$. This prevents the search from being initially biased in favor of any particular solution. By virtue of our theoretical results, we know that when the algorithm converges to a vector $\vct x$, its 
support $\sigma(\vct x) $ will correspond to a constrained dominant set, while its positive components will reflect the membership score of the selected vertices.  

Clearly, the behavior of the algorithm depends on the choice of the parameter $\alpha$. Our own experience is that $\alpha$ might affect the convergence time (number of steps of the replicator dynamics) as well as the distribution of the membership scores of the final solution (i.e., the components of the converged vector). In particular, we observed that the membership scores assigned to the constrained dominant-set vertices become larger and larger (thereby making the scores of the other dominant-set vertices smaller and smaller) as $\alpha$ increases. This phenomenon, however, manifests itself more sensibly, and might become an issue, only for large values of $\alpha$. No significant effect on the algorithm's performance has been observed for reasonable choices of the parameter. 
Accordingly, we recommend using a reasonably small value for $\alpha$, close to the lower bound predicted by our theoretical results. This is what we actually did in all the experiments reported below. 
As for the parameter $C$ in (\ref{eqn:Replicator2}), its function is only 
to scale the matrix $A - \alpha \hat I_S$ properly to avoid negative values. An obvious choice would be $C = \alpha$, 
which is the value we used in our experiments.

Although in the experiments reported in this thesis we used the replicator dynamics described above, we mention
a faster alternative to solve linearly constrained quadratic optimization problems like ours, namely {\em Infection and Immunization Dynamics} (InImDyn) \cite{RotPelBomCVIU2011}.
Each step of InImDyn has a linear time/space complexity as opposed to the quadratic per-step complexity of replicator dynamics, and is therefore to be preferred in the presence of large affinity matrices.

\section{Summary}

In this chapter, we briefly introduced the well-known graph and game theoretic clustering algorithm called Dominant Sets, and its variant Constrained Dominant Sets. 
\newpage
\chapter{Dominant Sets for ``Constrained'' Image Segmentation}

\label{chap:intro}

Image segmentation has come a long way since the early days of computer vision, and still remains a challenging task. Modern variations of the classical (purely bottom-up) approach, involve, e.g., some form of user assistance (interactive segmentation) or ask for the simultaneous segmentation of two or more images (co-segmentation). At an abstract level, all these variants can be thought of as ``constrained'' versions of the original formulation, whereby the segmentation process is guided by some external source of information. In this chapter, we propose a new approach to tackle this kind of problems in a unified way. Our work is based on some properties of a family of quadratic optimization problems related to {\em dominant sets}, a graph-theoretic notion of a cluster which generalizes the concept of a maximal clique to edge-weighted graphs. In particular, we show that by properly controlling a regularization parameter which determines the structure and the scale of the underlying problem, we are in a position to extract groups of dominant-set clusters that are constrained to contain predefined elements. In particular, we shall focus on interactive segmentation and co-segmentation (in both the unsupervised and the interactive versions). The proposed algorithm can deal naturally with several types of constraints and input modalities, including scribbles, sloppy contours and bounding boxes, and is able to robustly handle noisy annotations on the part of the user. Experiments on standard benchmark datasets show the effectiveness of our approach as compared to state-of-the-art algorithms on a variety of natural images under several input conditions and constraints.

\section{Introduction}
Segmentation is arguably one of the oldest and best-studied problems in computer vision, being a fundamental step in a variety of real-world applications, and yet remains a challenging task \cite{RichardSzeliski11} \cite{ForsPonse11}.Besides the standard, purely bottom-up formulation, which involves partitioning an input image into coherent regions, in the past few years several variants have been proposed which are attracting increasing attention within the community. Most of them usually take the form of a ``constrained'' version of the original problem, whereby the segmentation process is guided by some external source of information. For example, user-assisted (or ``interactive'') segmentation has become quite popular nowadays, especially because of its potential applications in problems such as image and video editing, medical image analysis, etc. \cite{GrabCutRotherKB04,iccvLempitsky09,MilCutCVPR14,BaiSapIJCV2009,LiSunTanShuACM2004,ProSapIP2007,BoyJolICCV2001,MorBarIP1998}. Given an input image and some information provided by a user, usually in the form of a scribble or of a bounding box, the goal is to provide as output a foreground object in such a way as to best reflect the user's intent. By exploiting high-level, semantic knowledge on the part of the user, which is typically difficult to formalize, we are therefore able to effectively solve segmentation problems which would be otherwise too complex to be tackled using fully automatic segmentation algorithms.

Another example of a  ``constrained'' segmentation problem is co-segmentation. Given a set of images, the goal here is to jointly segment same or similar foreground objects. The problem was first introduced by Rother \et \cite{CarThoAndVlaCVPR2006} who used histogram matching to simultaneously segment the foreground object out from a given pair of images. 
Recently, several techniques have been proposed which try to co-segment groups containing more than two images, even in the presence of similar backgrounds. Joulin \et \cite{ArmFraJeaCVPR2010}, for example, proposed a discriminative clustering framework, combining normalized cut and kernel methods and the framework has recently been extended in an attempt to handle multiple classes and a significantly larger number of images \cite{ArmFraJeaCVPR2012}.

In this chapter (which is an extended version of \cite{ZemeneP16}), we propose a unified approach to address this kind of problems which can deal naturally with various input modalities, or constraints, and is able to robustly handle noisy annotations on the part of the external source. In particular, we shall focus on interactive segmentation and co-segmentation (in both the unsupervised and the interactive versions).
	

Although various kinds of constraints can be envisaged to encode top-down information in segmentation processes, our work is focused on what we might refer to as ``first-order'' (or unary) constraints, which require that one or more ``seed'' points be part of the extracted group. Second- or higher-order constraints, of the type discussed for example in \cite{YuShi04,Kul+09,Eri+11}, which include must-link constraints (pairs of points that should belong to the same cluster) and cannot-link constraints (pairs of points that should belong to different clusters), will not be treated here, although it is not difficult to adapt our framework to deal with these cases.


	%
Our approach is based on some properties of a parameterized family of quadratic optimization problems related to dominant-set clusters, a well-known generalization of the notion of maximal cliques to edge-weighted graph which have proven to be extremely effective in a variety of computer vision problems, including (automatic) image and video segmentation \cite{PavPelCVPR2003,PavPel07} (see \cite{RotPel17} for a recent review).
In particular, we show that by properly controlling a regularization parameter which determines the structure and the scale of the underlying problem, we are in a position to extract groups of dominant-set clusters which are constrained to contain user-selected elements. We provide bounds that allow us to control this process, which are based on the spectral properties of certain submatrices of the original affinity matrix. The resulting algorithm has a number of interesting features which distinguishes it from existing approaches. Specifically: 1) it is able to deal in a flexible manner with {\em both} scribble-based and boundary-based input modalities (such as sloppy contours and bounding boxes); 2) in the case of noiseless scribble inputs, it asks the user to provide {\em only} foreground pixels; 3) it turns out to be {\em robust} in the presence of input noise, allowing the user to draw, e.g., imperfect scribbles or loose bounding boxes. Experimental results on standard benchmark datasets demonstrate the effectiveness of our approach as compared to state-of-the-art algorithms on a wide variety of natural images under several input conditions.
    \begin{figure*}[t]
		
		\centering
		\includegraphics[width=0.85\linewidth,trim=0cm 8cm 0cm 0cm,clip ]{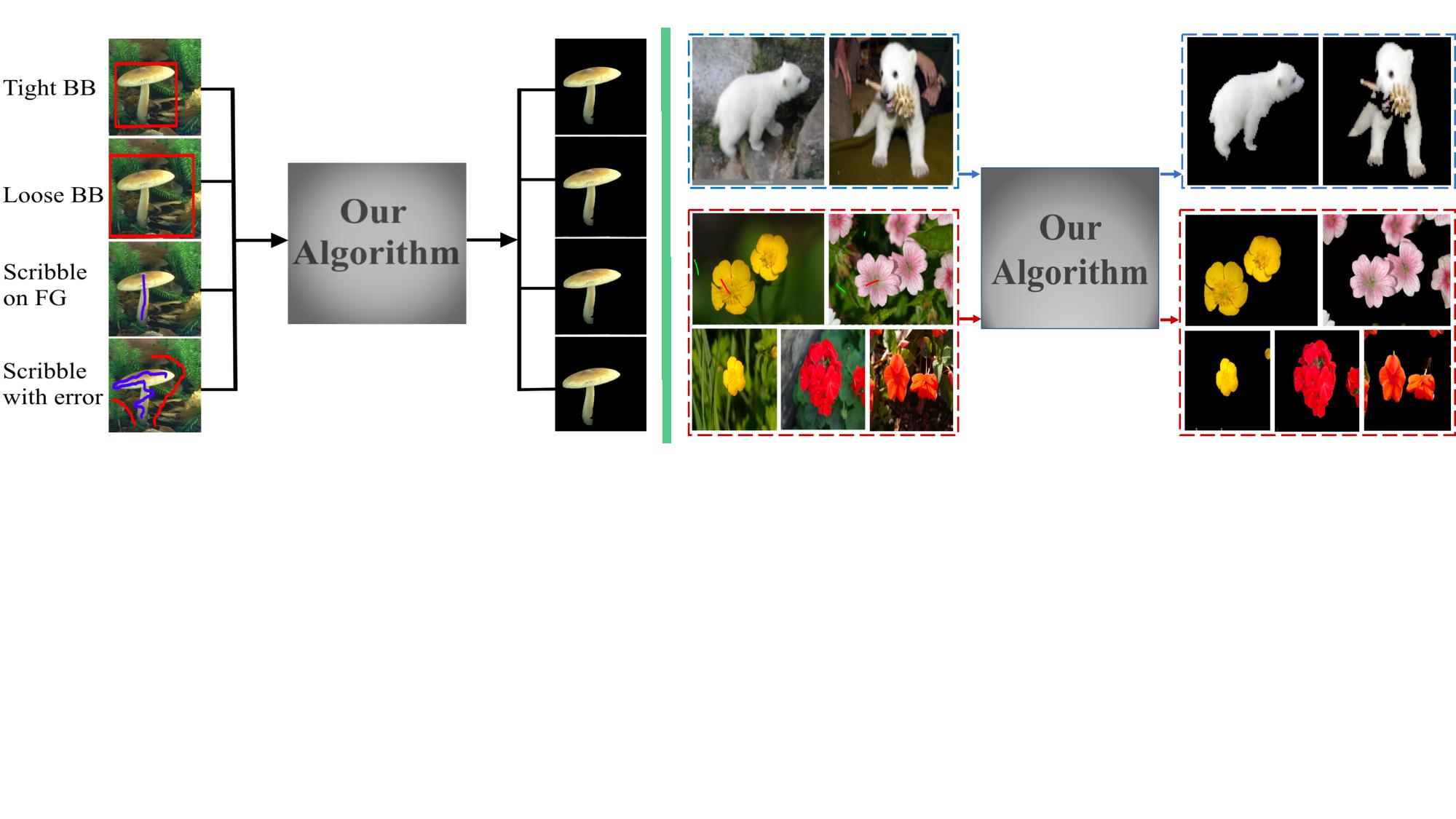}\\
		\caption{\small{ \bf Left:} An example of our interactive image segmentation method and its outputs, with different user annotation. Respectively from top to bottom, tight bounding box (Tight BB), loose bounding box (Loose BB), a scribble made (only) on the foreground object (Scribble on FG) and scribbles with errors. \textbf{Right:} Blue and Red dash-line boxes, show an example of our unsupervised and interactive co-segmentation methods, respectively. }
		\label{fig:InputModalities}
	\end{figure*}
Figure \ref{fig:InputModalities} shows some examples of how our system works in both interactive segmentation (in the presence of different input annotations) and co-segmentation settings.

\subsection{Background}

Existing interactive segmentation interfaces fall into two broad categories, depending on whether the user annotation is given in terms of a scribble or of a bounding box, and supporters of the two approaches have both good reasons to prefer one modality against the other.
For example, Wu et al. \cite{MilCutCVPR14} claim that bounding boxes are the most natural and economical form in terms of the amount of user interaction, and develop a multiple instance learning algorithm that extracts an arbitrary object located inside a tight bounding box at unknown location.
Yu et al. \cite{LOOSECUTcorr15} also support the bounding-box approach, though their algorithm is different from others in that it does not need bounding boxes tightly enclosing the object of interest, whose production of course increases the annotation burden. They provide an algorithm, based on a Markov Random Field (MRF) energy function, that can handle input bounding box that only loosely covers the foreground object. 
Xian et al. \cite{XiaZhaCheXuDinCoRR2015} propose a method which avoids the limitations of existing bounding box methods - region of interest (ROI) based methods, though they need much less user interaction, their performance is sensitive to initial ROI. 
	
On the other hand, several researchers, arguing that boundary-based interactive segmentation such as intelligent scissors \cite{MorBarIP1998} requires the user to trace the whole boundary of the object, which is usually a time-consuming and tedious process, support scribble-based segmentation. Bai et al. \cite{BaiWuCVPR2014}, for example, propose a model based on ratio energy function which can be optimized using an iterated graph cut algorithm, which tolerates errors in the user input.
In general, the input modality in an interactive segmentation algorithm affects both its accuracy and its ease of use. Existing methods work typically on a single modality and they focus on how to use that input most effectively. However, as noted recently by Jain and Grauman \cite{JainGraICCV2013}, sticking to one annotation form leads to a suboptimal tradeoff between human and machine effort, and they tried to estimate how much user input is required to sufficiently segment a novel input.

The co-segmentation problem has also been addressed using user interaction \cite{DhrAdaDevJieTsuCVPR2010, XinJiaLinMinTIP2015}. Here, a user adds guidance, usually in the form of scribbles, on foreground objects of some of the input images. Batra \et \cite{DhrAdaDevJieTsuCVPR2010} proposed an extension of the (single-image) interactive segmentation algorithm of Boykov and Jolly \cite{BoyJolICCV2001}. They also proposed an algorithm that enables users to quickly guide the output of the co-segmentation algorithm towards the desired output via scribbles. Given scribbles, both on the background and the foreground, on some of the images, they cast the labeling problem as energy minimization defined over graphs constructed over each image in a group. Dong \et \cite{XinJiaLinMinTIP2015} proposed a method using global and local energy optimization. Given background and foreground scribbles, they built a foreground and a background Gaussian mixture model (GMM) which are used as global guide information from users. By considering the local neighborhood consistency, they built the local energy as the local smooth term which is automatically learned using spline regression. The minimization problem of the energy function is then converted into constrained quadratic programming (QP) problem, where an iterative optimization strategy is designed for the computational efficiency.

\section{Application to interactive image segmentation}

In this section, we apply CDS to the interactive image segmentation problem. As input modalities we consider scribbles as well as boundary-based approaches (in particular, bounding boxes) and, in both cases,
we show how the system is robust under input perturbations, namely imperfect scribbles or loose bounding boxes.

In this application the vertices of the underlying graph $G$ represent the pixels of the input image (or superpixels, as discussed below), and the edge-weights reflect the similarity between them.
As for the set $S$, its content depends on whether we are using scribbles or bounding boxes 
as the user annotation modality. 
In particular, in the case of scribbles, $S$ represents precisely those pixels that have been
manually selected by the user. In the case of boundary-based annotation 
instead, it is taken to contain only the pixels comprising the box boundary,
which are supposed to represent the background scene.
Accordingly, the union of the extracted dominant sets, say $\mathcal{L}$ dominant sets are extracted which contain the set $S$,  as described in the previous section and below, $\mathbf{UDS}=\mathcal{D}_1 \cup \mathcal{D}_2 ..... \cup \mathcal{D}_{\mathcal{L}}$, represents either the foreground object or the background scene depending on the input modality. For scribble-based approach the extracted set, $\mathbf{UDS}$, represent the segmentation result, while in the boundary-based approach we provide as output the complement of the extracted set, namely $\mathbf{V}\setminus \mathbf{UDS}$. 

\begin{figure}
	\centering
	\includegraphics[width=0.95\linewidth, height=0.45\linewidth]{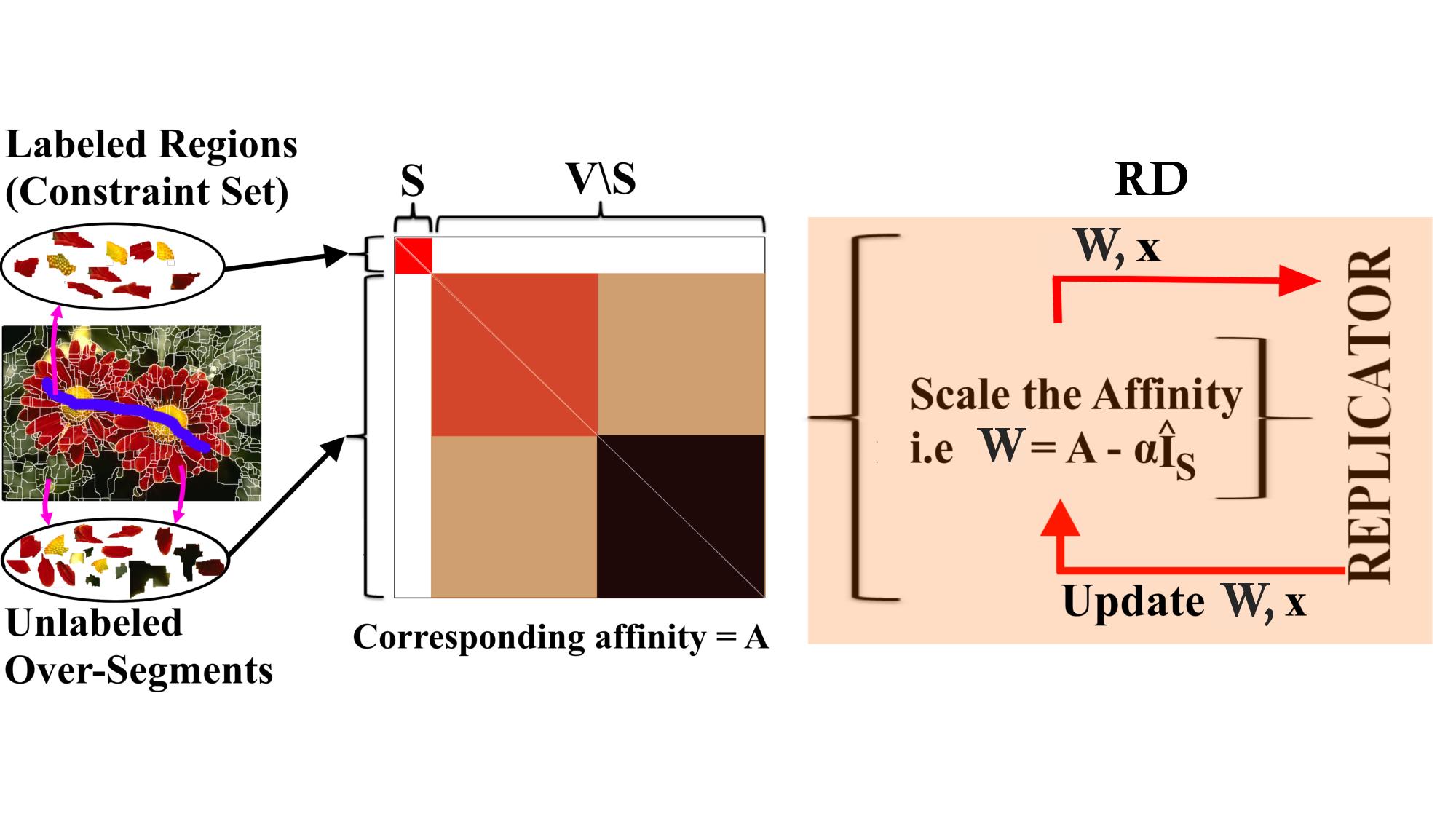}
	\caption{\small Overview of our interactive segmentation system. \textbf{Left:} Over-segmented image (output of the UCM-OWT algorithm \cite{MalikHierarchical}) with a user scribble (blue label). \textbf{Middle:} The corresponding affinity matrix, using each over-segments as a node, showing its two parts: $S$, the constraint set which contains the user labels, and $V\setminus S$, the part of the graph which takes the regularization parameter $\alpha$. \textbf{Right:} The optimization RD (Replicator Dynamics), starts from the barycenter and extracts the first dominant set and update $\mathbf{x}$ and $\mathbf{M}$, for the next extraction till all the dominant sets which contain the user labeled regions are extracted.}
	\label{fig:Framework}
\end{figure}

Figure \ref{fig:Framework} shows the pipeline of our system. 
Many segmentation tasks reduce their complexity by using superpixels (a.k.a. over-segments) as a preprocessing step \cite{MilCutCVPR14,LOOSECUTcorr15,HoiEfrHebICCV2005} \cite{WanJiaHuaZhaQuaCVPR2008,XiaQuaICCV2009}. 
While \cite{MilCutCVPR14} used SLIC superpixels \cite{SLIC-superpixels-TPAMI12}, \cite{LOOSECUTcorr15} used a recent superpixel algorithm \cite{ZhoJuWanWACV2015} which considers not only the color/feature information but also boundary smoothness among the superpixels. 
In this work, we used the over-segments obtained from Ultrametric Contour Map (UCM) which is constructed from Oriented Watershed Transform (OWT) using globalized probability of boundary (gPb) signal as an input \cite{MalikHierarchical}. %

We then construct a graph $G$ where the vertices represent over-segments and 
the similarity (edge-weight) between any two of them is obtained using a standard Gaussian kernel
$$\mat A^\sigma_{ij}=\ind{i\neq j}exp(\Vert\vct f_i-\vct f_j\Vert^2/{2\sigma^2})$$
where $\vct f_i$, is the feature vector of the $i^{th}$ over-segment, $\sigma$ is the free scale parameter, and $\ind{P}=1$ if $P$ is true, $0$ otherwise.

Given the affinity matrix $A$ and the set $S$ as described before, the system constructs
the regularized matrix $M=A-\alpha \hat I_S$, with $\alpha$ chosen as prescribed in (\ref{alphabound}).
Then, the replicator dynamics (\ref{eqn:Replicator}) are run (starting them from the simplex barycenter) until they converge to some solution vector $\x$. We then take the support of $\x$, remove the
corresponding vertices from the graph and restart the replicator dynamics until all
the elements of $S$ are extracted.

\subsection{Experiments and results}

As mentioned above, the vertices of our graph represents over-segments
and edge weights (similarities) are built from the median of the color of all pixels in RGB, HSV, and L*a*b* color spaces, and Leung-Malik (LM) Filter Bank \cite{MalikLeungLMfilterIJCV01}. The number of dimensions of feature vectors for each over-segment is then 57 (three for each of the RGB, L*a*b*, and HSV color spaces, and 48 for LM Filter Bank).

In practice, the performance of graph-based algorithms that use Gaussian kernel, as we do, is sensitive to the selection of the scale parameter $\sigma$. In our experiments, we have reported three different results based on the way $\sigma$ is chosen: $1)$ CDS\_Best\_Sigma, in this case the best parameter $\sigma$  is selected on a per-image basis, which indeed can be thought of as the optimal result (or upper bound) of the framework. $2) $ CDS\_Single\_Sigma, the best parameter in this case is selected on a per-database basis tuning $\sigma$ in some fixed range, which in our case is between 0.05 and 0.2. $3)$ CDS\_Self\_Tuning, the $\sigma^2$ in the above equation is replaced, based on \cite{ManPieNIPS2004}, by $\sigma_i*\sigma_j$, where $\sigma_i = mean(KNN(f_i))$, the mean of the K\_Nearest\_Neighbor of the sample $f_i$, K is fixed in all the experiment as 7. 

\textbf{Datasets:} We conduct four different experiments on the well-known GrabCut dataset \cite{GrabCutRotherKB04} which has been used as a benchmark in many computer vision tasks \cite{LiMN13}\cite{iccvLempitsky09,TangECCV14,OneCutICCV13,MilCutCVPR14,LOOSECUTcorr15} \cite{PriMorCohCVPR2010,YanCaiZheLuoIP2010}. The dataset contains 50 images together with manually-labeled segmentation ground truth. The same bounding boxes as those in \cite{iccvLempitsky09} is used as a baseline bounding box. We also evaluated our scribbled-based approach using the well known Berkeley dataset which contains 100 images.

\textbf{Metrics:}
We evaluate the approach using different metrics: error rate, fraction of misclassified pixels within the bounding box, Jaccard index which is given by, following \cite{EvaluationMetricsMcGuinnessO10}, $J$ = $\frac{|GT \cap O|}{|GT \cup O|}$, where $GT$ is the ground truth and $O$ is the output. The third metric is the Dice Similarity Coefficient ($DSC$), which measures the overlap between two segmented object volume, and is computed as $DSC = \frac{2*|GT \cap O|}{|GT|+|O|}$.

\textbf{Annotations:} 
In interactive image segmentation, users provide annotations which guides the segmentation.
A user usually provides information in different forms such as scribbles and bounding boxes. The input modality affects both its accuracy and ease-of-use \cite{JainGraICCV2013}. However, existing methods fix themselves to one input modality and focus on how to use that input information effectively. This leads to a suboptimal tradeoff in user and machine effort. Jain et al. \cite{JainGraICCV2013} estimates how much user input is required to sufficiently segment a given image. In this work as we have proposed an interactive framework, figure \ref{fig:InputModalities}, which can take any type of input modalities we will use four different type of annotations: bounding box, loose bounding box, scribbles - only on the object of interest -, and scribbles with error as of \cite{BaiWuCVPR2014}.

\subsubsection{Scribble based segmentation} \label{Sec.Scribble}
Given labels on the foreground as constraint set, we built the graph and collect (iteratively) all unlabeled regions (nodes of the graph) by extracting dominant set(s) that contains the constraint set (user scribbles). We provided quantitative comparison against several recent state-of-the-art interactive image segmentation methods which uses scribbles as a form of human annotation: \cite{BoyJolICCV2001}, Lazy Snapping \cite{LiSunTanShuACM2004}, Geodesic Segmentation \cite{BaiSapIJCV2009}, Random Walker \cite{GraPAMI2006}, Transduction \cite{DucAudKerPonFloCVPR2008} , Geodesic Graph Cut \cite{PriMorCohCVPR2010}, Constrained Random Walker \cite{YanCaiZheLuoIP2010}.

We have also compared the performance of our algorithm againts Biased Normalized Cut (BNC) \cite{MajiNishVishMalCVPR11}, an extension of normalized cut, which incorporates a quadratic constraint (bias or prior guess) on the solution $\x$, where the final solution is a weighted combination of the eigenvectors of normalized Laplacian matrix. In our experiments we have used the optimal parameters according to \cite{MajiNishVishMalCVPR11} to obtain the most out of the algorithm.
We also provide some qualitative comparisons with the Semi-Supervised Normalized Cut (SSNCut) algorithm recently introduced in 
\cite{ChewC15}, which incorporates (soft) must-link and cannot-link constraints.

Tables \ref{table:ScribblesResult} and \ref{table:ScribblesResultBerkeley} and the plots in Figure \ref{fig:ExamplarResults} show the respective quantitative and the several qualitative segmentation results. Most of the results, reported on table \ref{table:ScribblesResult}, are reported by previous works \cite{LOOSECUTcorr15,MilCutCVPR14,iccvLempitsky09,PriMorCohCVPR2010,YanCaiZheLuoIP2010}.
We can see that the proposed CDS outperforms all the other approaches.

\begin{table}
	{
		\centering
		\begin{tabular}{l|r}
			
			Methods                                        & Error Rate \\ \hline
			BNC \cite{MajiNishVishMalCVPR11}                & 13.9 \\ \hline
			Graph Cut \cite{BoyJolICCV2001}                & 6.7 \\ \hline
			Lazy Snapping \cite{LiSunTanShuACM2004}        & 6.7 \\ \hline
			Geodesic Segmentation \cite{BaiSapIJCV2009}    & 6.8 \\ \hline
			Random Walker \cite{GraPAMI2006}               & 5.4 \\ \hline
			Transduction \cite{DucAudKerPonFloCVPR2008}    & 5.4 \\ \hline
			Geodesic Graph Cut \cite{PriMorCohCVPR2010}    & 4.8 \\ \hline
			Constrained Random Walker \cite{YanCaiZheLuoIP2010} & 4.1 \\ \hline
			CDS\_Self Tuning (Ours)               & \textbf{3.57} \\ \hline
			CDS\_Single Sigma (Ours)              & \textbf{3.80} \\ \hline
			CDS\_Best Sigma (Ours)                & 2.72 \\ \hline
		\end{tabular}
		\caption{\small Error rates of different scribble-based approaches on the Grab-Cut dataset.}
		\label{table:ScribblesResult}
	}
\end{table}

\begin{table}
	{
		\centering
		\begin{tabular}{l|r}
			Methods                          & Jaccard Index \\ \hline
			MILCut-Struct \cite{MilCutCVPR14}             & 84 \\ \hline
			MILCut-Graph \cite{MilCutCVPR14}              & 83 \\ \hline
			MILCut \cite{MilCutCVPR14}                    & 78 \\ \hline
			Graph Cut \cite{GrabCutRotherKB04}            & 77 \\ \hline
			Binary Partition Trees \cite{SalGarIP2000}    & 71 \\ \hline
			Interactive Graph Cut \cite{BoyJolICCV2001}   & 64 \\ \hline
			Seeded Region Growing \cite{AdaBisPAMI1994}   & 59 \\ \hline
			Simple Interactive O.E\cite{FriJanRojACM2005} & 63 \\ \hline
			CDS\_Self Tuning (Ours)                       & \textbf{93} \\ \hline
			CDS\_Single Sigma (Ours)                      & \textbf{93} \\ \hline
			CDS\_Best Sigma (Ours)                        & 95 \\ \hline
		\end{tabular}
		\caption{\small Jaccard Index of different approaches -- first 5 bounding-box-based -- on Berkeley dataset.}
		\label{table:ScribblesResultBerkeley}
	}
\end{table}

\textbf{Error-tolerant Scribble Based Segmentation.} This is a family of scribble-based approach, proposed by Bai et. al \cite{BaiWuCVPR2014}, which tolerates imperfect input scribbles thereby avoiding the assumption of accurate scribbles. 
We have done experiments using synthetic scribbles and compared the algorithm against recently proposed methods specifically designed to segment and extract the object of interest tolerating the user input errors \cite{BaiWuCVPR2014,LiuSunShuACM2009,OzaKemAydACM2012,SubParSolKauCGF2013}.

Our framework is adapted to this problem as follows. We give for our framework the foreground scribbles as constraint set and check those scribbled regions which include background scribbled regions as their members in the extracted dominant set. Collecting all those dominant sets which are free from background scribbled regions generates the object of interest.

\begin{figure*}
	\centering
	\includegraphics[width=1\linewidth,trim=0.15cm 10cm 7cm 0cm,clip ]{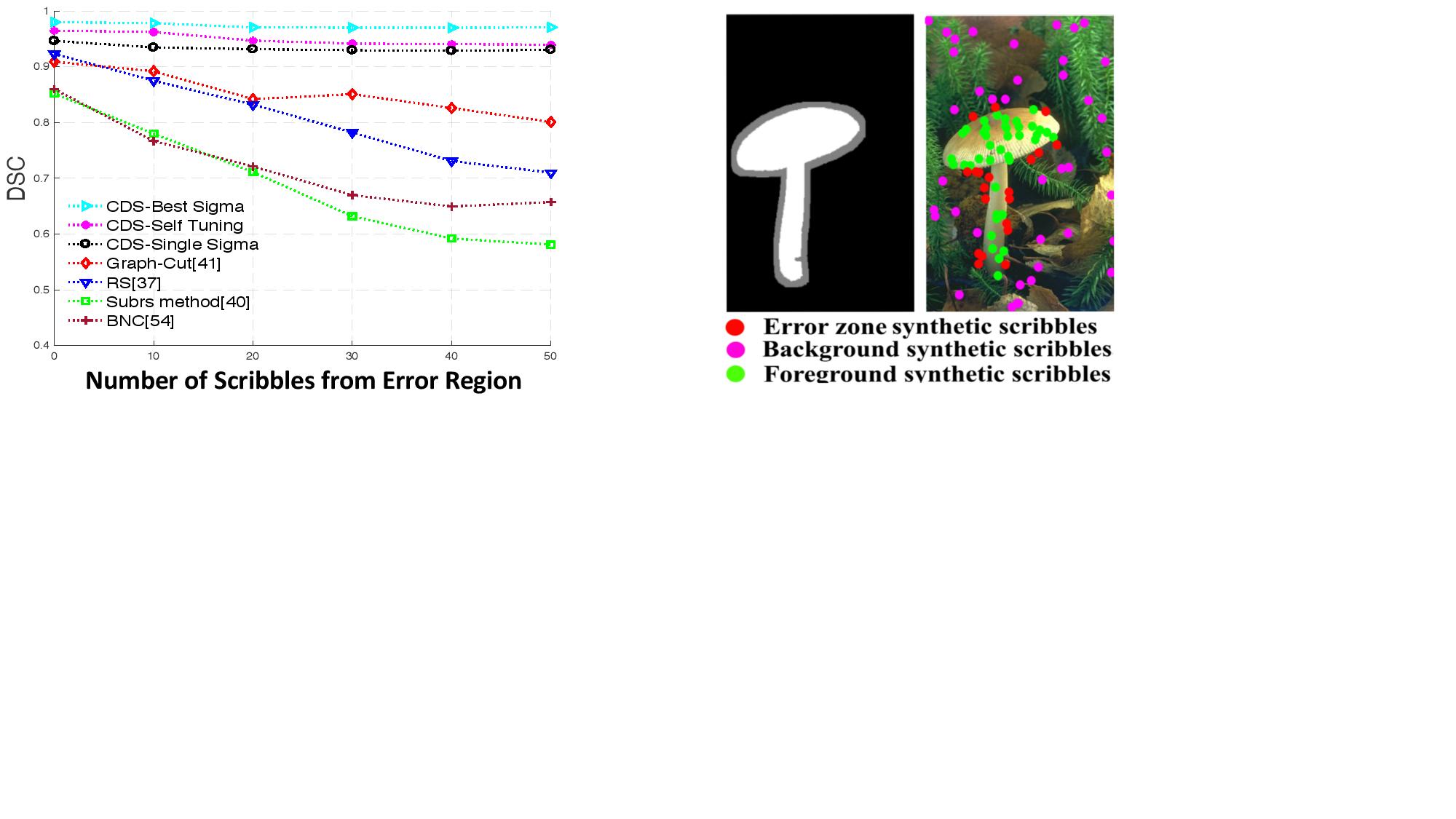}
	\caption{\small  \textbf{Left:} Performance of interactive segmentation algorithms, on Grab-Cut dataset, for different percentage of synthetic scribbles from the error region. \textbf{Right:} Synthetic scribbles and error region }
	\label{fig:SyntheticScribblesResult}
\end{figure*}

\textbf{Experiment using synthetic scribbles.}
Here, a procedure similar to the one used in \cite{SubParSolKauCGF2013} and \cite{BaiWuCVPR2014} has been followed.
First, 50 foreground pixels and 50 background pixels are
randomly selected based on ground truth (see Fig. \ref{fig:SyntheticScribblesResult}).
They are then assigned as foreground or background scribbles, respectively. Then
an error-zone for each image is defined as background pixels
that are less than a distance D from the foreground, in
which D is defined as 5 \%. We randomly select 0 to 50 pixels in the error zone and assign
them as foreground scribbles to simulate different degrees
of user input errors. We randomly select 0, 5, 10, 20, 30, 40, 50
erroneous sample pixels from error zone to simulate the error
percentage of 0\%, 10\%, 20\%, 40\%, 60\%, 80\%, 100\%
in the user input. It can be observed from figure \ref{fig:SyntheticScribblesResult} that our approach is not affected by the increase in the percentage of scribbles from error region.

\subsubsection{Segmentation using bounding boxes} \label{Sec.boundingBox}
The goal here is to segment the object of interest out from the background based on a given bounding box. The corresponding over-segments which contain the box label are taken as constraint set which guides the segmentation. The union of the extracted set is then considered as background while the union of other over-segments represent the object of interest.

\begin{figure*}[t]
	\centering
	\includegraphics[width=1\linewidth,trim=0cm 0cm 0cm 0cm,clip]{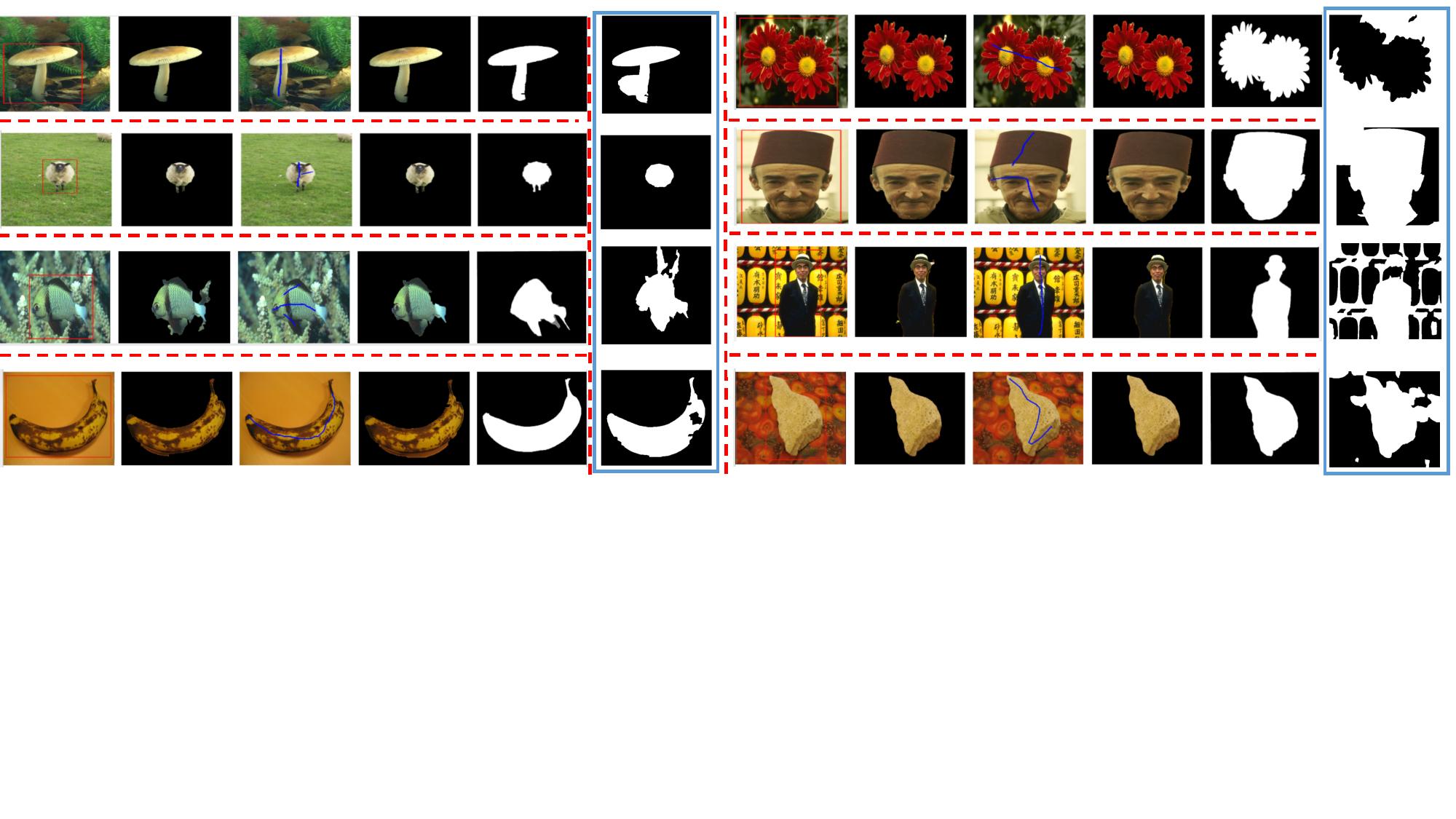}
	\caption{\small Exemple results of the interactive segmentation algorithm tested on Grab-Cut dataset. (In each block of the red dashed line) \textbf{Left:} Original image with bounding boxes of \cite{iccvLempitsky09}. \textbf{Middle left:} Result of the bounding box approach. \textbf{Middle:} Original image and scribbles (observe that the scribbles are only on the object of interest). \textbf{Middle right:} Results of the scribbled approach. \textbf{Right:} The ground truth. \textbf{Blue box:} Results of Semi-Supervised Normalized Cuts (SSNcut) \cite{ChewC15}.}
	\label{fig:ExamplarResults}
\end{figure*}

We provide quantitative comparison against several recent state-of-the-art interactive image segmentation methods which uses bounding box: LooseCut \cite{LOOSECUTcorr15}, GrabCut \cite{GrabCutRotherKB04}, OneCut \cite{OneCutICCV13}, MILCut \cite{MilCutCVPR14}, pPBC and \cite{TangECCV14}.
Table \ref{table:LoosCutApproach} and the pictures in Figure \ref{fig:ExamplarResults} show the respective error rates and the several qualitative segmentation results. Most of the results, reported on table \ref{table:LoosCutApproach}, are reported by previous works \cite{LOOSECUTcorr15,MilCutCVPR14,iccvLempitsky09,PriMorCohCVPR2010,YanCaiZheLuoIP2010}.

\textbf{Segmentation Using Loose Bounding Box.} This is a variant of the bounding box approach, proposed by Yu et.al \cite{LOOSECUTcorr15}, which avoids the dependency of algorithms on the tightness of the box enclosing the object of interest. The approach not only avoids the annotation burden but also allows the algorithm to use automatically detected bounding boxes which might not tightly encloses the foreground object. It has been shown, in \cite{LOOSECUTcorr15}, that the well-known GrabCut algorithm \cite{GrabCutRotherKB04} fails when the looseness of the box is increased. Our framework, like \cite{LOOSECUTcorr15}, is able to extract the object of interest in both tight and loose boxes. Our algorithm is tested against a series of bounding boxes with increased looseness. The bounding boxes of \cite{iccvLempitsky09} are used as boxes with 0\% looseness. A looseness $L$ (in percentage) means an increase in the area of the box against the baseline one. The looseness is increased, unless it reaches the image perimeter where the box is cropped, by dilating the box by a number of pixels, based on the percentage of the looseness, along the 4 directions: left, right, up, and down.

For the sake of comparison, we conduct the same experiments as in \cite{LOOSECUTcorr15}: 41 images out of the 50 GrabCut dataset \cite{GrabCutRotherKB04} are selected as the rest 9 images contain multiple objects while the ground truth is only annotated on a single object. As other objects, which are not marked as an object of interest in the ground truth,  may be covered when the looseness of the box increases, images of multiple objects are not applicable
for testing the loosely bounded boxes \cite{LOOSECUTcorr15}. Table \ref{table:LoosCutApproach} summarizes the results of different approaches using bounding box at different level of looseness. As can be observed from the table, our approach performs well compared to the others when the level of looseness gets increased. When the looseness $L=0$, \cite{MilCutCVPR14} outperforms all, but it is clear, from their definition of tight bounding box, that it is highly dependent on the tightness of the bounding box. It even shrinks the initially given bounding box by 5\% to ensure its tightness before the slices of the positive bag are collected. For looseness of $L=120$ we have similar result with LooseCut \cite{LOOSECUTcorr15} which is specifically designed for this purpose. For other values of $L$ our algorithm outperforms all the approaches.

\begin{table}[t]
	\centering
	\begin{tabular}
		{p{3.2cm}    | p{0.8cm} | p{0.8cm}| p{0.8cm}| p{0.8cm}}
		\hline\hline
		Methods      & $ L = 0\% $ & $ L = 120\% $ & $ L = 240\% $ & $ L = 600\% $  \\ \hline
		GrabCut \cite{GrabCutRotherKB04}      & 7.4 &  10.1 &  12.6     &  13.7   \\ \hline
		OneCut \cite{OneCutICCV13}            & 6.6 & 8.7  &  9.9      &  13.7  \\ \hline
		pPBC \cite{TangECCV14}                & 7.5  &  9.1  &  9.4      &  12.3  \\ \hline
		MilCut \cite{MilCutCVPR14}            & \textbf{3.6}&  -    &  -        &  -       \\ \hline
		LooseCut \cite{LOOSECUTcorr15}        & 7.9 & \textbf{5.8}  &  6.9      &  6.8     \\ \hline
		CDS\_Self Tuning (Ours)               & 7.54 &  6.78 &  \textbf{6.35}      &  7.17 \\ \hline
		CDS\_Single Sigma (Ours)              & 7.48 &  5.9  &  \textbf{6.32}      & \textbf{6.29} \\ \hline
		CDS\_Best Sigma (Ours)                & 6.0  &  4.4  &  4.2     &  4.9 \\ \hline
	\end{tabular}
	\caption{\small Error rates of different bounding-box approaches with different level of looseness as an input, on the Grab-Cut dataset. $ L = 0\% $ implies a baseline bounding box as those in \cite{iccvLempitsky09}}
	\label{table:LoosCutApproach}
\end{table}

\textbf{Complexity.} 
In practice, over-segmenting and extracting features may be treated as a pre-processing step which can be done
before the segmentation process. Given the affinity matrix, we used replicator dynamics (\ref{eqn:Replicator}) to exctract
constrained dominant sets. Its computational complexity per step is $O(N^2)$, with $N$ being the total number of nodes of the graph.
Given that our graphs are of moderate size (usually less than 200 nodes) the algorithm is fast and converges 
in fractions of a second, with a code written in Matlab and run on a core i5 6 GB of memory.
As for the pre-processing step, the original \textit{gPb-owt-ucm} segmentation algorithm was very slow to be used as a practical tools. Catanzaro et al. \cite{CatSuSunLeeMurKeuICCV2009} proposed a faster alternative, which reduce the runtime from 4 minutes to 1.8 seconds, reducing the  computational complexity and using parallelization which allow \textit{gPb} contour detector and \textit{gPb-owt-ucm} segmentation algorithm practical tools. For the purpose of our experiment we have used the Matlab implementation which takes around four minutes to converge, but in practice it is possible to give for our framework as an input, the GPU implementation \cite{CatSuSunLeeMurKeuICCV2009} which allows the convergence of the whole framework in around 4 seconds.

\begin{figure*}[t]
	\centering
	\includegraphics[width=1\linewidth ,trim=0cm 0cm 0cm 0cm,clip]{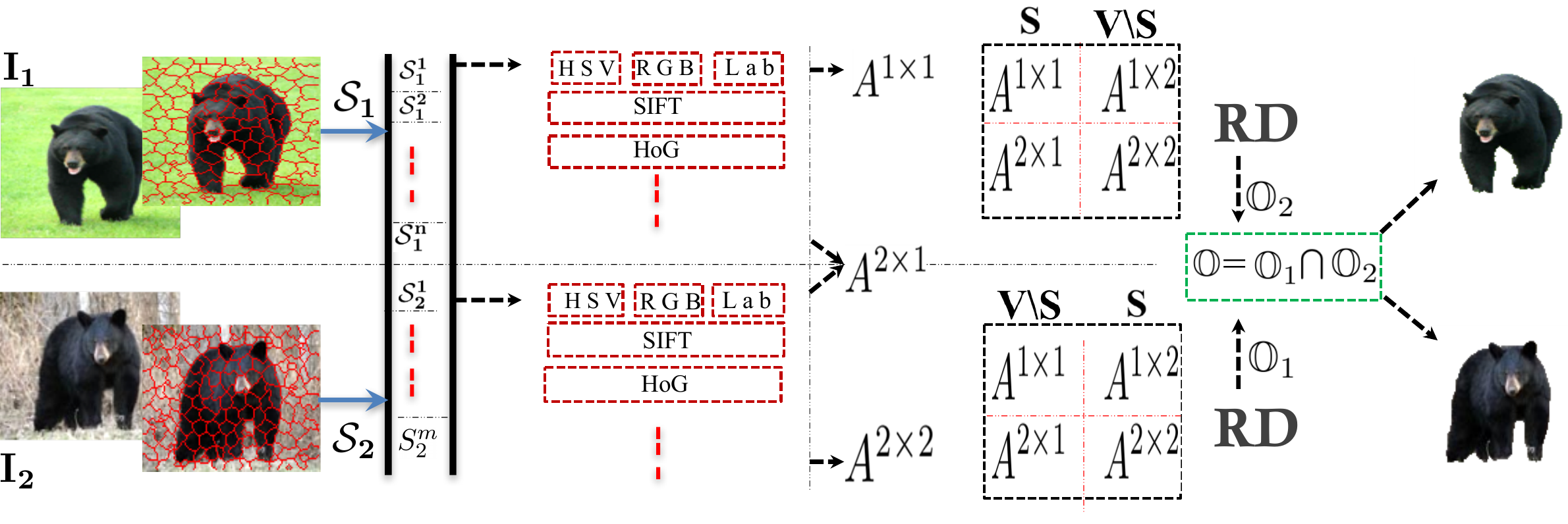}
	\caption{\small Overview of our unsupervised co-segmentation algorithm.}
	\label{fig:FrameworkCo-Seg}
\end{figure*}

\section{Application to co-segmentation}

\begin{figure}
	\centering
	\includegraphics[width=0.8\linewidth ,trim=0cm 3cm 0cm 2cm,clip]{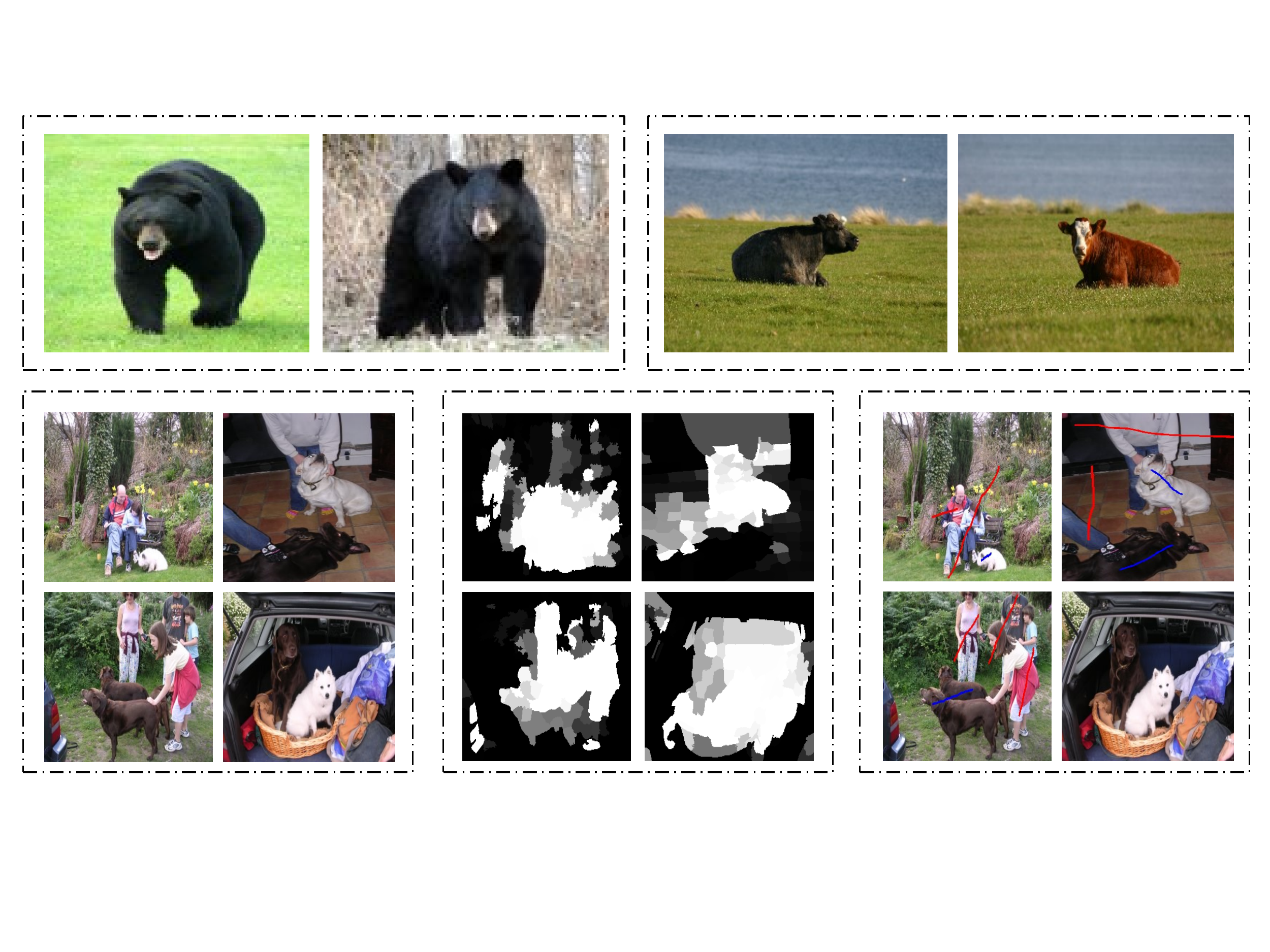}
	\caption{The challenges of co-segmentation. Example image pairs: \textbf{(top left)} similar foreground objects with significant variation in background, \textbf{(top right)} foreground objects with similar background. The \textbf{bottom} part shows why user interaction is important for some cases. The \textbf{bottom left} is the image, \textbf{bottom middle} shows the objectness score, and the \textbf{bottom right} shows the user label.}
	
	\label{fig:ImgPairExemplar}
\end{figure}

In this section, we describe the application of constrained dominant sets (CDS) to co-segmentation, both unsupervised and interactive. 
Among the difficulties that make this problem a challenging one, we mention the similarity among the different backgrounds and the similarity of object and background \cite{SarCarVlaCVPR2011} (see, e.g., the top row of Figure \ref{fig:ImgPairExemplar}). 
A measure of ``objectness'' has proven to be effective in dealing with such problems and improving the co-segmentation results \cite{SarCarVlaCVPR2011,AvikChaVelECCV16}. However, this measure alone is not enough especially when one aims to solve the problem using global pixel relations. One can see from Figure \ref{fig:ImgPairExemplar} (bottom) that the color of the cloth of the person, which of course is one of the objects, is similar to the color of the dog which makes systems that are based on objectness measure fail. Moreover the object may not be the one which we want to co-segment.

\begin{figure*}[t]
	\centering
	\includegraphics[width=1\linewidth ,trim=0cm 3cm 0cm 1cm,clip]{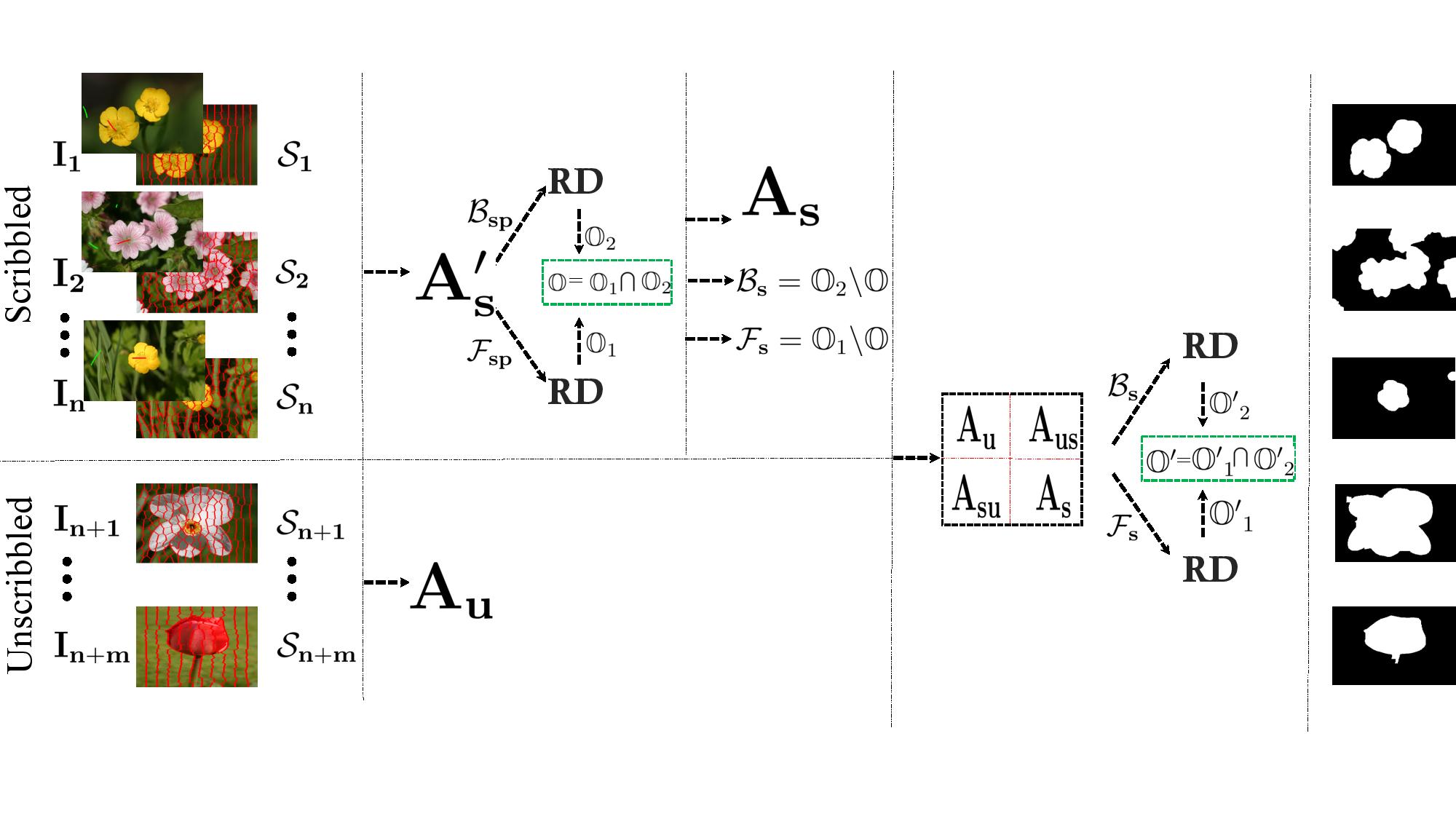}
	\caption{\small Overview of our interactive co-segmentation algorithm.} 
	\label{fig:interactiveco-segmentationworkflow}
\end{figure*}

Figures \ref{fig:FrameworkCo-Seg} and  \ref{fig:interactiveco-segmentationworkflow} show the pipeline of our unsupervised
and interactive co-segmentation algorithms, respectively. 
In figure \ref{fig:FrameworkCo-Seg}, $\vct{I}_{1}$ and $\vct{I}_{2}$ are the given pair of images while $\mathcal{S}_{1}$ and $\mathcal{S}_{2}$ represent the corresponding sets of superpixels. The affinity is built using the objectness score of the superpixels and using different handcrafted features extracted from the superpixels. The set of nodes $V$ is then divided into the constraint set ($S$) and the non-constraint set ($V\backslash S$). We run the CDS algorithm twice: first, setting the nodes of the graph that represent the first image as constraint set and $\mathbb{O}_2$ represents our output. Second we change the constraint set $S$ with nodes that come from the second image and $\mathbb{O}_1$ represents the output. The intersection $\mathbb{O}$ refines the two results and represents the final output of the proposed unsupervised co-segmentation approach.

Our interactive co-segmentation approach, as shown in Figure \ref{fig:interactiveco-segmentationworkflow}, needs user interaction which guides the segmentation process putting scribbles (only) on some of the images with ambiguous objects or background. $\vct{I}_{1},\vct{I}_{2},...\vct{I}_{n}$ are the scribbled images and $\vct{I}_{n+1}, ..., \vct{I}_{n+m}$ are unscribbled ones. The corresponding sets of superpixels are represented as $\mathcal{S}_{1},\mathcal{S}_{2},...\mathcal{S}_{n},...\mathcal{S}_{n+1},...\mathcal{S}_{n+m}$. $\vct{A'}_\vct{s}$ and $\vct{A}_\vct{u}$ are the affinity matrices built using handcrafted feature-based similarities among superpixels of scribbled and unscribbled images respectively. Moreover, the affinities incorporate the objectness score of each node of the graph. $\mathcal{B}_{\vct{s}\vct{p}}$ and $\mathcal{F}_{\vct{s}\vct{p}}$ are (respectively) the background and foreground superpixels based on the user provided information. The CDS algorithm is run twice over $\vct{A'}_\vct{s}$ using the two different user provided information as constraint sets which results outputs $\mathbb{O}_1$ and $\mathbb{O}_2$. The intersection of the two outputs, $\mathbb{O}$, help us get new foreground and background sets represented by $\mathcal{B}_\vct{s}$, $\mathcal{F}_\vct{s}$. Modifying the affinity $\vct{A'}_\vct{s}$, putting the similarities among elements of the two sets to zero, we get the new affinity $\vct{A}_\vct{s}$. We then build the biggest affinity which incorporates all images' superpixels. As our affinity is symmetric, $\vct{A}_{\vct{u}\vct{s}}$ and $\vct{A}_{\vct{s}\vct{u}}$ are equal and incorporates the similarities among the superpixels of the scribbled and unscribbled sets of images. Using the new background and foreground sets as two different constraint sets, we run CDS twice which results outputs $\mathbb{O'}_1$ and $\mathbb{O'}_2$ whose intersection ($\mathbb{O'}$) represents the final output.

\subsection{Graph representation and affinity matrix}
Given an image, we over-segment it to get its superpixels  $\mathcal{S}$, which are considered as vertices of a graph. We then extract different features from each of the superpixels. The first features we consider are obtained from the different color spaces: RGB, HSV and CIE Lab. Given the superpixels, say size of $n$, of an image $i$, $\mathcal{S}_i$, $\mathcal{F}_c^i$ is a matrix of size $n \times 9$ which is the mean of each of the channels of the three color spaces of pixels of the superpixel. The mean of the SIFT features extracted from the superpixel $\mathcal{F}_s^i$ is our second feature. The last feature which we have considered is the rotation invariant histogram of oriented gradient (HoG), $\mathcal{F}_h^i$. 

The dot product of the SIFT features  is considered as the SIFT similarity among the nodes, let us say the corresponding affinity matrix is $A_s$. Motivated by \cite{MorML2016}, the similarity among the nodes of image $i$ and image $j$ ($i\neq j$), based on color, is computed from their Euclidean distance $\mathcal{D}_c^{i\times j}$ as 

\[ A_c^{i\times j} = max(\mathcal{D}_c) - \mathcal{D}_c^{i\times j} + min(\mathcal{D}_c)\]

The HoG similarity among the nodes, $A_h^{i\times j}$, is computed in a similar way , as $A_c$, from the diffusion distance. All the similarities are then min max normalized.

We then construct the  $A_c^{i\times i}$, the similarities among superpixels of image $i$,  which only considers adjacent superpixels as follows. First, construct the dissimilarity graph using their Euclidean distance considering their average colors as weight. Then, compute the geodesic distance as the accumulated edge weights along their shortest path on the graph. Assuming the computed geodesic distance matrix is $\mathcal{D}_{geo}$, the weighted edge similarity of superpixel $p$ and superpixel $q$, say $e_{p,q}$, is computed as

\begin{equation}
e_{p,q} = 
\begin{cases}
0, \hspace*{1cm}\text{if\quad $p$ and $q$ are not adjacent}, & \\
max(\mathcal{D}_{geo}) - \mathcal{D}_{geo}(p,q)+ min(\mathcal{D}_{geo}),&\text{otherwise}
\end{cases}
\end{equation}
$A_h^{i\times i}$ for HoG is computed in a similar way while and $A_s^{i\times i}$ for SIFT is built by just keeping adjacent edge similarities.

Assuming we have $I$ images, the final affinity $A_\gamma$ ($\gamma$ can be $c$, $s$ or $h$ in the case of color, SIFT or HOG respectively) is built as
$$
A_\gamma = 
\begin{pmatrix} 
A_\gamma^{1 \times 1} & \cdots  & A_\gamma^{1 \times I} \\ 
\vdots & ~~ \ddots ~~ & \vdots &\\
A_\gamma^{I \times 1} & \cdots  & A_\gamma^{I \times I}\\ 
\end{pmatrix}
$$ 

As our goal is to segment common foreground objects out, we should consider how related backgrounds are eliminated. As shown in the example image pair of Figure \ref{fig:ImgPairExemplar} (top right), the two images have a related background to deal with it which otherwise would be included as part of the co-segmented objects.
To solve this problem we borrowed the idea from  \cite{WanShuYicJiaCVPR2014} which proposes a robust background
measure, called boundary connectivity. Given a superpixel $\mathcal{SP}_i$, it computes, based on the background measure, the backgroundness probability $\mathcal{P}_b^i$. We compute the probability of the superpixel being part of an object $\mathcal{P}_f^i$ as its additive inverse, $\mathcal{P}_f^i$ = 1 - $\mathcal{P}_b^i$. From the probability $\mathcal{P}_f$ we built a score affinity $A_m$ as 

\[A_m(i,j) = \mathcal{P}_f^i* \mathcal{P}_f^j\]

\subsection{Optimization}

We model the foreground object extraction problem as the optimization of the similarity values among all image superpixels.
The objective function is designed to assign the object region a positive membership score and the background region zero membership score, respectively. The optimal object region is then obtained by maximizing the objective function. Let the membership score of $N$ superpixels be $\{x_i\}_{i=1}^N$, the $(i,j)$ entry of a matrix $A_z$ is $z_{ij}$. 

Our objective function, combining all the aforementioned terms ($A_c$,$A_s$,$A_h$ and $A_m$), is thus defined, based on equation (\ref{eqn:parQP}), as:

\begin{equation}
\sum\limits_{i=1}^{N} \sum\limits_{j=1}^{N} \left\{ \frac{1}{2}\underbrace{ x_ix_jm_{ij}}_\text{objectness score} +    \frac{1}{6}x_ix_j\underbrace{\left(c_{ij} + s_{ij} + h_{ij} \right)}_\text{feature similarity} - \alpha x_ix_j \right\}
\end{equation}

The parameter $\alpha $ is fixed based on the (non-)constraint set of the nodes. For the case of unsupervised co-segmentation, the nodes of the pairs of images are set (interchangeably) as constraint set where the intersection of the corresponding results give us the final co-segmented objects.

In the interactive setting, every node $i$ (based on the information provided by the user) has three states: $i \in FGL $, ($i$ is labeled as foreground label), $i \in BGL$ ( $i$ is labeled as background label) or $i \in V\backslash ({FGL \cup BGL})$ ($i$ is unlabeled).
Hence, the affinity matrix $A=(a_{ij})$ is modified by setting $a_{ij}$ to zero if nodes $i$ and $j$ have different labels
(otherwise we keep the original value).

\begin{figure}
	\centering
	\includegraphics[width=0.70\linewidth,trim=0cm 0cm 0cm 0cm,clip]{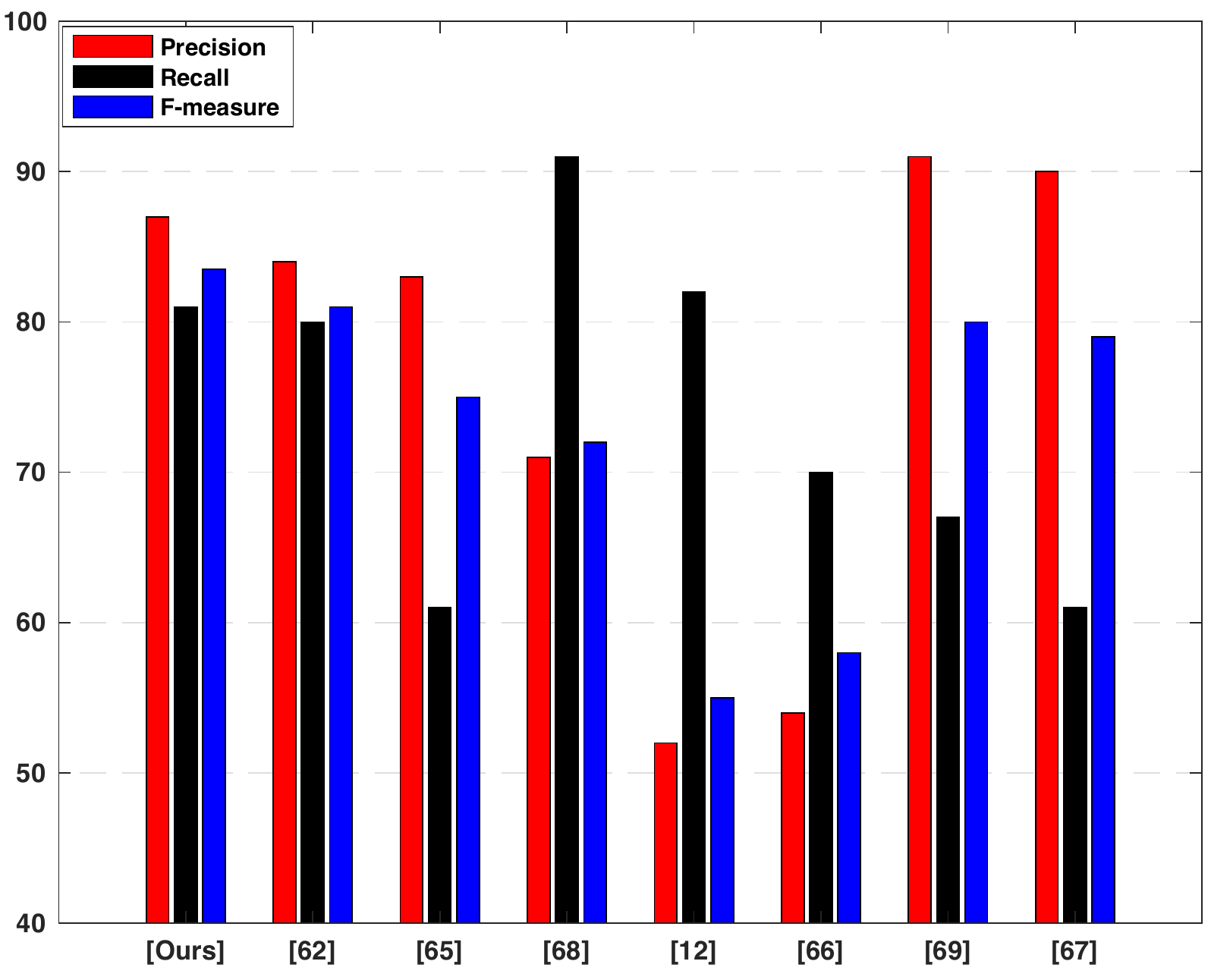}
	\caption{\small Precision, Recall and F-Measure of our unsupervised co-segmentation algorithm and other 
	state-of-the art approaches on the image pair dataset.}
	\label{fig:Imagepairesult}
\end{figure}

 \begin{figure*}
 	\centering
 	\includegraphics[width=0.85\linewidth]{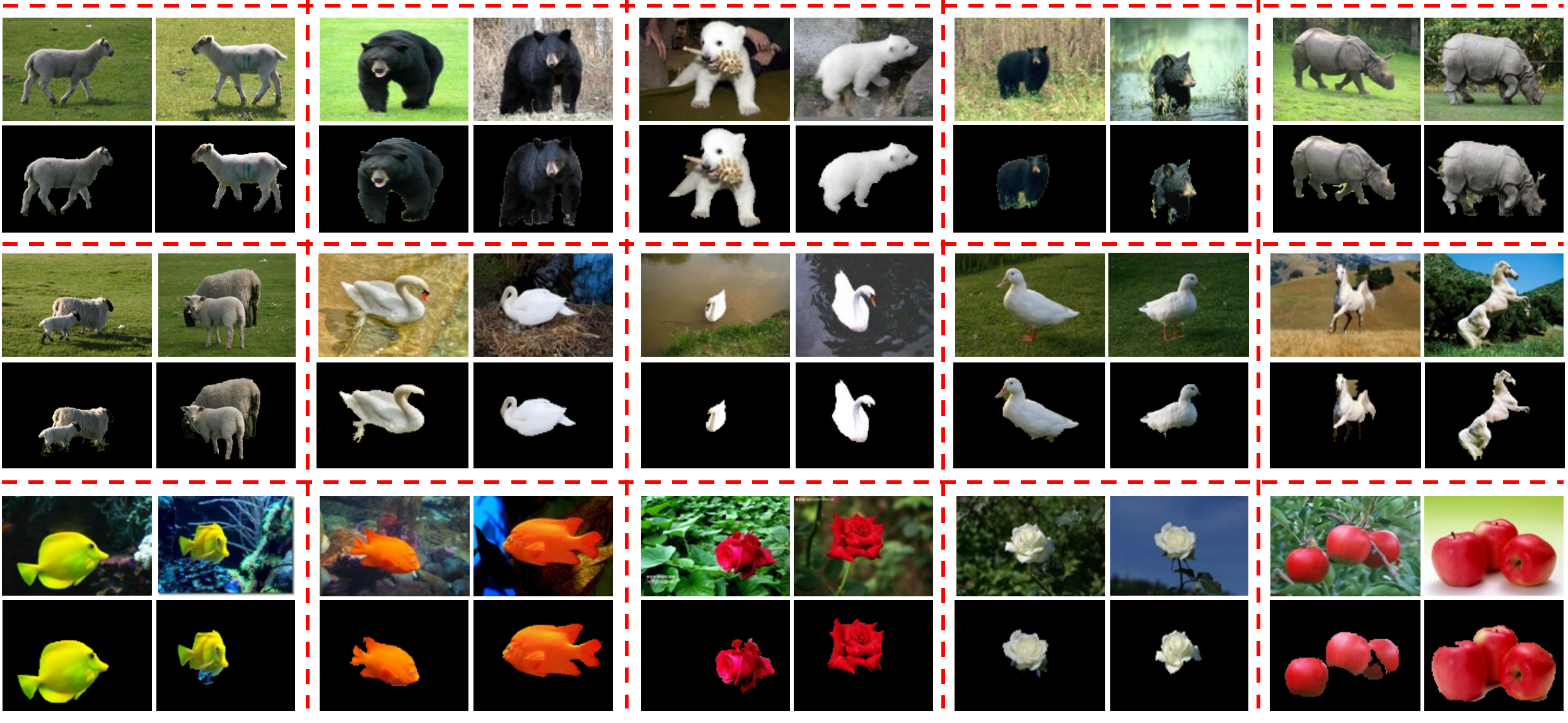}
 	\caption{\small Some qualitative results of our unsupervised method tested on the image pair dataset. \textbf{Upper row:} Original image \textbf{Lower row:} Result of the proposed unsupervised algorithm.}
 	\label{fig:ExamplarResultsCoSeg}
 \end{figure*}

\subsection{Experiments and results}

To evaluate the effectiveness of our approach, we conducted extensive experiments on standard benchmark datasets that are widely used to evaluate co-segmentation algorithms, namely, image pairs \cite{HonKinPAMI2011} and MSRC \cite{MicArmJohCeCVPR2013}. The image pairs dataset consists of 210 images (105 image pairs) of different animals, flowers, human objects, buses, etc. Each image pair contains one or more similar objects. Some of them are relatively simple and others include complex image pairs containing foreground objects with higher appearance variations or low-contrast objects with complex backgrounds. 

The MSRC dataset contains 14 categories with 418 images in total. We evaluated our interactive co-segmentation algorithm on nine selected object classes of MSRC dataset (bird, car, cat, chair, cow, dog, flower, house, sheep), which  contains 25\texttildelow30 images per class. We put foreground and background scribbles on 15\texttildelow20 images per class. Each image was over-segmented to 78\texttildelow83 SLIC superpixels using the VLFeat toolbox. 

In order to directly compare the performance of our algorithm with state-of-the-art approaches, in the experiments reported here we used precision, recall and F-measure, which were computed based on the output mask and the ground-truth segmentation. Precision is calculated as the ratio of correctly detected objects to the number of detected object pixels, while recall is the ratio of correctly detected object pixels to the number of ground truth pixels. The F-measure is computed as customary as
$$
F_\beta = \frac{(1+\beta^2) \times Precision \times Recall}{\beta^2 \times Precision + Recall}
$$
where we set $\beta^2 = 0.3$ as in \cite{HonKinPAMI2011,HuaXiaZhuTIP2013,AvikChaVelECCV16}.

\begin{figure}
	\centering
	\includegraphics[width=1\linewidth ,trim=0cm 0cm 0cm 0cm,clip]{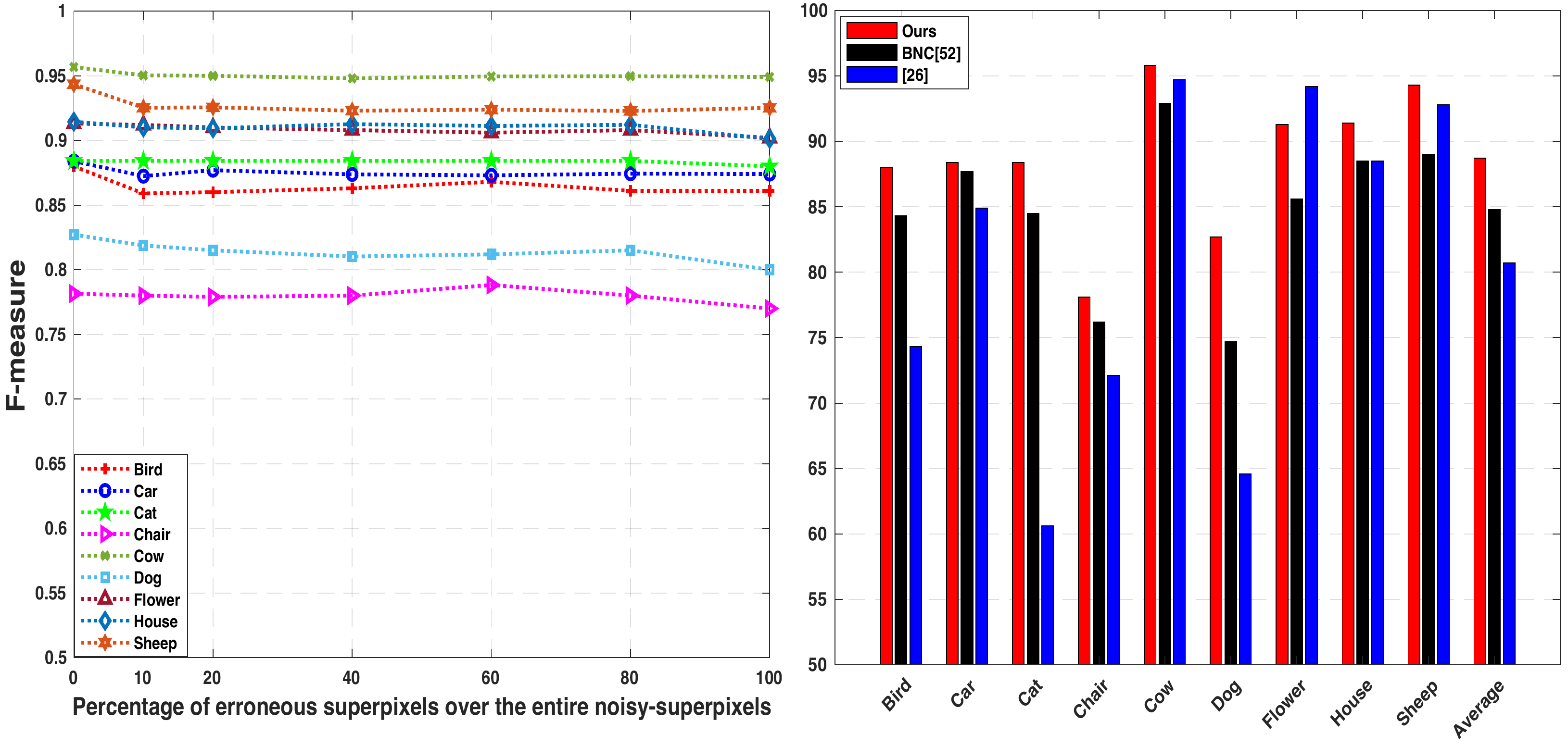}
	\caption{\textbf{Left:} Performance of our interactive image co-segmentation framework with different percentage of erroneous superpixels. \textbf{Right:} F-Measure based performance comparison of our interactive co-segmentation method with state-of-the-art methods on MSRC dataset.}
	\label{fig:MSRC_Result}
\end{figure}

We have applied Biased Normalized Cut (BNC) \cite{MajiNishVishMalCVPR11}
on co-segmentation problem on the MSRC dataset by using the same similarity matrix we used to test our method, and the comparison result of each object class is shown in Figure \ref{fig:MSRC_Result}. As can be seen, our method significantly surpasses BNC and  \cite{XinJiaLinMinTIP2015} in average F-measure. Furthermore, we have tested our interactive co-segmentation method, BNC and  \cite{XinJiaLinMinTIP2015} on image pairs dataset by putting scribbles on one of the two images. As can be observed from Table \ref{table:ImagePair}, our algorithm substantially outperforms BNC and \cite{XinJiaLinMinTIP2015} in precision and F-measure
(the recall score being comparable among the three competing algorithms). 

We have also examined our unsupervised co-segmentation algorithm by using image pairs dataset, the barplot in Figure \ref{fig:Imagepairesult} shows the quantitative result of our algorithm comparing to the state-of-the-art methods \cite{AvikChaVelECCV16,ChWoJaChaCVPR2015,XiaZhiqBaTIP2014}. As shown here, our algorithm achieves the best F-measure comparing to all other state-of-the-art methods. The qualitative performance of our unsupervised algorithm is shown in Figure \ref{fig:ExamplarResultsCoSeg} on some example images taken from image pairs dataset. As can be seen, our approach can effectively detect and segment the common object of the given pair of images.

\begin{table}[bth]
	
	\begin{center}
		\begin{tabular}{lllllllllllllll} 
			\hline\noalign{\smallskip}
			Metrics & $ Precision$  & $ Recall$ & $F-measure$ \\
			\noalign{\smallskip}
			\hline
			\noalign{\smallskip}
			\cite{XinJiaLinMinTIP2015}   & 0.5818 &  0.8239 &  0.5971 \\    
			BNC & 0.6421 & \textbf{0.8512} &  0.6564\\ 
			Ours & \textbf{0.7076} & 0.8208 &  \textbf{0.7140}\\ 	\hline	
		\end{tabular}
	\end{center}
	\caption{\small Results of our interactive co-segmentation method on Image pair dataset putting scribbles on one of the image pairs.}
	\label{table:ImagePair}
\end{table}

Finally, to assess the robustness of our interactive co-segmentation algorithm we conducted the following experiment on 
the MSRC dataset. We first generated random noise-scribbles by flipping 10\texttildelow20 superpixel labels from foreground to background and vice versa. We then randomly selected from $0\%,$ to $100\%$ erroneous superpixels from the noise-scribbles and ran our algorithm. As can be observed from Figure \ref{fig:MSRC_Result}(left) our algorithm performs consistently well, thereby confirming the behavior observed with error-tolerant scribble-based segmentation (Section 5.1.1), and 
previous experimental findings with dominant sets \cite{PavPel07,RotPelPAMI2013,RotPel17}.

\section{Summary}

In this chapter, we have introduced the notion of a {\em constrained dominant set} and have demonstrated its applicability
to problems such as interactive image segmentation and co-segmentation (in both the unsupervised and the interactive flavor).
In our perspective, these can be thought of as ``constrained'' segmentation problems involving an external source of information (being it, for example, a user annotation or a collection of related images to segment jointly) which somehow drives the whole segmentation process. 

The approach is based on some properties of a family of quadratic optimization problems related to dominant sets which show that, by properly selecting a regularization parameter that controls the structure of the underlying function, we are able to ``force'' all solutions to contain the constraint elements. 
The proposed method is flexible and is capable of dealing with various forms of constraints and input modalities, such as 
scribbles and bounding boxes, in the case of interactive segmentation. 
Extensive experiments on benchmark datasets have shown that our approach considerably improves the state-of-the-art results on the problems addressed. This provides evidence that constrained dominant sets hold promise as a powerful and principled framework to address a large class of computer vision problems formulable in terms of constrained grouping. Indeed, we mention that they are already being used in problems such as content-based image retrieval \cite{ZemAlePalICPR2016}, multi-target tracking \cite{TesfayeZPPS17} and image geo-localization \cite{EyaYonHarAndMarMubPAMI}.

The kind of constraints we dealt with in this thesis might be called first-order, or unary, positive constraints, as they
refer to a situation whereby one wants to {\em include} one or more given vertices into the final cluster solution. 
Of course, other types of constraints can be (and have indeed been) considered when doing clustering.
Using the terminology introduced above, for example, one might want to enforce (first-order) negative constraints, according to which certain vertices have to be {\em excluded} from the extracted cluster. This situation can easily be addressed within our framework by simply setting the initial values of the replicator dynamics corresponding to the to-be-excluded vertices to zero.
Note that by combining unary (negative and positive) constraints, and by employing the simple peel-off strategy described in \cite{PavPel07} to extract multiple clusters, it would be straightforward to generalize the proposed framework to 
multi-cluster versions of the problem involving multi-label seeds.
Second-order constraints, better known in the literature as must-link and cannot-link constraints \cite{YuShi04,Kul+09,Eri+11,ChewC15}, can also be easily incorporated. For example, a cannot-link constraint which prescribes that vertices $i$ and $j$ should not be part of the same cluster, can be enforced by setting $a_{ij}=0$ in the affinity matrix. In fact, a result proven in \cite{Alb+09} shows that, by doing so, no dominant set can contain both vertices. Similarly, must-link constraints might be enforced by setting $a_{ij}$ to a sufficiently large value, e.g., the maximum entry in the affinity matrix (see \cite{Kam+03} for a similar idea).
 
The reason why in this work we focused primarily on first-order positive constraints is that, despite their simplicity, they allow us to address in a unified manner various well-known segmentation settings which have been traditionally treated separately in the literature. Of course, the combination of various forms of pairwise or higher-order constraints might give rise to a more general and flexible segmentation framework, and it is our plan to investigate these ideas in our future work.


\newpage
\chapter{Constrained Dominant Sets for Image Retrieval}

\section{Multi-features Fusion Using Constrained Dominant Sets for Image Retrieval}

\label{chap:intro}

Aggregating different image features for image retrieval has recently shown its effectiveness. While highly effective, though, the question of how to uplift the impact of the best features for a specific query image persists as an open computer vision problem. In this paper, we propose a computationally efficient approach to fuse several hand-crafted and deep features, based on the probabilistic distribution of a given membership score of a constrained cluster in an unsupervised manner. First, we introduce an incremental nearest neighbor (NN) selection method, whereby we dynamically select k-NN to the query. We then build several graphs from the obtained NN sets and employ constrained dominant sets (CDS) on each graph G to assign edge weights which consider the intrinsic manifold structure of the graph, and detect false matches to the query. Finally, we elaborate the computation of feature positive-impact weight (PIW) based on the dispersive degree of the characteristics vector. To this end, we exploit the entropy of a cluster membership-score distribution. In addition, the final NN set bypasses a heuristic voting scheme. Experiments on several retrieval benchmark datasets show that our method can improve the state-of-the-art result.

\subsection{Introduction}

The goal of semantic image search, or content-based image retrieval (CBIR), is to search for a query image from a given image dataset. This is done by computing image similarities based on low-level image features, such as color, texture, shape and spatial relationship of images. Variation of images in illumination, rotation, and orientation has remained a major challenge for CBIR. Scale-invariant feature transform (SIFT) \cite{Lowe04} based local feature such as Bag of words (BOW) \cite{SivicZICCV03}, \cite{JainBJG12}, \cite{YangNXLZP12}, has served as a backbone for most image retrieval processes. Nonetheless, due to the inefficiency of using only a local feature to describe the content of an image, local-global feature fusion has recently been introduced.

\begin{figure*}[t]

	\includegraphics[width=1\linewidth ,trim=0cm 3.5cm 0cm 0cm,clip]{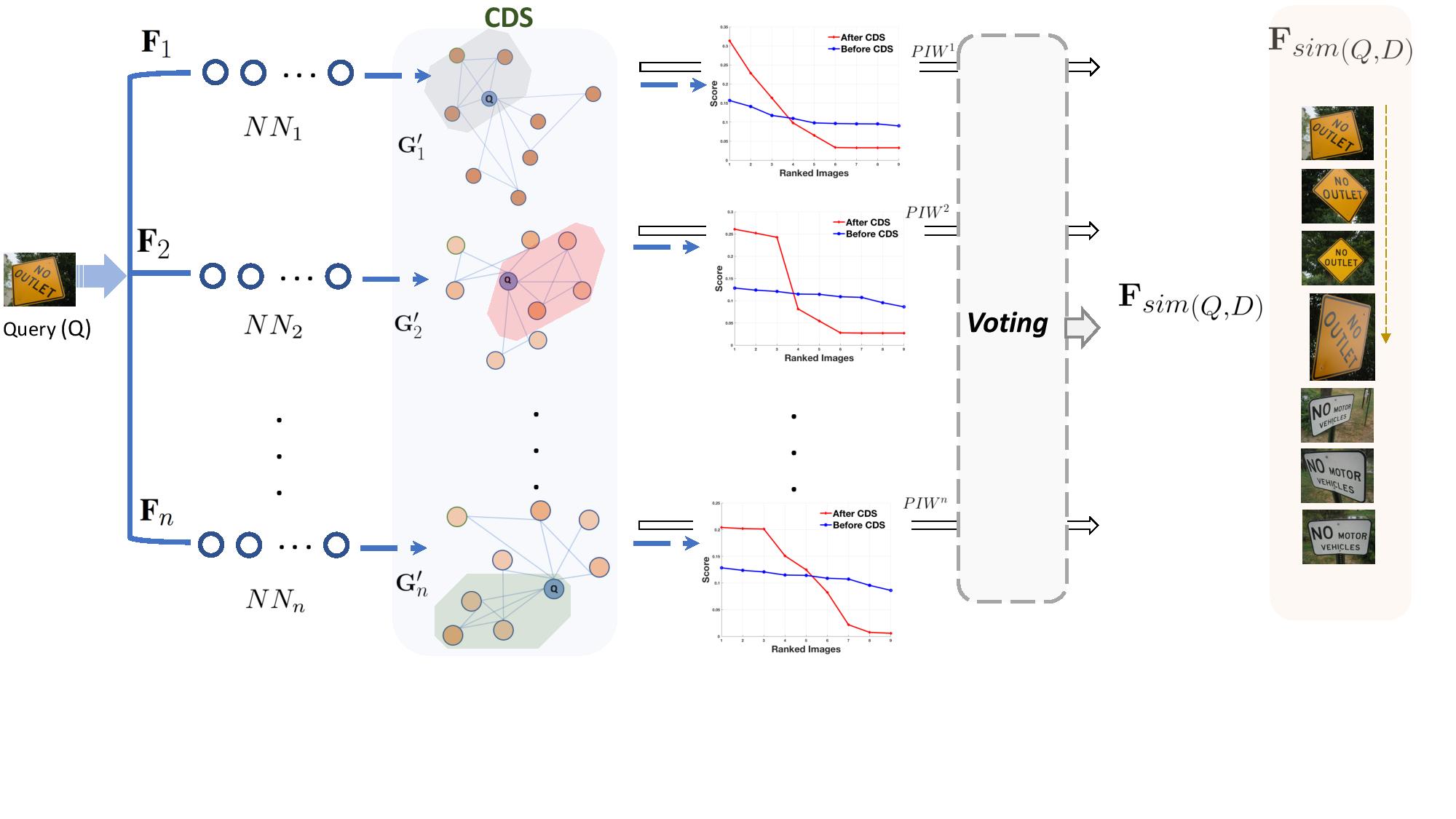}
	
	\caption{ Overview of the proposed image retrieval framework. Based on the given features, $F_1, F_2, . . . F_n,$ we first incrementally collect the $NN's$ to the query $Q,$ denoted as $NN_1, NN_2, . . . NN_n.$ Next, for each $NN$ we build the corresponding graph $G'_1, G'_2, . . . G'_n,$ and then, we apply $CDS$ on each graph to learn the $PIW$ of each feature, $PIW_1, PIW_2, . . . PIW_n,$ in the subsequent plot, the blue and red curves depict the ranked score of NN's before and after the application of CDS,  respectively. Following, the final candidates, which come from each feature, pass through a voting scheme. Finally, using the obtained votes and PIW's we compute the final similarity, $F_{sim} (Q, D)$, between the query and the dataset images by equ. \ref{equ:FinSim} .}
	\label{fig:Workflow}
\end{figure*}

Multi-feature based CBIR attacks the CBIR problem by introducing an approach which utilizes multiple low-level visual features of an image. Intuitively, if the to-be-fused feature works well by itself, it is expected that its aggregation with other features will improve the accuracy of the retrieval. Nevertheless, it is quite hard to learn in advance the effectiveness of the to-be-fused features for a specific query image. Different methods have recently been proposed to tackle this problem \cite{YangMD15}, \cite{ZhaYanCorYuMetPAMI2015}, \cite{ZheWanTiaHeLiuTiaCVPR2015}. Zhang et al. \cite{ZhaYanCorYuMetPAMI2015} developed a graph-based query specific fusion method, whereby local and global rank lists are merged with equal weight by conducting a link analysis on a fused graph.
Zheng et al. \cite{ZheWanTiaHeLiuTiaCVPR2015} proposed a score level fusion model called Query Adaptive Late Fusion (QALF) \cite{ZheWanTiaHeLiuTiaCVPR2015}, in which, by approximating a score curve tail with a  reference collected on irrelevant data, they able to estimate the effectiveness
of a feature as negatively related to the area under the normalized curve. Yang \et\cite{YangMD15} used a mixture Markov model to combine given graphs into one. Unlike \cite{ZhaYanCorYuMetPAMI2015} where graphs are equally weighted, \cite{YangMD15} proposed a method to compute a weight which quantifies the usefulness of the given graph based on a naive Bayesian formulation, which depends only on the statistics of image similarity scores. 

However, existing multi-feature fusion methods have different drawbacks. For instance, \cite{ZheWanTiaHeLiuTiaCVPR2015}, \cite{ZhaYanCorYuMetPAMI2015}, \cite{DenJiLiTaGaICCV2013}, \cite{ZhangYWLT13} heavily rely on a pre-calculated and offline stored data, which turns out to be computationally expensive when new images are constantly added to the dataset. On the other hand, Ensemble Diffusion (ED)\cite{Bai17} requires $O(n^3)$ to perform a similarity diffusion. In addition to that, its feature-weight computation approach is not a query specific. 

Inspired by \cite{ZheWanTiaHeLiuTiaCVPR2015}, in this work we present a novel and simple CBIR method based on a recently introduced constrained cluster notion. Our approach presents two main advantages. Firstly, compared to the state of the art methods, it can robustly quantify the effectiveness of features for a specific query, without any supervision. Secondly, by diffusing the pairwise similarity between the nearest neighbors, our model can easily avoid the inclusion of false positive matches in the final shortlist. Towards this end, we first dynamically collect the nearest neighbors to the query, therefore, for each feature, we will have a different number of NNs. Subsequently, we set up the problem as finding a cluster from the obtained NNs, which is constrained to contain the given query image. To this end, we employ a graph-theoretic method called constrained dominant sets \cite{ZemeneP16}. Here is our assumption: if the nearest neighbor to the query image is a false match, after the application of CDS its membership score to the resulting constrained cluster should be less than the fixed threshold $\zeta,$ which leads us to detect and exclude the outliers. Furthermore, we introduce the application of entropy to quantify the effectiveness of the given features based on the obtained membership score. In contrast to \cite{ZheWanTiaHeLiuTiaCVPR2015}, our method does not need any reference or external information to learn a query specific feature-weight. Fig. \ref{fig:Workflow} shows the pipline of the proposed method.

In particular, we make the following contributions. 1) Compared to the previous work \cite{ZhaYanCorYuMetPAMI2015}, \cite{ZheWanTiaHeLiuTiaCVPR2015}, we propose a simple but efficient entropy-based feature effectiveness weighting system; in addition to that, we demonstrate an effective way of outlier or false nearest neighbor detection method. 2) Most importantly, our proposed model is a generic approach, which can be adapted to distinct computer vision problems, such as object detection and person re-identification. 3) We show that our unsupervised graph fusion model easily alleviates the asymmetry neighborhood problem. 

This chapter is structured as follows. In section 2 we briefly survey literature relevant to our problem, followed by technical details of the proposed approach in Sec. 3. And, in Sec. 4 we show the performance of our framework on different benchmark datasets.


\subsection{Related Work}

CBIR has become a well-established research topic in the computer vision community. The introduction of SIFT feature plays a vital role in the application of BOW model on the image retrieval problem. Particularly, its robustness in dealing with the variation of images in scale, translation, and rotation provide a significant improvement in the accuracy of similar image search. Sivic et al. \cite{SivicZICCV03} first proposed BOW-based image retrieval method by using SIFT, in that, local features of an image are quantized to visual words. Since then, CBIR has made a remarkable progress by incorporating k-reciprocal neighbor \cite{QinGamBosQuaGooCVPR2011}, query expansion \cite{ChumMPM11}, \cite{QinGamBosQuaGooCVPR2011}, \cite{ToliasJ14}, large visual codebook \cite{PhilbinCISZ07}, \cite{AvrithisK12}, diffusion process \cite{YangMD15} \cite{ZemeneAP16}, \cite{LeulThesis} and spacial verification \cite{PhilbinCISZ07}.  Furthermore, several methods, which consider a compact representation of an image to decrease the memory requirement and boost the search efficiency have been proposed. Jegou et al. \cite{JegouPDSPSPAMI12} developed a Vector of Locally Aggregated Descriptor(VLAD), whereby the residuals belonging to each of the codewords are accumulated.

While SIFT-based local features have considerably improved the result of image search, it does not leverage the discriminative information encoded in the global feature of an image, for instance, the color feature yields a better representation for smooth images. This motivates the introduction of multiple feature fusion for image retrieval. In \cite{ZhaYanCorYuMetPAMI2015}, a graph-based query specific fusion model has been proposed, in which multiple graphs are combined and re-ranked by conducting a link analysis on a fused graph. Following, \cite{YangMD15} developed a re-ranking algorithm by fusing multi-feature information, whereby they apply a locally constrained diffusion process (LCDP) on the localized NNs to obtain a consistent similarity score.

Although the aggregation of handcrafted local and global features has shown promising results, the advent of a seminal work by A.Krizhevsky \et \cite{KrizhevskySH12} in 2012 changed the focus of the computer vision community. Since then, convolutional neural network (CNN) feature has been used as a main holistic cue in different computer vision problems, including CBIR. Despite its significant improvement on the result of image retrieval, CNN feature still can not endow the demanded accuracy on different benchmark retrieval datasets, especially without the use of fine-tuning. Thus, aggregating graphs which are built from a hand-engineered and CNN-based image features has shown improvement in the accuracy of the retrieval \cite{SivicZICCV03}, \cite{JegouPDSPSPAMI12}, \cite{PerronninDCVPR07}, \cite{JegouZCVPR14}, \cite{DoTCCVPR15}, \cite{KalantidisMO16}.

In addition to that, Yang \et \cite{YangMD15} applied a diffusion process to understand the intrinsic manifold structure of the fused graph. Despite a significant improvement on the result, employing the diffusion process on the final (fused) graph restricts the use of the information which is encoded in the pairwise similarity of the individual graph. Instead, our proposed framework applies CDS on each graph which is built from the corresponding feature. Thus, we are able to propagate the pairwise similarity score throughout the graph. Thereby, we exploit the underutilized pairwise similarity information of each feature and alleviate the negative impact of the inherent asymmetry of a neighborhood.

\subsection{Proposed Method}

\subsubsection{Incremental NN Selection}

In this subsection, we show an incremental nearest neighbor collection method to the given query image. We start with an intuitive clustering concept that similar nodes with common features should have an approximate score distribution, while outliers, or nodes which do not belong to a similar semantic class, have different score values. Accordingly, we propose a technique to search for the transition point where our algorithm starts including the outlier nodes. To this end, we examine how distinctive two subsequent nodes are in a ranked list of neighbors. Thus, we define a criterion called neighbors proximity coefficient$(NPC),$ which is defined as the ratio of two consecutive NNs in the given ranked list. Therefore, images are added only if the specified criterion is met, which is designed in such a way that only images that are very likely to be similar to the query image are added. Thereby, we are able to decrease the number of false matches to the query in the k-nearest neighbors set. 

Given an initial ranked list $R$. And then, we define top-k nearest neighbors (kNN) to query $Q$ as

\begin{equation}
	kNN(q, k) = 
	\begin{cases}
		\textrm{Add} \: n_i & if  \quad   \frac{Sim(q,n_{i+1})}{Sim(q,n_{i})} > $NPC$    \\
		0 & otherwize  
	\end{cases}
\end{equation} 

where $ |kNN(q, k)| = k,$ and $|.|$ represents the cardinality of a set.

\begin{equation}
kNN(q, k)= \{n_1, n_2, . . . n_k\},\quad where \; kNN(q,k) \subseteq R
\end{equation}

\subsubsection{Graph Construction}\label{PIWusingCDS}

Different features, $F = F_1, F_2 . . . F_n,$ are extracted from images in the dataset D and the query image $Q$, where each feature encodes discriminative information of the given image in different aspects. We then compute the distance between the given images based on a distance metric function $d'(I_i,I_j),$ where $I_i$ and $I_j$ denote the given feature vector extracted from image $i$ and $j$ respectively. Following, we compute symmetric affinity matrices $A_1'$, $A_2'$, . . . $A_n'$  from each distance matrix $D_i$ using a similarity function $S(D_i)$. We then apply minimax normalization on each similarity matrix as: $	A_i= \frac{V_\alpha^{ij} - min(V_\alpha)}{max(V_\alpha) - min(V_\alpha)}$, where $V_\alpha$ is a column vector taken from matrix $A'_i$, which comprises the pairwise similarity score between a given image $V_\alpha^i$ and images in the dataset $V^j,$ which is denoted as $V_\alpha^{ij}$. Next, we build undirected edge-weighted graphs with no self-loops $G_1, G_2 . . . G_n$ from the affinity matrices $A_1, A_2, . . . A_n,$ respectively. Each graph $G_n$  is defined as $Gn = (V_n, E_n, w_n),$ where $V_n = { 1, . . . ,n }$ is vertex set, $E_n \subseteq V_n \times V_n$ is the edge set, and $w_n: E \longrightarrow {\rm I\!R}_+^*$  is the (positive) weight function. Vertices in G correspond to the given images, edges represent neighborhood relationships, and edge-weights reflect similarity between pairs of linked vertices.

\subsubsection{PIW Using Entropy of CDS}

Since the nearest neighbor selection method heavily relies on the initial pairwise similarity, it is possible that the NN set easily includes false matches to the given query. This usually happens due to the lack of technics which consider the underlying structure of the data manifold,  especially the inherent asymmetry of a neighborhood is a major shortcoming of such systems. For instance, although $Sim(n_i, q) = Sim(q, n_i),$ the nearest neighbor relationship between query $Q$ and image $n_i$ may not be symmetric, which implies that  $m_i \in kNN(q, k)$ but $m_i \notin kNN(n_i, k).$
As demonstrated in the past retrieval works, the k-reciprocal neighbors \cite{QinGamBosQuaGooCVPR2011} and similarity diffusion process \cite{IscenTAFC17} have been vastly taken as the optimal options to tackle this issue. However, the existing methods are not computationally efficient. In this work, we remedy the existing limitations using an unsupervised constrained clustering algorithm whereby we exploit the pairwise similarity to find a cohesive cluster which incorporates the specified query.


\subsubsection{Constrained Clustering for Coherent Neighbor Selection}\label{kRN}

\begin{figure*}[t]
	
	\centering
	
	\includegraphics[width=1\linewidth ,trim=0cm 6.8cm 0cm 0cm,clip]{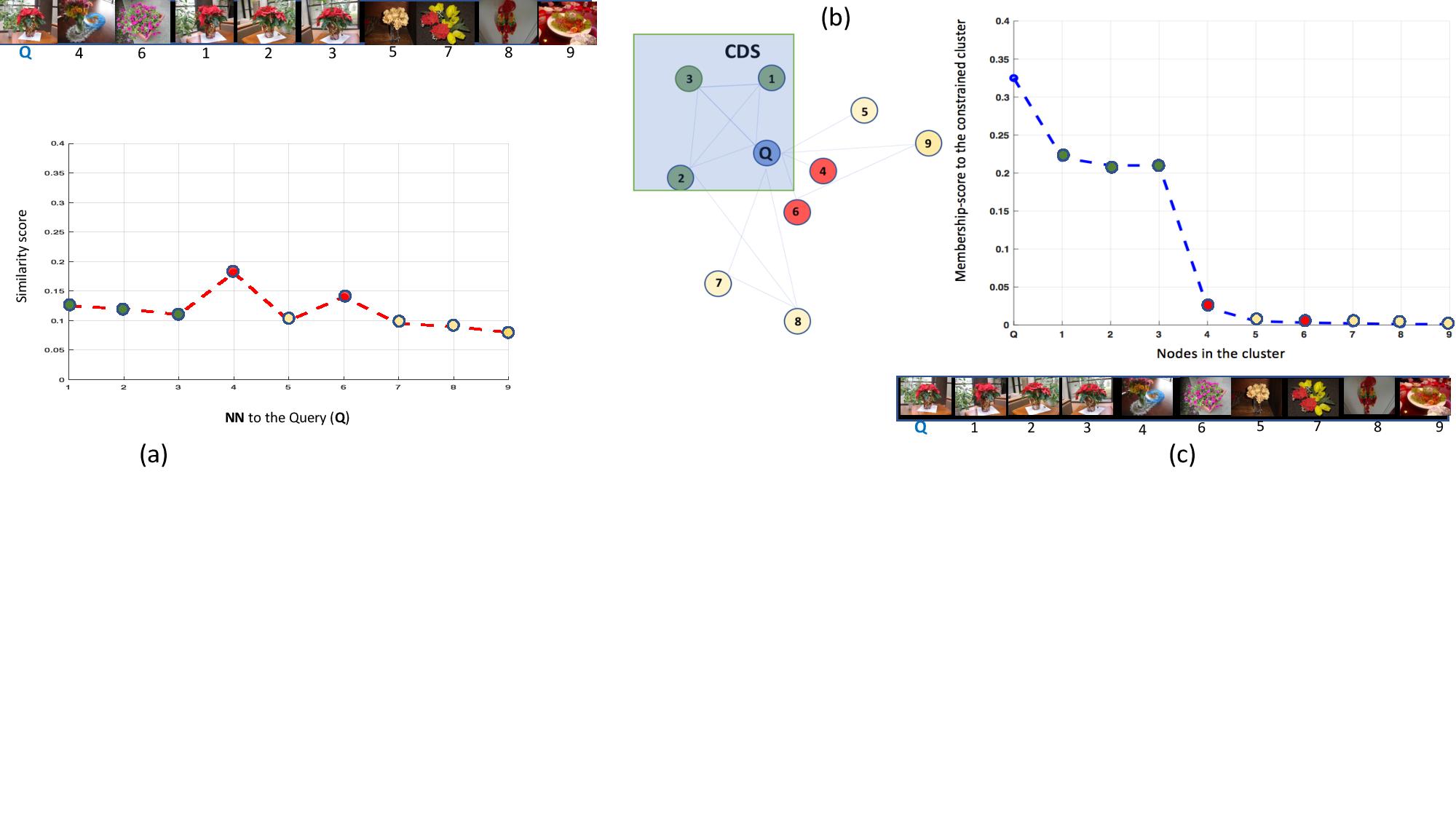}
	
	\caption{(a) Initial score distribution of the top k nearest neighbors to the query Q, green and red points denote the false-negative and false-posetive NNs. (b) Graph $G'$, built from the initial pairwise similarity of the k-nearest neighbor set. And the blue box contains the CDS nodes which are obtained by running CDS on graph $G'.$ (c) The resulting constrained dominant set membership-score distribution.}
	\label{fig:CharacvecExam}
\end{figure*}

Towards collecting true matches to the query image, we employ an unsupervised clustering algorithm on the top of the previous steps. Our hypothesis to tackle the asymmetry problem between the given query and its nearest neighbors is that images which are semantically similar to each other tend to be clustered in some feature space. As can be seen in the synthetic example (See Fig. \ref{fig:CharacvecExam}), retrieved image $i_4$ and $i_6$ are outliers or false positives to the query image $Q$. We can confirm this by observing the common neighbors of $Q$ with $i_4$ and $i_6$. But due to the lack of contextual information, the system considers them as a true match (neighbor) to the query. In our proposed model, to attack this issue, we represent the set of $kNN$ as a graph $G'$ accordingly to subsection \ref{PIWusingCDS}. Then, we treat outliers finding problem as an unsupervised clustering problem. We first convert graph $G'$ into a symmetric affinity matrix $A$, where the diagonal corresponding to each node is set to 0, and the $ij-th$ entry denotes the edge-weight $w_{ij}$ of the graph so that $A_{ij}\equiv A_{ji}$. Accordingly, given graph $G'$ and query $Q$, we cast detecting outliers from a given $NN$ set as finding the most compact and coherent cluster from graph $G',$ which is constrained to contain the query image $Q.$ To this end, we adopt constrained dominant sets \cite{ZemeneP16}, \cite{ZemAP19} which is a generalization of a well known graph-theoretic notion of a cluster. We are given a symmetric affinity matrix $A$ and parameter  $\mu > 0,$  and then we define the following parametrized quadratic program

\begin{equation} \label{eqn:parQP}
\begin{array}{ll}
\text{maximize }  &  f_Q^\mu(X) = X' (A - \mu \hat \Gamma_Q) X \\

&  f_Q^\mu(X) = X' \hat{A} X \\

\text{subject to} &  X \in \Delta
\end{array}
\end{equation}

where a prime denotes transposition and  
$$
\Delta=\left\{ X \in R^n~:~ \sum_{i=1}^n X_i = 1, \text{ and } X_i \geq 0 \text{ for all } i=1 \ldots n \right\}
$$ 
$\Delta$ is the standard simplex of $R^n$. $\hat \Gamma_Q$ represents $n \times n$ diagonal matrix whose diagonal elements are set to zero in correspondence to the query $Q$ and to 1 otherwise. And $\hat{A}$ is defined as,

$$
\hat{A} = A - \mu \hat \Gamma_Q  = 
\begin{pmatrix} 
0 & .    & .  &.  \\ 
. & ~~ - \mu ~~ &  .  &.\\
.& .    & - \mu  &. \\ 

.& .   &  .   & -\mu\\ 
\end{pmatrix}
$$ 
where the dots denote the $ij$ th entry of matrix $A.$ Note that matrix $\hat{A}$ is scaled properly to avoid negative values.

Let $Q \subseteq V,$ with $Q \neq \emptyset$ and let $\mu > \lambda_{max}(A_{V \backslash Q} ),$ where $\lambda_{max}(A_{V\backslash q})$ is the largest eigenvalue of the principal submatrix of $A$ indexed by the element of $V\backslash q.$ If $X$ is a local maximizer of $ f_Q^\mu(X)$ in $\Delta,$ then $\delta(X) \cap Q \neq \emptyset,$ where, $\delta(X) = {i \in V : X_i > 0.}$ We refer the reader to \cite{ZemeneP16} for the proof.

The above result provides us with a simple technique to determine a constrained dominant set which contains the query vertex $Q.$ Indeed, if $Q$ is the vertex corresponding the query image, by setting 

\begin{equation}
\mu > \lambda (A_{V\backslash Q})
\end{equation}
we are guaranted that all local solutions of eq ($\ref{eqn:parQP}$) will have a support that necessarily contains the query element.
The established correspondence between dominant set (coherent cluster) and local extrema of a quadratic form over the standard simplex allow us to find a dominant set using straightforward continuous optimization techniques known as replicator dynamics, a class of dynamical systems arising in evolutionary game theory [21]. The obtained solution provides a principled measure of a cluster cohesiveness as well as a measure of vertex participation. Hence, we show that by fixing an appropriate threshold $\zeta$ on the membership score of vertices, to extract the coherent cluster, we could easily be able to detect the outlier nodes from the k-nearest neighbors set. For each $X^i,$ $\zeta^i$ is dynamically computed as
\begin{equation}
\zeta^i = \Lambda(1 - max(X^i) + min(X^i))/L
\end{equation}

where $max(X)$ and $min(X)$  denote the maximum and minimum membership score of $X^i,$ respectively. $\Lambda$ is a scaling parameter and $L$ stands for length of $X^i.$ Moreover, we show an effective technique to quantify the usefulness of the given features based on the dispersive degree of the obtained characteristics vector $X.$ 

\subsubsection{PIW Using Entropy of Constrained Cluster.}

Entropy has been successfully utilized in a variety of computer vision applications, including object detection \cite{SznitmanBFF13}, image retrieval \cite{DeselaersWN06} and visual tracking \cite{MaLFZ15}. In this chapter, we exploit the entropy of a membership-score of nodes in the constrained dominant set to quantify the usefulness of the given features. To this end, we borrowed the concept of entropy in the sense of information theory (Shannon entropy). We claim that the discriminative power of a given feature is inversely proportional to the entropy of the score distribution, where the score distribution is a stochastic vector. Let us say we are given a random variable $C$ with possible values $c_1, c_2, . . . c_n,$ according to statistical point of view the information of the event ($C = c_i$) is inversely proportional to its likelihood, which is denoted by $I(C_i)$ and defined as 

\begin{equation}
I(C_i) = log \Big( \frac{1}{P(c_i)}\Big) = -log (p(c_i)).
\end{equation}
 Thus, as stated by \cite{Shannon}, the entropy of $C$ is the expected value of I, which is given as 
 \begin{equation}
 	H(C) = - \sum_{i=1}^N P(c_i)log(P(c_i).
 \end{equation}
For each characteristic vector {\small $X^i,X^{i + 1} . . . X^z,$} where {\small $X^i = \big\{X^i_\mu, X^i_{\mu + 1}. . . X^i_{n}\big\},$} we compute the entropy $H(exp({X^i}))$. Each $X^i$ corresponds to the membership score of nodes in the CDS, which is obtained from the given feature $F^i$.  Assume that the top NNs obtained from feature x are irrelevant to the query Q, thus the resulting CDS will only contain the constraint element Q. Based on our previous claim, since the entropy of a singleton set is 0, we can infer that the feature is highly discriminative. Although this conclusion is right, assigning a large weight to feature with irrelevant NNs will have a negative impact on the final similarity. To avoid such unintended impact, we consider the extreme case where the entropy is 0. Following, we introduce a new term $C_a,$ which is obtained from the cardinality of a given cluster, $K_c,$ as 	$Ca^i = \frac{K_c ^i}{\sum_{i=1}^z  K_c^i}.$ As a result, we formulate the PIW computation from the additive inverse of the entropy $\varepsilon ^i =  1-H(X^i),$ and $C_a^i,$ as: 
\begin{equation}\label{eq.PIW}
PIW^i = \frac{\vartheta ^i}{\sum_{i=1}^z  \vartheta^i} \quad Thus, \sum_{i=1}^{z} PIW^i = 1
\end{equation}
where $\vartheta ^i = \varepsilon^i + C_a^i,$ and $i$ represents the corresponding feature.

\subsubsection{Naive Voting Scheme and Similarity Fusion}

In this section, we introduce a simple yet effective voting scheme, which is based on the member nodes of k-nearest neighbor sets and the constrained dominant sets, let {\small $NN_1, NN_2 . . . NN_z$} and {\small $CDS_1, CDS_2 . . . CDS_z$} represent the $NN$ and $CDS$ sets respectively, which are obtained from $G'_1,G'_2 . . . G'_z.$ Let us say $\xi = 2(z-1) - 1,$ and then  we build $\xi$  different combinations of $NN$ sets, $\varphi_1, \varphi_2 . . . \varphi_{\xi}.$ Each $\varphi$ represents an intersection between $z - 1$ unique combinations of $NN$ sets. We then form a super-set $\varpi$ which contains the union of $\varphi$ sets, with including repeated nodes. Assume that {\small $NNs = \{NN_1, NN_2, NN_3\}, \xi = 3,$} thus each $\varphi$ set contains the intersection of two $NN$ sets as {\small $\varphi_1 = \{NN_1 \cap NN_2\}, \varphi_2 = \{NN_1 \cap NN_3\}$} and {\small $\varphi_3=\{NN_2 \cap NN_3\}.$} Hence the resulting $\varpi$ is defined as $\varpi = (\varphi_1 \ominus \varphi_2 \ominus \varphi_3),$ where $(.\ominus.)$ is an operator which returns the union of given sets, including repeated nodes. We have also collected the union of $CDS$ sets as {\small $\omega = (CDS_1 \ominus CDS_2 \ominus CDS_3).$} Following, we compute $\kappa$ as $(\kappa = \varphi_1 \cap \varphi_2 \cap . . . \varphi_{\xi}).$ Thereby we find super-sets $\varpi, \omega$ and $\kappa.$ Next, we design three different counters, which are formulated to increment when the NN node appears in the corresponding super-sets. Based on the value obtained from each counter, we finally compute the vote scores for each $NN$ node to the query as $v_1 = v_1/\eta, v_2 = v_2/\theta$ and $v_3 = v_3/\iota,$ where $\eta, \theta$ and $\iota$ are parameters which are fixed empirically. Note that the outlier detecting capability of our framework is encoded in the voting process. Thus, if a NN node $n_{i}$ is contained in more than one cluster, its probability to be given a large weight is higher. This is due to the number of votes it gets from each cluster.

\subsubsection{Final Similarity.}

After obtaining the aforementioned terms, we compute the final similarity as follows: say we are given $n$ features, $Q$ is the query image and $D$ denotes image  dataset, then the initial similarity of $D$ to $Q$, with respect to feature $F_i, i = 1 . . . n,$ ,is given as $S_{D, Q}^{(i)}.$ Let $PIW_{Q}^{(i)}, i = 1 . . . n,$ encode the weight of feature $F_{i}$ for query $Q,$ and then the final similarity score, $F _{sim (Q, D)} $, between $Q$ and $D$ is given as

\begin{equation}
N_{s} = \prod_{i = 1}^{k} (S_{D, Q}^{(i)})^{PIW_Q^{(i)}} \\
\end{equation}

\begin{equation}\label{equ:FinSim}
F _{sim (Q, D)} = \lambda N_s + (1 - \lambda) \sum_{\Omega = 1}^{\Psi} v_\Omega
\end{equation}
where $\Psi = 3,$ is the total number of voter sets. And $\lambda \in [0, 1]$ defines the penalty factor which penalizes the similarity fusion, when $\lambda = 1$ only $F_s$ is considered, otherwise, if $\lambda = 0,$ only $v$ is considered. 


\subsection{Experiments}

In this section, we present the details about the features, datasets and evaluation methodology we used along with rigorous experimental analysis.

\subsubsection{Datasets and Metrics} 

To provide a thorough evaluation and comparison, we evaluate our approach on INRIA Holiday, Ukbench, Oxford5k and Paris6k datasets. 

\textbf{Ukbench Dataset \cite{NisterS06}.} Contains 10,200 images which are categorized into 2,550 groups, each group consists of three similar images to the query which undergo severe illumination and pose variations. Every image in this dataset is used as a query image in turn while the remaining images are considered as dataset images, in ``leave-one-out" fashion. As customary, we used the N-S score to evaluate the performance of our method, which is based on the average recall of the top 4 ranked images.

\textbf{INRIA Holiday Dataset \cite{JegDouSchECCV2008}.} Comprises 1491 personal holiday pictures including 500 query images, where most of the queries have one or two relevant images. Mean average precision (MAP) is used as a performance evaluation metric.

\textbf{Oxford5k Dataset \cite{PhilbinCISZ07}.} It is one of the most popular retrieval datasets, which contains 5062 images, collected from flicker-images by searching for landmark buildings in the Oxford campus. 55 queries corresponding to 11 buildings are used.

\textbf{Paris6k Dataset \cite{PhilbinCISZ08}.} Consists of 6392 images of Paris landmark buildings with 55 query images that are manually annotated.

\begin{figure*}
	
	\begin{center}
		
		\includegraphics[width=1\linewidth ,trim=0cm 5.5cm 0cm 0cm,clip]{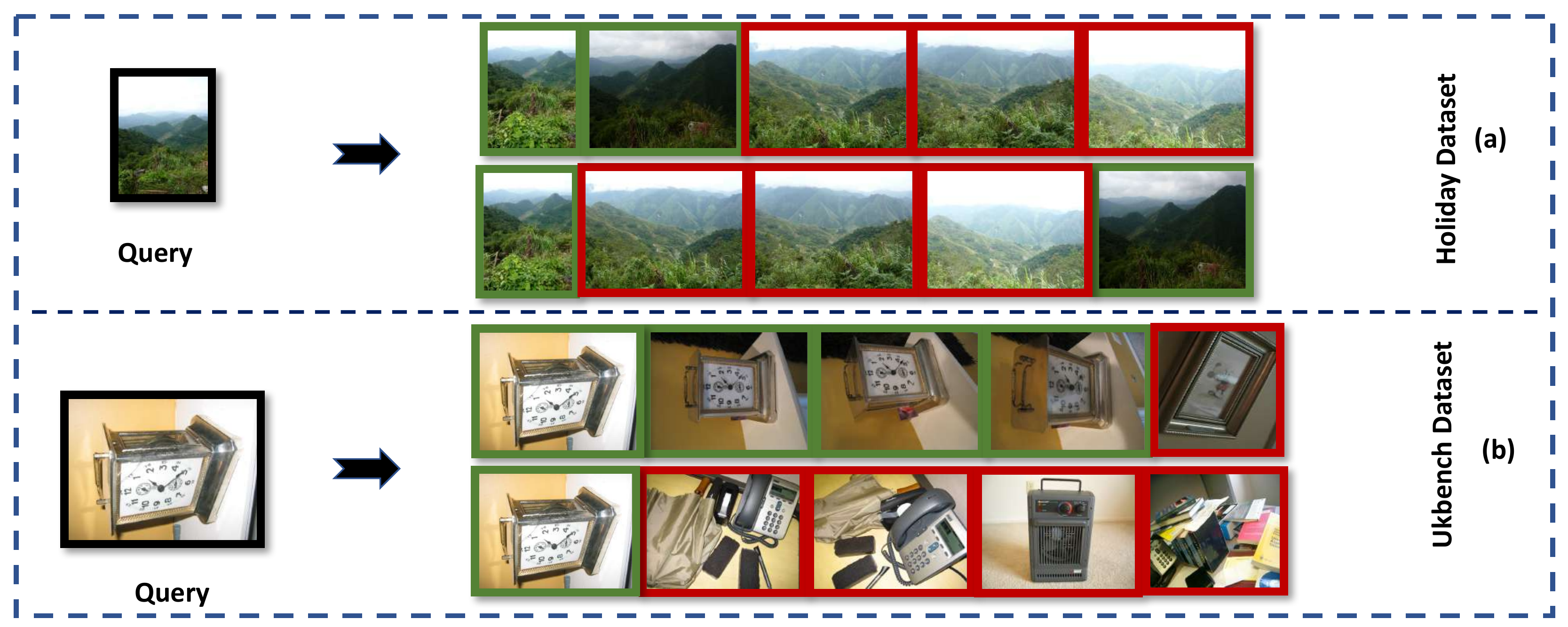}
	\end{center}
	
	\caption{ five relevant images to the query where the green and red frame indicate the True and False posetives to the query, respectively. \textbf{Top-row (a) and (b):} show the top five relevant images of our proposed method. \textbf{Bottom row (a) and (b):} show the top five relevant images obtained from a Naive fusion of several features.}
	\label{fig:HOLex}
\end{figure*}

\subsubsection{Image Features}

\textbf{Object Level Deep Feature Pooling (OLDFP)\cite{MopuriB15}.} OLDFP is a compact image representation, whereby images are represented as a vector of pooled CNN features describing the underlying objects. Principal Component Analysis (PCA) has been employed to reduce the dimensionality of the compact  representation. We consider the top 512-dimensional vector in the case of the Holiday dataset while considering the top 1024-dimensional vector to describe images in the Ukbench dataset.  As suggested in \cite{MopuriB15}, we have applied power normalization (with exponent 0.5), and l2 normalization on the PCA projected image descriptor.

\textbf{BOW.} Following \cite{ZhengWLT14}, \cite{ZheWanTiaHeLiuTiaCVPR2015}, we adopt Hamming Embedding \cite{JegDouSchECCV2008}. SIFT descriptor and Hessian-Affine detector are used in feature extraction, and we used 128-bit vector binary signatures of SIFT. The Hamming threshold and weighting parameters are set to 30 and 16 respectively, and three visual words are provided for each key-point. Flickr60k data \cite{JegDouSchECCV2008} is used to train a codebook of size 20k. We also adopt root sift as in \cite{ArandjelovicZ12}, average IDF as defined in \cite{ZhengWLT13} and the burstiness weighting \cite{JegouDS09}.

\textbf{NetVLAD \cite{ArandjelovicGTP16}.}  NetVLAD is an end-to-end trainable CNN architecture that incorporates the generalized VLAD layer.


\textbf{HSV Color Histogram.} Like \cite{YangMD15}, \cite{ZheWanTiaHeLiuTiaCVPR2015}, for each image, we extract 1000-dimensional HSV color histograms where the number of bins for H, S, V are 20, 10, 5 respectively.


\begin{table*}[h]
	\centering
	\caption{ The performance of baseline features on Holidays, Ukbench, Oxford5k and Paris6k datasets. }
	\smallskip
	\begin{tabular}{|c|p{1.4cm}|p{1.4cm}|p{0.8cm}|c|c|p{0.8cm}|p{0.8cm}|p{0.8cm}|p{0.8cm}|} 
		\hline
		 Datasets & Metrics&  NetVLAD \cite{ArandjelovicGTP16} &  BOW &  OLDFP& HSV& $R_{res}$\cite{IscenTAFC17}&  $G_{res}$\cite{IscenTAFC17}& $R_{vgg}$ \cite{IscenTAFC17} & $G_{vgg}$\cite{IscenTAFC17} \\\hline\hline
		\textbf{Holidays} & MAP& 84&80 &87 &65& -& -&-&-\\
		\textbf{Ukbench} &N-S score& 3.75& 3.58& 3.79&3.19& &-&-&-\\
		\textbf{Oxford5k} &MAP& 69& -&- &-& 95.8&87.7&93&-\\
		\textbf{Paris6k} &MAP& -& -&- &-& 96.8 & 94.1 & 96.4 &95.6\\
		
		\hline
	\end{tabular}
	\label{table:ImagePair}
\end{table*}


\subsubsection{Experiment on Holiday and Ukbench Datasets}

As it can be seen in Fig.\ref{fig:HOLex}(a), the noticeable similarity between the query image and the irrelevant images, in the Holiday dataset, makes the retrieval process challenging. For instance, (See Fig.\ref{fig:HOLex}(a)), at a glance all images seem similar to the query image while the relevant are only the first two ranked images. Moreover, we can observe that the proposed scheme is invariant to image illumination and rotation change. Table \ref{table:comparison} shows that our method significantly improves the MAP of the baseline method \cite{MopuriB15} on Holiday dataset by 7.3 $\%$ while improving the state-of-the-art method by 1.1 $\%$. Likewise, it can be seen that our method considerably improves the N-S score of the baseline method \cite{MopuriB15} on the Ukbench dataset by 0.15 while improving the state-of-the-art method by 0.03. 
\begin{figure*}
	\centering
	\includegraphics[width=1\linewidth ,trim=0cm 0cm 0cm 0cm,clip]{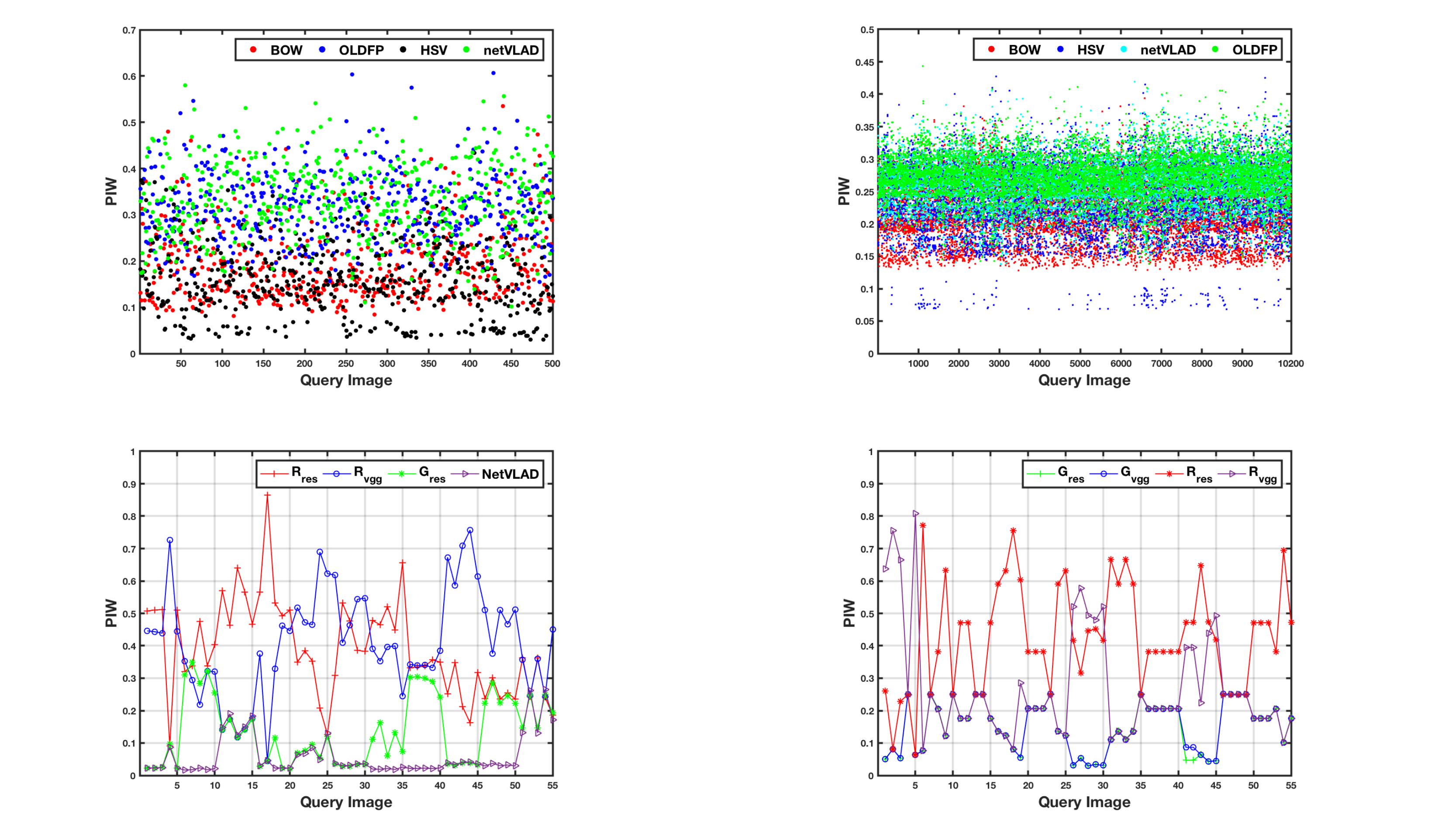}
	\caption{Feature positive-impact weights (PIW's) learned by our algorithm. Top-left, top-right, bottom-left, and bottom-right: on Holiday, Ukbench, Oxford5k and Paris6k datasets, respectively.}
	\label{fig:learnedPIW}
\end{figure*}

\begin{figure*}
	\centering
	\includegraphics[width=1\linewidth ,trim=0cm 9.5cm 0cm 0cm,clip]{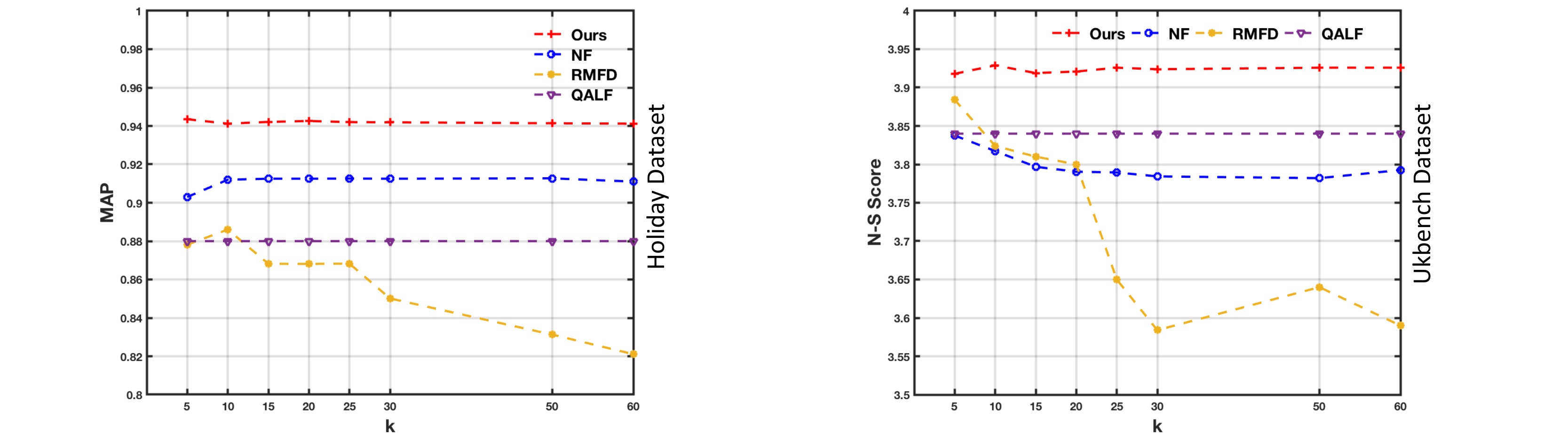}
	\caption{Comparison with state-of-the-art fusion methods with respect to varying k. Naive Fusion (NF), Reranking by Multi-feature Fusion (RMFD) \cite{YangMD15}, and QALF \cite{ZheWanTiaHeLiuTiaCVPR2015}.}
	\label{fig:WithPIWewithourPIW}
\end{figure*}

Furthermore, to show how effective the proposed feature-weighting system is, we have experimented by fusing the given features with and without PIW. Naive fusion (NF) denotes our approach with a constant PIW for all features used, thus the final similarity $F_s$ defined as $F_s = \frac{1}{k} (\prod_{i = 1}^k (S_{D, Q}^{(i)})).$ In Fig.\ref{fig:WithPIWewithourPIW} we have demonstrated the remarkable impact of the proposed PIW. As can be observed, our scheme effectively uplifts the impact of a discriminative feature while downgrading the inferior one. Note that in the PIW computation we have normalized the minimum entropy (See eq.\ref{eq.PIW}), thus its values range between 0 and 1. Accordingly, one implies that the feature is highly discriminative, while zero shows that the feature is indiscriminate.

\begin{figure*}
	\centering
	\includegraphics[width=1\linewidth ,trim=0cm 0cm 0cm 0cm,clip]{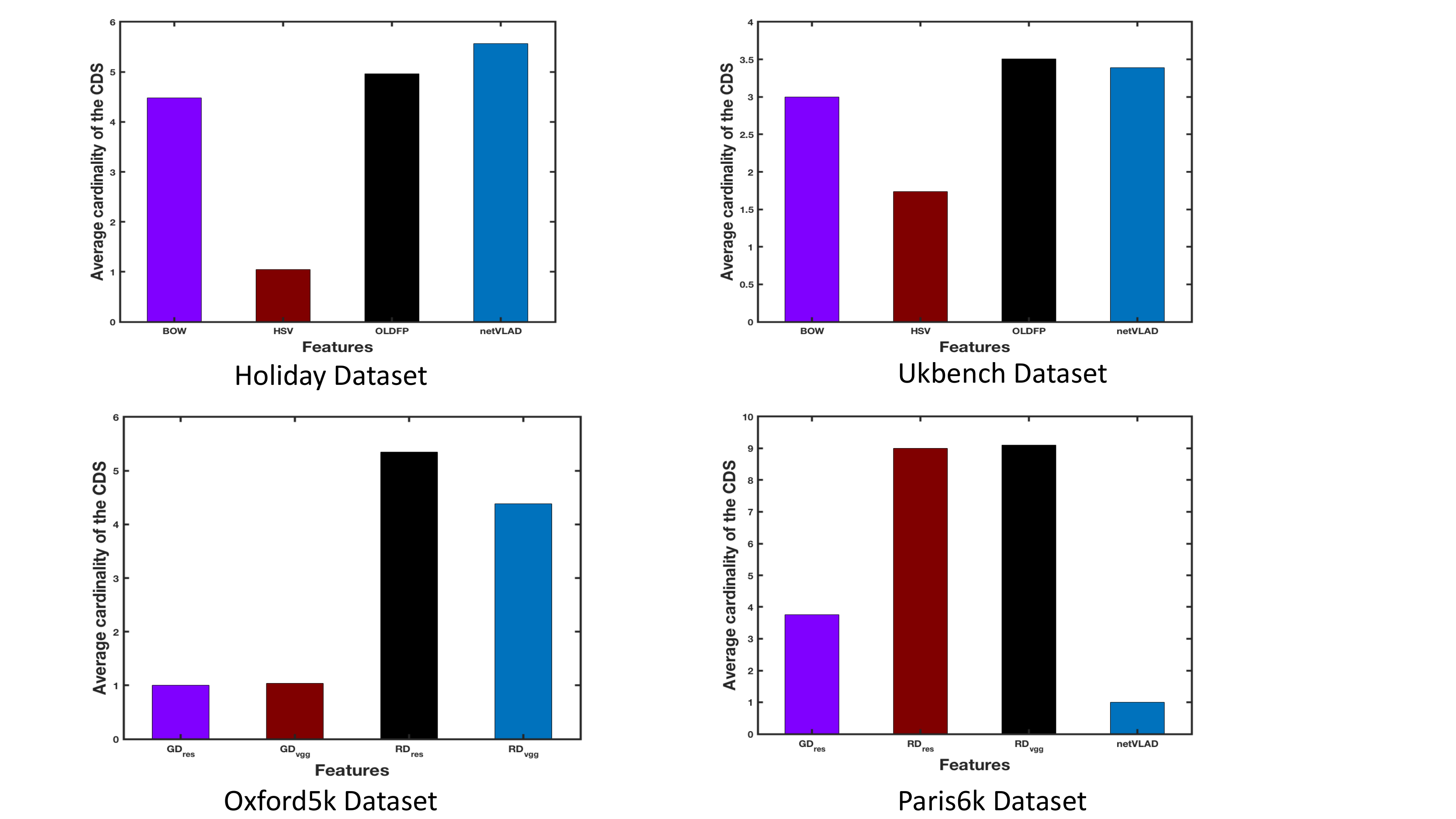}
	\caption{The cardinality of constrained dominant sets for the given features.}
	\label{fig:ClusterCardinality}
\end{figure*}
In order to demonstrate that our scheme is robust in handling outliers, we have conducted an experiment by fixing the number of NNs (disabling the incrimental NNs selection) to different numbers. As is evident from Fig.\ref{fig:WithPIWewithourPIW}, the performance of our method is consistent regardless of the number of $kNN$. As elaborated in subsection \ref{kRN}, the robustness of our method to the number of $k$ comes from the proposed outlier detection method. Since the proposed outliers detector is formulated in a way that allows us to handle the outliers, we are easily able to alleviate the false matches which are incorporated in the nearest neighbors set. This results in finding a nearly constant number of nearest neighbors regardless of the choice of $k$.

\begin{table}[h]
	\centering
	\caption{\small Comparison among various retrieval methods with our method on benchmark datasets, where QALF is implemented with the same baseline similarities used in our experiments.}
	\begin{tabular}{|c|c|c|p{0.7cm}|p{0.6cm}|p{0.6cm}|p{0.5cm}|p{0.5cm}|p{0.5cm}|p{0.5cm}|p{0.5cm}|c|} 
		\hline
		Datasets & Metrics& Baselines  & QALF\cite{ZheWanTiaHeLiuTiaCVPR2015} &  \cite{YangMD15} &NF& ED\cite{BaiZWBLT17} & \cite{GordoARL16}&\cite{RadenovicTC16} &\cite{XuSQWX17}&\cite{BabenkoSCL14}&   Ours \\
	
		\hline\hline
		\textbf{Ukbench} &\scriptsize{N-S score}& 3.79\cite{MopuriB15} & 3.84 & 3.86 & 3.86&3.93&-&-&-&3.76&\textbf{3.94}      \\
		\textbf{Holiday}   &\scriptsize{MAP}& 87\cite{MopuriB15} &  88& 88&91&93 &90&83&89&77&\textbf{94}       \\   		 
		\textbf{Oxford5k}  &\scriptsize{MAP}&  95.8\cite{IscenTAFC17}&  \textbf{-} & 76.2& 94.4&-&89.1&79.7&81.4&67.6&\textbf{96.2}       \\ 
		\textbf{Paris6k}  & \scriptsize{MAP}& 96.8\cite{IscenTAFC17}&  - & 83.3 & - &-&91.2&83.8&88.9&-& \textbf{97.4}\\ 
		\hline
	\end{tabular}
	\label{table:comparison}
\end{table}

\subsubsection{Experiment on Oxford5k and Paris6k Datasets}
In the same fashion as the previous analysis, we have conducted extensive experiments on the widely used Oxford5k and Paris6k datasets. Unlike the Holiday and Ukbench datasets, we adapt affinity matrices which are obtained through a diffusion process on a regional $Resnet$ and $VGG$ representation \cite{IscenTAFC17}, and they are denoted as $R_{res}$ and $R_{vgg}$ respectively, as well as affinity matrices $G_{res}$ and $G_{vgg}$ which are also obtained through a diffusion process on a global $Resnet$ and $VGG$ representation, respectively. Table \ref{table:comparison} shows that the proposed method slightly improves the state-of-the-art result. Even if the performance gain is not significant, our scheme marginally achieves better MAP over the state-of-the-art methods. Furthermore, as shown in Fig \ref{fig:learnedPIW}, the proposed model learns the PIW of the given features effectively. Therefore, a smaller average weight is assigned to $G_{vgg}$ and $NetVLAD$ feature comparing to $R_{res}$ and $R_{vgg}$.

\subsubsection{Robustness of Proposed PIW}
As can be seen in Fig \ref{fig:learnedPIW}, for all datasets, our algorithm has efficiently learned the appropriate weights to the corresponding features. Fig. \ref{fig:learnedPIW} shows how our algorithm assigns PIW in a query adaptive manner. In Holiday and Ukbench datasets, the average weight given to HSV feature is much smaller than all the other features used. Conversely, a large PIW is assigned to OLDFP and NetVLAD features. Nevertheless, it is evident that in some cases a large value of PIW is assigned to HSV and BOW features as well, which is appreciated considering its effectiveness on discriminating good and bad features in a query adaptive manner.

\subsubsection{Impact of Parameters}

\begin{figure*}
	\centering
	\includegraphics[width=1\linewidth ,trim=0cm 0cm 0cm 0cm,clip]{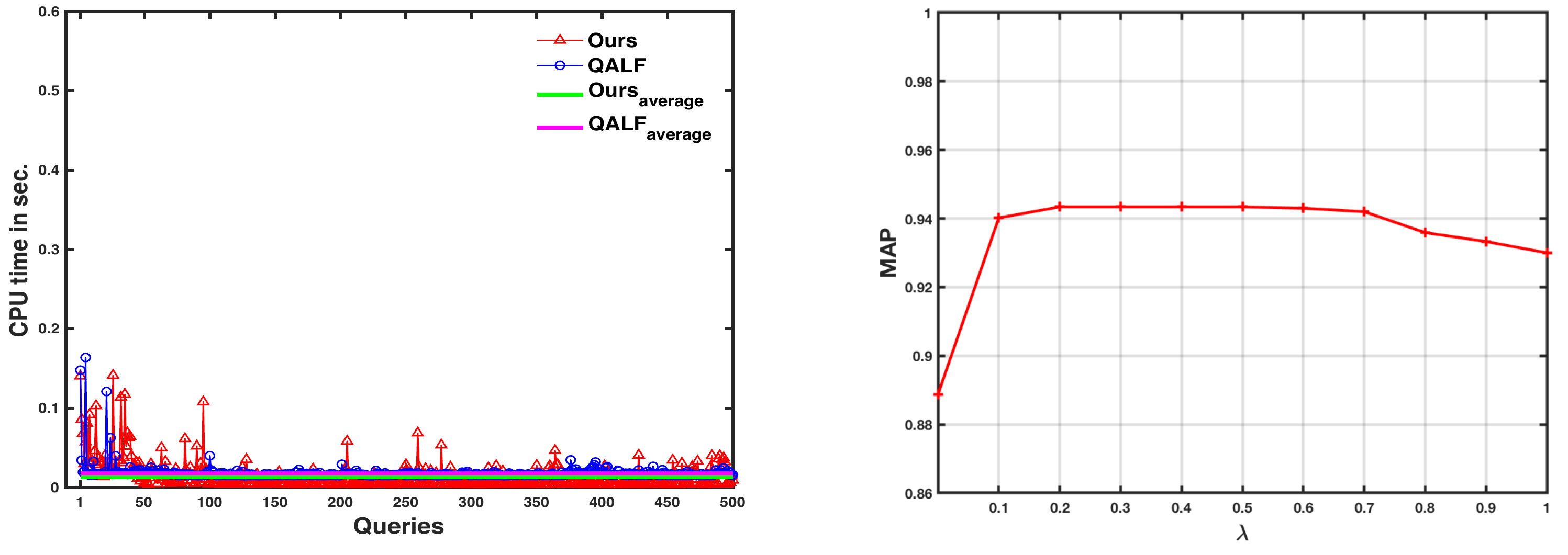}
	\caption{Left: Time complexity of our algorithm (red) and QALF\cite{ZheWanTiaHeLiuTiaCVPR2015} (blue) on Holiday dataset. Right: The impact of $\lambda$ on the retrieval performance, on Holiday dataset.}
	\label{fig:ComplexityeLambda}
\end{figure*}

To evaluate the robustness of our method we have performed different experiments by changing one parameter at a time. Thereby, we have observed that setting  $\Lambda$ to a large value results in assigning insignificant PIW to indiscriminate features. The reason is that after the application of CDS, the cluster membership-score of the dissimilar images to the query will become smaller. Thus, since the threshold fixed to choose the true neighbors is tighter, the resulting constrained dominant set will be forced  to yield a singleton cluster. As a result, we obtained a very small PIW due to the cardinality of the constrained-cluster. In addition to that, we observe that the MAP start to decline when $\lambda$ is set to a very large value (See. Fig \ref{fig:ComplexityeLambda}, right).

\subsubsection{Impact of Cluster Cardinality}
On the Ukbench dataset, as can be observed in Fig. \ref{fig:ClusterCardinality}, the average cardinality of the constrained clusters which is obtained from HSV and BOW feature is 3 and 1.7, respectively. In contrast, for NetVLAD and OLDFP, the average cluster cardinality is 3.4 and 3.5, respectively . Similarly, in the case of the Holiday dataset, the cluster cardinality obtained from HSV feature is one while for BOW, NetVLAD and OLDFP is 4.5, 5 and 5.6, respectively. Thus, from this, we can draw our conclusion that the cardinality of a constrained dominant set, in a certain condition, has a direct relationship with the effectiveness of the given feature.

\subsubsection{Computational Time}
In Fig. \ref{fig:ComplexityeLambda} we depict the query time taken to search for each query image, red and blue lines represent our method and QALF, respectively. The vertical axis denotes the CPU time taken in seconds, and the horizontal axis shows the query images. As can be seen from the plot, the proposed framework is faster than the fastest state-of-the-art feature-fusion method \cite{ZheWanTiaHeLiuTiaCVPR2015}. As for time complexity, in our experiment we used a replicator dynamics to solve problem (\ref{eqn:parQP}), hence, for a graph with N nodes, the time complexity per step is $O(N^2),$ and the algorithm usually takes a few steps to converge, while that of \cite{Bai17} is $O(N^3).$ However, we note that by using the Infection-immunization  algorithm \cite{BuloPB11} we can achieve even faster convergence as its per-step complexity would be linear in the number of nodes.

\subsection{Summary}
In this chapter, we addressed a multi-feature fusion problem in CBIR. We developed a novel and computationally efficient CBIR method based on a constrained-clustering concept. In particular, we showed an efficient way of estimating a positive impact weight of features in a query-specific manner. Thus it can be readily used for feature combination. Furthermore, the proposed scheme is fully unsupervised, and can easily be able to detect false-positive NNs to the query, through the diffused similarity of the NNs. To demonstrate the validity of our method, we performed extensive experiments on benchmark datasets. Besides the improvements achieved on the state-of-the-art results, our method shows its effectiveness in quantifying the discriminative power of given features. Moreover, its effectiveness on feature-weighting can also be exploited in other computer vision problems, such as person re-identification, object detection, and image segmentation.

\section{Constrained Dominant Sets for Image Retrieval}

	Learning new global relations based on an initial affinity of the database objects has shown significant improvements in similarity retrievals. Locally constrained diffusion process is one of the recent effective tools in learning the intrinsic manifold structure of a given data. Existing methods, which constrained the diffusion process locally, have problems - manual choice of optimal local neighborhood size, do not allow for intrinsic relation among the neighbors, fix initialization vector to extract dense neighbor - which negatively affect the affinity propagation. We propose a new approach, which alleviate these issues, based on some properties of a family of
	quadratic optimization problems related to dominant sets, a well-known graph-theoretic notion of a cluster
	which generalizes the concept of a maximal clique to edge-weighted graphs. In particular, we show that by properly controlling a regularization parameter which determines the structure and the scale of the underlying problem, we are in a position to extract dominant set cluster which is constrained to contain user-provided query.
	Experimental results on standard benchmark datasets show the effectiveness of the proposed approach.

\subsection{Similarity Diffusion for Image Retrieval} 
\label{introduction}

Retrieval has recently attracted considerable attention within the computer vision community, especially because of its potential applications such as database retrieval, web and mobile image search. Given user provided query, the goal is to provide as output a ranked list of objects that best reflect the user's intent. Classical approaches perform the task based on the (dis)similarity between the query and the database objects. The main limitation of such classical retrieval approaches is that they does not allow for the intrinsic relation among the database objects.

Recently, various techniques, instead of simply using the pairwise similarity, they try to learn a better similarities that consider manifold structures of the underlying data. Qin \et  in \cite{QinGamBosQuaGooCVPR2011} try to alleviate the asymmetry problem of the k-nearest neighbor (k-nn) using the notion of k-reciprocal nearest neighbor. In \cite{HouKriKroSchZimSSDBM2010} the notion of shared nearest neighbor is used to build secondary similarity measure, which stabilize the performance of the search, based on the primary distance measure.
In \cite{EgoKelGutTIP2010} shape meta-similarity measure, which is computed as the {\textbf{L1}} distance between new vector representation which considers only the k-nn set of similarities fixing all others to 0, was proposed. Choosing the right size of the neighbor is important. In \cite{KonDonBisACCV2009}, the notion of shortest path was used to built a new affinity for retrieval.


Diffusion process is one of the recent effective tools in learning the intrinsic manifold structure of a given data \cite{DonBisCVPR2013,YanLatCVPR2011,YanPraLatPAMI2013}. Given data, a weighted graph is built where the nodes are the objects and the edge weight is a function of the affinity between the objects. The pairwise affinities are then propagated following structure of the weighted edge links in the graph. The result of the affinity propagation highly depends on the quality of the pairwise similarity \cite{LafLeePAMI2006,SzuJaaNIPS2001}. Inaccurate Pairwise similarity results in a graph with much noise which negatively affects the diffusion process.  Constraining the diffusion process locally alleviate this issue \cite{SzuJaaNIPS2001,YanPraLatPAMI2013,DonBisCVPR2013}. Dominant neighbor (DN) and k-nn are two notions used by the recent existing methods to constrain the diffusion process locally \cite{DonBisCVPR2013,YanLatCVPR2011,YanPraLatPAMI2013}. In \cite{DonBisCVPR2013}, it has been shown that affinity learning constraining relation of an object to its neighbors effectively improves the retrieval performance and was able to achieve 100 \% bull's eye score in the well known MPEG datset. The author of \cite{DonBisCVPR2013} put automatically selecting local neighborhood size (K) as the main limitation of the approach and is still an open problem. The influence of selecting different K values was also studied which proved that the parameter is a serious problem of the approach. For MPEG7 dataset, the choice is insignificant while for the other two datasets YALE and ORL choosing the reasonable K is difficult which resulted in a decrease in performance for the right value of K. Moreover, it is obvious that the selection of k-nn is prone to errors in the pairwise similarities \cite{YanPraLatPAMI2013}. Since any k-nn decision procedure relies only on affinities of an object to all other objects, k-nn approach is handicapped in resisting errors in pairwise affinities and in capturing the structure of the underlying data manifold.

Yang \et  in \cite{YanPraLatPAMI2013}, to avoid the above issues, proposed the notion of dominant neighbors (DN). Instead of the k-nn, here a compact set from the k-nn which best explains the intrinsic relation among the neighbors is considered to constrain the diffusion process. However, the approach follows heuristic based k-nn initialization scheme. To capture dominant neighbors, the approach first choose a fixed value of K, collect the K nearest neighbors and then initialize the dynamics, the dynamics which extracts dense neighbors, to the barycenter of the face of the simplex which contains the neighbors. It is obvious to see that the approach is still dependent on K. Moreover, as fixing K limits the dynamics to a specified face of the simplex, objects out of k-nn($q$) which form a dominant neighbor with $q$ will be loosed. The chosen k-nn may also be fully noisy which might not have a compact structure.

In this section, we propose a new approach to retrieval which can deal naturally with the above problems. Our approach is based on some properties of a family of quadratic optimization problems related to dominant sets, a well-known graph-theoretic notion of a cluster which generalizes the concept of a maximal clique to edge-weighted graphs. In particular, we show that by properly controlling a regularization parameter which determines the structure and the scale of the underlying problem, we are in a position to extract dominant-set cluster which is constrained to contain user-specified query. 

The resulting algorithm has a number of interesting features which distinguishes it from existing approaches.
Specifically: 1) it is able to constrain the diffusion process locally extracting dense neighbors whose local neighborhood size (K) is fixed automatically; different neighbors can have different value of K. 2)it has no any initialization step; the dynamics, to extract the dense neighbors, can start at any point in the standard simplex 3) it turns out to be {\em robust} to noisy affinity matrices.

The rest of this chapter is organized as follows. In the next section we will discuss the most related works to our approach. The experimental results are given in section \ref{sec:Experiments}.

\subsection{Diffusion Process}

Given a set of, say $n$, objects, the relation among them can be represented as an undirected edge-weighted graph $G = (V, E,w)$, where $V = \{1, . . . , n\}$ is the vertex set, $E \subseteq V \times V$ is the edge set, and $w : E \rightarrow R_+^*$ is the (positive) weight function. Vertices in $G$ correspond to data points, edges represent neighborhood relationships, and edge-weights reflect similarity between pairs of linked vertices. As customary, the graph $G$ is represented with the corresponding weighted adjacency (or similarity) matrix, which is the $n \times n$ nonnegative, symmetric matrix $A = (a_{ij})$ defined as $a_{ij} = w(i, j)$, if $(i, j) \in E$, and $a_{ij} = 0$ otherwise. A diffusion process then starts from a predefined initialization, say $\mathcal{V}$ and propagate the affinity value through the underlying manifold based on a predefined transition matrix, say $\mathcal{T}$, and diffusion scheme ($\mathcal{S}$).

Off-the-shelf diffusion processes, which basically differ based on the choice of $\mathcal{V}$, $\mathcal{T}$ and $\mathcal{S}$, the most related ones to this work are \cite{YanPraLatPAMI2013} and \cite{YanTezLatCVPR2009}. In both cases, the diffusion process is locally constrained. While in \cite{YanTezLatCVPR2009} the notion of k-nn is used to constrain the diffusion process locally, dominant neighbor notion ($\mathcal{DN}$) is used by \cite{YanPraLatPAMI2013}. 

\subsubsection{Nearest Neighbors}
In the first case, the edge-weights of the k-nn are kept i.e define locally constrained affinity $\mathcal{L} = (l_{ij})$ defined as $l_{ij} = w(i, j)$, if $(i, j) \in$ k-nn($q$), and $l_{ij} = 0$ otherwise. Then the diffusion process, setting $\mathcal{V}$ as the affinity $A$ , is performed by the following update rule.

\begin{equation}
\mathcal{V}_{t+1} = \mathcal{L}\mathcal{V}\mathcal{L}
\label{eq:diffusion}
\end{equation}

Nearest neighbors constrained diffusion process, alleviating the issue of noisy pairwise similarity, significantly increases the retrieval performance. However, the approach has two serious limitations: First, automatically selecting local neighborhood size (K) is very difficult and is still an open problem \cite{DonBisCVPR2013}. In \cite{DonBisCVPR2013} the influence of selecting different K values was studied which proved that the parameter is a serious problem of the approach. For MPEG7 dataset, the choice was insignificant while for the other two datasets, YALE and ORL, choosing the reasonable K was difficult which even resulted in a decrease in performance, for ORL from 77.30\% to 73.40\% and for YALE 77.08\% to 73.39\%, for the right value of K. Moreover, it is obvious that the selection of k-nn is prone to errors in the pairwise similarities \cite{YanPraLatPAMI2013}. 

\subsubsection{Dominant Neighbors}

Yang \et  in \cite{YanPraLatPAMI2013}, to avoid the above issues, proposed the notion of dominant neighbors ($\mathcal{DN}$). Instead of the k-nn, here a compact set from the k-nn which best explains the intrinsic relation among the neighbors is considered to constrain the diffusion process. To do so, the author used the dominant set framework by Pavan and Pelillo \cite{PavPel07}.

A dominant neighbor ($\mathcal{DN}$) is set as a dominant set, say $\mathcal{DS}$, from the k-nn which contains the user provided query $q$, lets call it $\mathcal{DS}(q)$.

\subsubsection{Proposed Method}

Given a query $q$, we scale the affinity and run the replicator \ref{eqn:Replicator}, say the dynamics converges to $\vct{x}^*$. The support of $\vct{x}^*$, $\sigma(\vct{x}^*)$, is the constrained dominant set which contains the query $q$, let us call it $\mathcal{CDS}(q)$. The edge-weights of the $\mathcal{CDS}(q)$ are then kept i.e define locally constrained affinity $\mathcal{L} = (l_{ij})$ defined as $l_{ij} = w(i, j)$, if $(i, j) \in \mathcal{CDS}(q)$, and $l_{ij} = 0$ otherwise. The diffusion process is then performed by the same update rule as in \ref{eq:diffusion}. For the proof of convergence of the update rule we refer the reader to \cite{YanLatCVPR2011}.

\subsection{Experiments}
\label{sec:Experiments}
The performance of the approach is presented in this section. The approach was tested against three well known data sets in the field of retrieval: MPEG7(shape), YALE(faces) and ORL(faces). For all test data sets the number of iterations for the update rule is set to 200. A given pairwise distance $\mathcal{D}$ is transformed to similarity (edge-weight) using a standard Gaussian kernel $$\mat A^\sigma_{ij}=\ind{i\neq j}exp(-\mathcal{D}/{2\sigma^2})$$ where $\sigma$ is the free scale parameter, and $\ind{P}=1$ if $P$ is true, $0$ otherwise. $\mathcal{L}$ is then built, from $\mat A$,  using the constrained dominant set framework. The diffusion process is then computed using the update rule \ref{eq:diffusion} which resulted in the final learned affinity for ranking. 

\begin{table}[t]
	\label{table:MPEG7}
	\centering
	\begin{tabular}{|c|c|c|c|c|c|c|} 	
		\hline\noalign{\smallskip}
		MPEG7&  B1  & B2 & B3 & B4 & B5 & B6(Ours)\\
		\noalign{\smallskip}
		\hline
		\noalign{\smallskip}
		A1   & 99.91 &   99.93 &  \textbf{100}  & \textbf{100}    & 99.88 & \textbf{100}\\ \hline
		A2   & 99.92 &   99.93 &  \textbf{100}  & \textbf{100}    & 99.88 & \textbf{100}\\ \hline
		A3	 & 99.93 &   99.94 &  \textbf{100}  & \textbf{100}    & 99.88 & \textbf{100}\\ 	
		A4   & 99.92 &   99.94 &  \textbf{100}  & \textbf{100}    & 99.88 & \textbf{100}\\ \hline
	\end{tabular}
	
	\caption{\small Results on MPEG7 dataset. Bull's eye score for the first 40 elements}
\end{table}

A similar experimental analysis as of \cite{DonBisCVPR2013} has been conducted. In \cite{DonBisCVPR2013}, a generic framework with 72 different variant of diffusion processes was defined which are resulted from three steps: initialization, definition of transition matrix and diffusion process. In our experiment, the update scheme is fixed to \ref{eq:diffusion} which has proven to be effective. The four different types of initialization schemes are Affinity Matrix {\bf A} (A1) \cite{WanTuCVPR2012}, Identity Matrix {\bf I} (A2), Transition Matrix {\bf P} which is the standard random walk transition matrix (A3) \cite{MinKyoCVPR2010} and Transition Matrix {\bf P$_{kNN}$} which is the random walk transition matrix constrained to the k-nearest neighbors (A4) \cite{MinKyoCVPR2010}. Including our transition matrix (B6), we have in total 6 different types of transition matrices: {\bf P} (B1), Personalized PageRank Transition Matrix {\bf P$_{PPR}$} (B2) \cite{MinKyoCVPR2010}, {\bf P$_{kNN}$} (B3), Dominant Set Neighbors {\bf P$_{DS}$} \cite{YanPraLatPAMI2013} (B4), and Affinity Matrix {\bf A} (B5) 

\begin{table*}[bth]
	
	\begin{center}
		\begin{tabular}{|c|c|c|c|c|c|} 
			\hline\noalign{\smallskip}
			$\mathcal{R}$ &  20  & 25 & 30 & 35 & 40 \\
			\noalign{\smallskip}
			\hline
			\noalign{\smallskip}
			B3   & 94.321 &  97.871 &  \textbf{98.614}  & 99.357    & \textbf{100}\\ 
			B4   & 94.296 &  97.846 &  \textbf{98.614}  & 99.357    & \textbf{100}\\ 
			Ours & \textbf{94.354} &    \textbf{97.896} &  \textbf{98.614}  & \textbf{99.360}& \textbf{100}\\ 	\hline	
			
		\end{tabular}
	\end{center}
	\caption{\small Results on MPEG7 dataset varying the first $\mathcal{R}$ returned objects}
	\label{table:MPEG7varyingR}
\end{table*}

\textbf{Metric:} The Bull's eye score is used as a measure of retrieval accuracy. It measures the percentage of objects sharing the same class with a query $q$ in the top $\mathcal{R}$ retrieved shapes. Let us say $\mathcal{C}$
is the set of objects in the same class of the query $q$ and $\mathcal{O}$ is the set of top $\mathcal{R}$ retrieved shapes. The Bull's eye score ($\mathcal{B}$) is then computed as $\mathcal{B}$=$\frac{ |\mathcal{O} \cap \mathcal{C}|}{|\mathcal{C}|}$

\textbf{MPEG7:} a well known data set for testing performance of retrieval and shape matching algorithms. It comprises 1400 silhouette shape images of 70 different categories with 20 images in each categories. In all reported results, Articulated Invariant Representation (AIR) \cite{GopTurCheECCV2010}, best performing shape matching algorithm, is used as the input pairwise distance measure. The retrieval performance is measured fixing $\mathcal{R}$ to 40.

The retrieval performance has also been tested by varying the first $\mathcal{R}$ returned objects, the set in which instances of the same category are checked in. For the purpose of this experiment we use the best diffusion variants (B3 and B4 initialized with A2). The performance of the algorithms is shown in Table \ref{table:MPEG7varyingR}.

MPEG7 has been used, most frequently, for testing retrieval algorithms. Table \ref{table:MPEGperformance}  shows the comparison against different state-of-the-art approaches.

\begin{table}[bth]
	\begin{center}
		\begin{tabular}{|c|c|c|c|c|c|c|} 
			\hline\noalign{\smallskip}
			\cite{LinJacPAMI2007}& \cite{BaiYanLatLiuTuPAMI2010} &  \cite{GopTurCheECCV2010} & \cite{LinYanLatECCV2010}  & \cite{YanPraLatPAMI2013} & \cite{DonBisCVPR2013} & Ours\\
			\noalign{\smallskip}
			\hline
			\noalign{\smallskip}
			85.40 &   91.61 & 93.67    & 95.96  & 99.99 & \textbf{100} & \textbf{100} \\ \hline
		\end{tabular}
	\end{center}
	\caption{\small Retrieval performance comparison on MPEG7 dataset. \textbf{Up:} methods, \textbf{Down:} Bull's eye score for the first 40 elements}
	\label{table:MPEGperformance}
\end{table}

\textbf{YALE:} \cite{BelHesKriPAMI1997}  a popular benchmark for face clustering which consists of 15 unique people with 11 pictures for each under different conditions: normal, sad, sleepy, center light, right light, ... etc that include variations of pose, illumination and expression. Similar procedure of \cite{JiaWanTuICCV2011} and \cite{DonBisCVPR2013} were followed to build the distance matrix. Down sample the image, normalize to 0-mean and 1-variance, and compute the Euclidean distance between the vectorized representation. The retrieval performance is measured fixing $\mathcal{R}$ to 15.

\begin{table}[bth]
	\label{table:YaleResult}
	\begin{center}
		\begin{tabular}{|c|c|c|c|c|c|c|} 
			\hline\noalign{\smallskip}
			YALE &  B1  & B2 & B3 & B4 & B5 & B6(Ours)\\
			\noalign{\smallskip}
			\hline
			\noalign{\smallskip}
			A1   & 71.74 &   71.24 &  \textbf{75.59}  & 75.31& 70.25 & 75.15\\ 
			A2   & 71.96 &   70.69 &           77.30  & 76.20& 69.92 & \textbf{77.41}\\ 
			A3	 & 72.07 &   70.57 &           74.93  & 76.14& 70.30 & \textbf{75.37}\\ 	
			A4   & 72.23 &   70.74 &           77.08  & 76.10& 70.25 & \textbf{77.36}\\ \hline
		\end{tabular}
	\end{center}
	\caption{\small Results on YALE dataset. Bull's eye score for the first 15 elements}
\end{table}

Results of the algorithm on YALE data set varying $\mathcal{R}$ is shown in Table \ref{table:YALEvaryingR}.
\begin{table}[bth]
	
	\begin{center}
		\begin{tabular}{|c|c|c|c|c|c|} 
			\hline\noalign{\smallskip}
			$\mathcal{R}$ &  20  & 25 & 30 & 35 & 40 \\
			\noalign{\smallskip}
			\hline
			\noalign{\smallskip}
			B3   & 71.240 &  \textbf{74.105} &  77.303  & 79.559    & 80.826\\ 
			B4   & 70.854 &  72.176 & 76.198  & 77.741    & 79.063\\ 
			Ours & \textbf{71.350} &    74.050 &  \textbf{77.411}  & \textbf{80.000}& \textbf{81.653}\\ 	\hline	
			
		\end{tabular}
	\end{center}
	\caption{\small Results on YALE dataset varying the first $\mathcal{R}$ returned objects}
	\label{table:YALEvaryingR}
\end{table}

\textbf{ORL:} face data set of 40 different persons with 10 grayscale
images per person with slight variations of pose, illumination, and expression. Similar procedure as of YALE data set was followed and The retrieval performance is measured fixing $\mathcal{R}$ to 15.

\begin{table}[bth]
	\label{table:TableOrlResult}
	\begin{center}
		\begin{tabular}{|c|c|c|c|c|c|c|} 
			\hline\noalign{\smallskip}
			ORL &  B1  & B2 & B3 & B4 & B5 & B6(Ours)\\
			\noalign{\smallskip}
			\hline
			\noalign{\smallskip}
			A1   & 72.75 &   73.48 &  \textbf{74.25}  & 73.90& 70.58 & \textbf{74.25}\\ 
			A2   & 72.75 &   73.75 &  \textbf{77.42}  & 74.82& 70.15 & \textbf{77.42}\\ 
			A3	 & 73.12 &   73.75 &  \textbf{75.52}  & 75.35& 71.05 & \textbf{75.52}\\ 	
			A4   & 73.12 &   73.75 &  \textbf{77.32}  & 75.50& 71.40 & \textbf{77.32}\\ \hline
		\end{tabular}
	\end{center}
	\caption{\small Results on ORL dataset. Bull's eye score for the first 15 elements}
\end{table}

Results of the algorithm on ORL data set varying $\mathcal{R}$ is shown in Table \ref{table:ORLvaryingR}.
\begin{table}[bth]
	
	\begin{center}
		\begin{tabular}{|c|c|c|c|c|c|} 
			\hline\noalign{\smallskip}
			$\mathcal{R}$ &  20  & 25 & 30 & 35 & 40 \\
			\noalign{\smallskip}
			\hline
			\noalign{\smallskip}
			B3   & \textbf{70.950} &  \textbf{75.250} &  \textbf{77.425}  & \textbf{79.275}   & \textbf{80.550}\\ 
			B4   & 68.850 &  72.900 & 74.825  & 76.775    & 77.700\\ 
			Ours & \textbf{70.950} &  \textbf{75.250} &  \textbf{77.425}  & \textbf{79.275}   & \textbf{80.550}\\ 	\hline	
			
		\end{tabular}
	\end{center}
	\caption{\small Results on ORL dataset varying the first $\mathcal{R}$ returned objects}
	\label{table:ORLvaryingR}
\end{table}

\subsection{Summary}

In this work, we have developed a locally constrained diffusion process which, as of existing methods, has no problems such as choosing optimal local neighbor size and initializing the dynamics to extract dense neighbor which constrain the diffusion process. The framework alleviates the issues with an up-tick in the results.

\newpage
\chapter{Deep Constrained Dominant Sets for Person Re-identification}
\label{chap:intro}

In this work, we propose an end-to-end constrained clustering scheme to tackle the person re-identification (re-id) problem. Deep neural networks (DNN) have recently proven to be effective on person re-identification task. In particular, rather than leveraging solely a probe-gallery similarity, diffusing the similarities among the gallery images in an end-to-end manner has proven to be effective in yielding a robust probe-gallery affinity. However, existing methods do not apply probe image as a constraint, and are prone to noise propagation during the similarity diffusion process. To overcome this, we propose an intriguing scheme which treats person-image retrieval problem as a {\em constrained clustering optimization} problem, called deep constrained dominant sets (DCDS). Given a probe and gallery images, we re-formulate person re-id problem as finding a constrained cluster, where the probe image is taken as a constraint (seed) and each cluster corresponds to a set of images corresponding to the same person. By optimizing the constrained clustering  in an end-to-end manner, we naturally leverage the contextual knowledge of a set of images corresponding to the given person-images. We further enhance the performance by integrating an auxiliary net alongside DCDS, which employs a multi-scale ResNet. To validate the effectiveness of our method we present experiments on several benchmark datasets and show that the proposed method can outperform state-of-the-art methods.

\begin{figure}[t]

	\begin{center}
		
		\includegraphics[width=1\linewidth ,trim=0cm 0cm 0cm 0cm,clip]{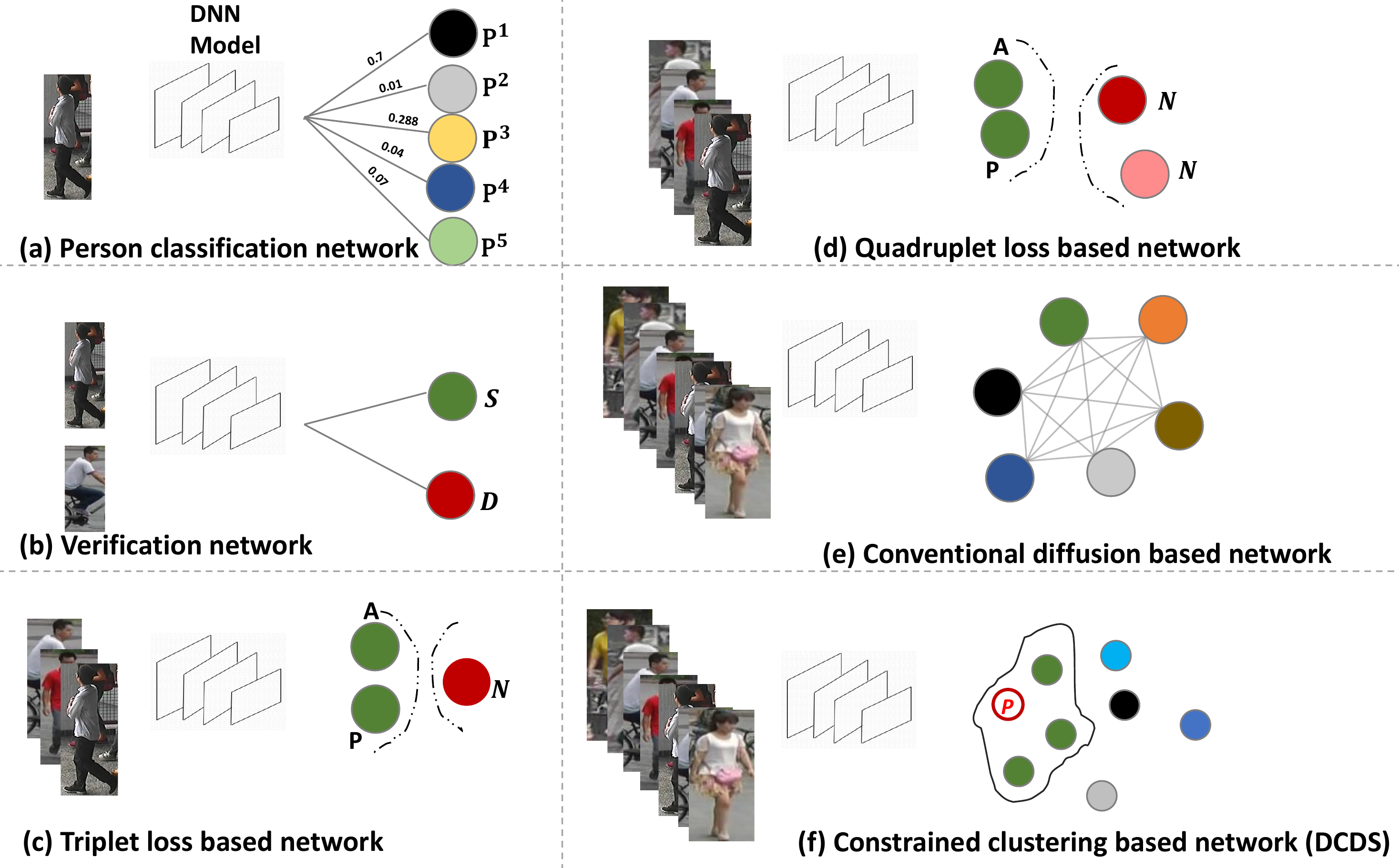}
	\end{center}
	
	\caption{Shows a variety of existing classification and similarity-based deep person re-id models. (a) Depicts a classification-based deep person re-id model, where $P^i$ refers to the $i^{th}$ person. (b) Illustrates a verification network whereby  the similarity $S$ and dissimilarity D for  a pair of images is found.  (c) A Triplet loss based DNN, where $A, P, N$ indicate anchor, positive, and negative samples, respectively. (d) A quadruplet based DNN 
	(e) Conventional diffusion-based DNN, which leverages the similarities among all the images in the gallery to learn a better similarity. (f) The proposed deep constrained dominant sets (DCDS), where, $\textcolor{red}{P}$ indicates the constraint (probe-image); and, images in the constrained cluster, the enclosed area, indicates the positive samples to the probe image.}
	
	\label{fig:metaphor}
\end{figure}


\section{Introduction}
Person re-identification aims at retrieving the most similar images to the probe image, from a large scale gallery set captured by camera networks. Among the challenges which hinder person re-id tasks, include background clutter, Pose, view and illumination variation can be mentioned.

Person re-id can be taken as a person retrieval problem based on the ranked similarity score, which is obtained from the pairwise affinities between the probe and the dataset images. However, relying solely on the pairwise affinities of probe-gallery images, ignoring the underlying contextual information between the gallery images often leads to  an undesirable similarity ranking.  To tackle this,  several works  have been reported, which employ similarity diffusion to estimate a second order similarity that considers the intrinsic manifold structure of the given affinity matrix \cite{BaiBT17}, \cite{LoyLG13}, \cite{DonoserB13}, \cite{BaiZWBLT17}. Similarity diffusion is a process of exploiting the contextual information between all the gallery images to provide a  context sensitive similarity. Nevertheless, all these methods do not leverage the advantage of deep neural networks. Instead, they employ the similarity diffusion process as a post-processing step on the top of the DNN model. Aiming to improve the discriminative power of a DNN model, there have been recent works which incorporate a similarity diffusion process in an end-to-end manner \cite{ShenLXYCW18DGSRW},\cite{ShenLYCW18DSGNN},\cite{Chen0LSW18}. Following \cite{BertasiusTYS17}, which applies a random walk in an end-to-end fashion for solving semantic segmentation problem, authors in \cite{ShenLXYCW18DGSRW} proposed a group-shuffling random walk network for fully utilizing the affinity information between gallery images in both the training and testing phase. 
Also, the authors of \cite{ShenLYCW18DSGNN} proposed similarity-guided graph neural network (SGGNN) to exploit the relationship between several prob-gallery image similarities. 

However, most of the existing graph-based end-to-end learning methods apply the similarity diffusion without considering any constraint or attention mechanism to the specific query image. Due to  that the second order similarity  these methods yield is highly prone to noise. To tackle this problem, one possible mechanism could be to guide the similarity propagation by providing seed (or constraint) and let the optimization process estimate the optimal similarity between the seed and  nearest neighbors, while treating  the seed as our attention point. To formalize this idea, in this chapter,  we model person re-id problem as finding an internally coherent and externally incoherent constrained cluster in an end-to-end fashion. To this end, we adopt a graph and game theoretic method called constrained dominant sets in an end-to-end manner. To the best of our knowledge, we are the first ones to integrate the well known unsupervised clustering method called dominant sets in a DNN model. To summarize, the contributions of the proposed work are:
\begin{itemize}
	\item  For the very first time, the  dominant sets clustering method is integrated in a DNN and optimized  in end-to-end fashion.
	\item A one-to-one correspondence between person re-identification and constrained clustering problem is established.
	\item  State-of-the-art results are significantly improved.
	
\end{itemize}

The chapter is structured as follow. In section 2, we  review the related works. In section 3, we  discuss the proposed method with a brief introduction to dominant sets and constrained dominant sets. Finally, in section 4, we  provide an extensive experimental analysis on three different benchmark datasets.

\section{Related works}

Person re-id is one of the challenging computer vision tasks due to  the variation of illumination condition, backgrounds, pose and viewpoints. Most recent methods train DNN models with different learning objectives including verification, classification, and similarity learning \cite{ChengGZWZ16}, \cite{ZhaoLZW17}, \cite{VariorHW16}, \cite{AhmedJM15}, \cite{groupLoss}. For instance, verification network (V-Net) \cite{LiZXW14}, Figure \ref{fig:metaphor}(b), applies a binary classification of image-pair representation which is trained under the supervision of binary softmax loss. Learning accurate similarity and robust feature embedding has a vital role in the course of person re-identification process. Methods which integrate siamese network with contrastive loss are a typical showcase of deep similarity learning for person re-id \cite{ChenCZH17}. The optimization goal of these models is to estimate the minimum distance between the same person images, while maximizing the distance between images of different persons. However, these methods focus on the pairwise distance ignoring the contextual or relative distances. Different schemes have tried to overcome these shortcomings. In Figure \ref{fig:metaphor}(c),  triplet loss is exploited to enforce the correct order of relative distances among image triplets \cite{ChengGZWZ16}, \cite{DingLWC15}, \cite{ZhaoLZW17} . In Figure \ref{fig:metaphor}(d), Quadruplet loss \cite{ChenCZH17} leverages the advantage of both contrastive and triplet loss, thus it is able to maximize the intra-class similarity while minimizing the inter-class similarity. Emphasizing the fact that these methods entirely neglect the global structure of the embedding space, \cite{Chen0LSW18}, \cite{EleziTVP18}, \cite{ShenLXYCW18DGSRW},\cite{MeierEADS18}, \cite{ShenLYCW18DSGNN} proposed graph based end-to-end diffusion methods shown in Figure \ref{fig:metaphor}(e).

\textbf{Graph based end-to-end learning.} Graph-based methods have played a vital role in the rapid growth of computer vision applications in the past.   However, lately, the advent of deep convolutional neural networks and their tremendous achievements in the field has attracted great attention of researchers. Accordingly, researchers have made a significant effort to integrate, classical methods, in particular, graph theoretical methods, in end-to-end learning. Shen \et\cite{ShenLYCW18DSGNN} developed two constructions of deep convolutional networks on a graph, the first one is based upon hierarchical clustering of the domain, and the other one is based on the spectrum of graph Laplacian. Yan \et\cite{YanXL18} proposed a model of dynamic skeletons called Spatial-Temporal Graph Convolutional Networks (ST-GCN), which provides a capability to automatically learn both the spatial and temporal pattern of data. Bertasius \et \cite{BertasiusTYS17} designed a convolutional random walk (RWN), where by jointly optimizing the objective of pixelwise affinity and semantic segmentation they are able to address the problem of blobby boundary and spatially fragmented predictions. Likewise, \cite{ShenLXYCW18DGSRW} integrates random walk method in end-to-end learning to tackle person re-identification problem. In \cite{ShenLXYCW18DGSRW}, through the proposed deep random walk and the complementary feature grouping and group shuffling scheme, the authors  demonstrate that one can estimate a robust probe-gallery affinity. Unlike recent Graph neural network (GNN) methods \cite{ShenLYCW18DSGNN}, \cite{KipfW16}, \cite{ShenLXYCW18DGSRW}, \cite{Chen0LSW18}, Shen \et \cite{ShenLYCW18DSGNN} learn the edge weights by exploiting the training label supervision,  thus they are able to learn more accurate feature fusion weights for updating nodes feature.

\textbf{Recent applications of dominant sets.} Dominant sets (DS) clustering \cite{PavPel07} and its constraint variant constrained dominant sets (CDS) \cite{ZemeneP16} have been employed in several recent computer vision applications ranging from person tracking \cite{TesfayeZPP16}, \cite{TesfayeZPPS17}, geo-localization \cite{ZemeneTIPPS19}, image retrieval \cite{ZemeneAP16}, \cite{AlemuPelillo}, 3D object recognition \cite{WangPS17}, to Image segmentation and co-segmentation \cite{ZemAP19}, \cite{aslan2019weakly}. Zemene \et \cite{ZemeneP16}  presented CDS with its applications to interactive Image segmentation. Following, \cite{ZemAP19} uses CDS to tackle both image segmentation and co-segmentation in interactive and unsupervised setup. Wang \et \cite{WangPS17} recently used dominant sets clustering in a recursive manner to select representative images from a collection of images and applied a pooling operation on the refined images,  which survive at the recursive selection process. Nevertheless, {\em none of the above works have attempted to leverage the dominant sets algorithm in an end-to-end manner.}

In this work, unlike most of the existing graph-based DNN model, we propose a constrained clustering based scheme in an end-to-end fashion,  thereby,  leveraging the contextual information hidden in the relationship among person images. In addition, the proposed scheme significantly magnifies the inter-class variation of different person-images while reducing the intra-class variation of the same person-images. The big picture of our proposed method is depicted in Figure \ref{fig:metaphor}(f), as can be seen, the objective here is to find a coherent constrained cluster which incorporates the given probe image $P$.
\section{Our Approach}
In this work, we cast probe-gallery matching as optimizing a constrained clustering problem, where the probe image is treated as a constraint, while the positive images to the probe are taken as members of the constrained-cluster. Thereby, we integrate such clustering mechanism into a deep CNN to learn a robust  features through the leveraged contextual information. This is achieved by traversing through the global structure of the given graph to induce a compact set of images based on the given initial similarity(edge-weight).

\begin{figure*}[t]
	
	\begin{center}
		
		\includegraphics[width=0.9\linewidth ,trim=0cm 0cm 0cm 0cm,clip]{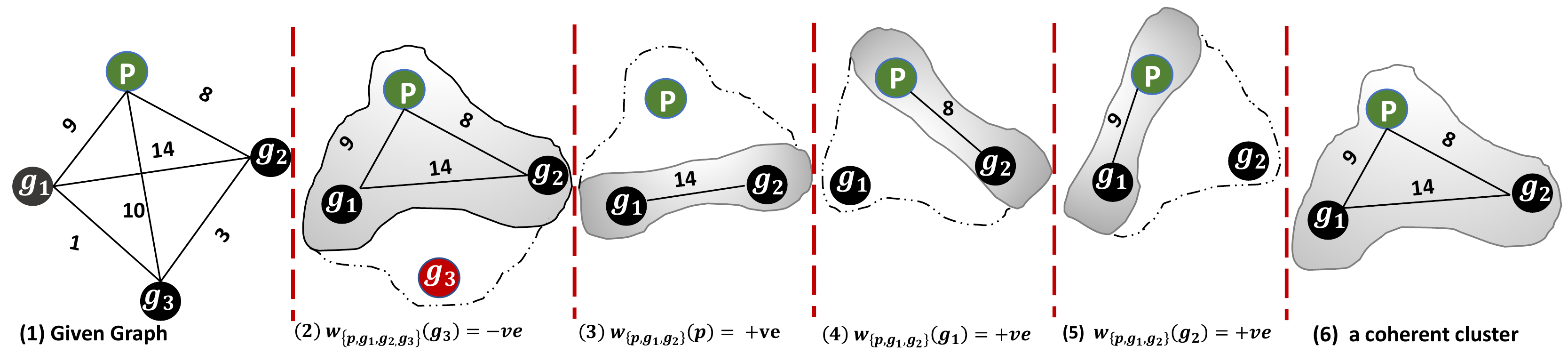}
	\end{center}
	\caption{Let $S= \{P,g_1, g_2, g_3\}$ comprises probe, $P,$ and gallery images $g_i$. As can be observed from the above toy example, the proposed method assess the contribution of each participant node $i \in S$ with respect to the subset $S \backslash i$. (1) shows graph G, showing the pairwise similarities of query-gallery images. (2-5) show the relative weight, $W_{\Gamma\}}(i)$
		( Equ. \ref{eqn:wdegree}), of each node with respect to the overall similarity between set $\Gamma \backslash i$ (shaded region) and $i$. 
		(2) shows that if the Node $\{g_3 \}$ is added with Node $\{ P, g_1, g_2\}$  it has a negative impact on the coherency of the cluster, since $W_{p,g_1,g_2, g_3}{(g_3)} < 0$. (3) shows that clustering $\{P \}$ with $\{g_1 \}$ and $\{ g_2\}$ has a positive contribution to the compactness of set $\{P, g_1, g_2\}.$ (4), similarly, shows the relative weight of $g_1,$ $W_{p,g_1,g_2}{(g_1)} > 0.$ (5) shows the relative weight of $g_2, W_{p,g_1,g_2}{(g_2)} > 0$. And, (6) is a coherent subset (dominant set cluster) extracted from the graph given in (1). 
	}
	\label{fig:exampler}
\end{figure*}

\begin{figure*}[t]
	
	\begin{center}
		
		\includegraphics[width=1\linewidth ,trim=0cm 0cm 0cm 0cm,clip]{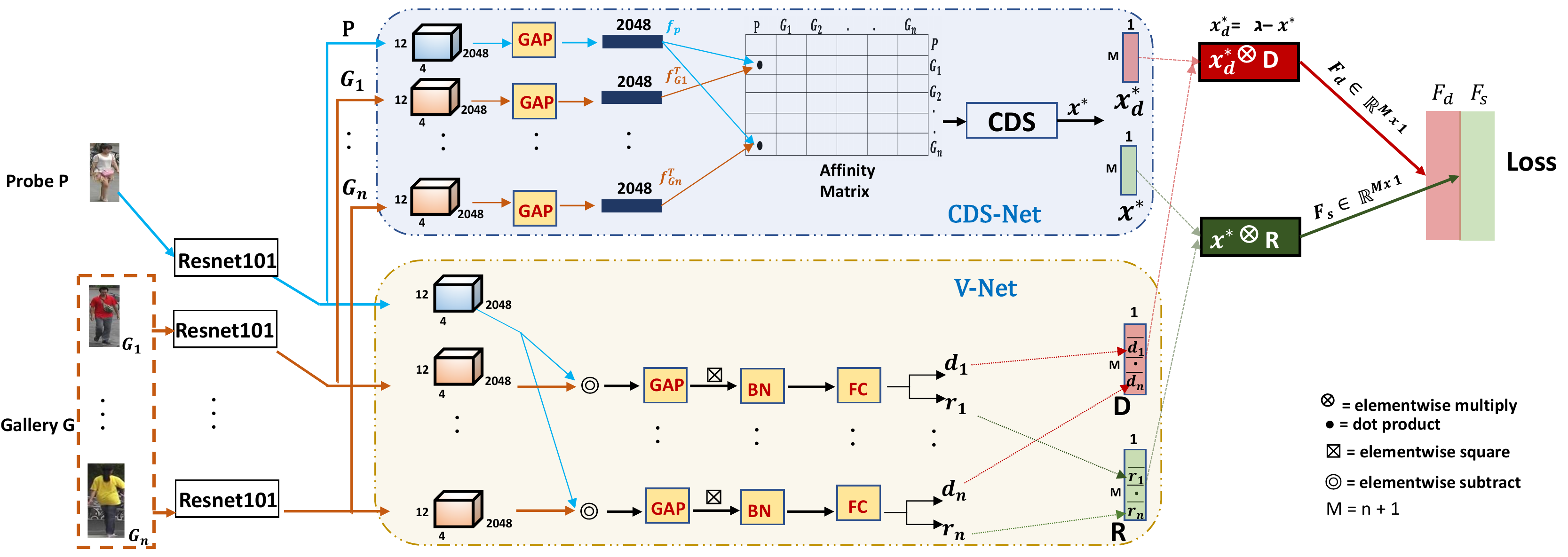}
	\end{center}
	\caption{Workflow of the proposed DCDS. Given n number of gallery images, $G,$ and probe image $P$, we first extract their Resent101 features right before the global average pooling (GAP) layer, which are then fed to CDS-Net (upper stream)  and V-Net (lower stream) branches. 
	In the CDS-branch, after applying GAP, we compute the similarity between $M^2$ pair of probe-gallery image features, $f_p$ and $f_{Gi}^T$ using their dot products, where $T$ denotes a transpose. Thereby, we obtain $M \times M$ affinity matrix. Then, we run CDS taking the probe image as a constraint to find the solution $x^* \in {\rm I\!R}^{M \times 1}$ (similarity), and the dissimilarity, $x^*_d,$ is computed as an additive inverse of the similarity $x^*.$ Likewise, in the lower stream we apply elementwise subtraction on $M$ pair of probe-gallery features.  This is followed by GAP, batch normalization (BN), and fully connected layer (FC) to obtain probe-gallery similarity score, $R \in {\rm I\!R}^{M \times 1},$ and probe-gallery dissimilarity score, $D \in {\rm I\!R}^{M \times 1}$. Afterward, we  elementwise multiply $x^*$  and $R,$ and $x^*_d$ and $D$, to find the final similarity, $F_s,$ and disimilarity, $F_d,$ scores, respectively. Finally, to find the prediction loss of our model, we apply a cross entropy loss, the ground truth ($G_t$) is given as $G_t  \in {\rm I\!R}^{M \times 1}$. }
	\label{fig:pipline}
\end{figure*}

\subsection{Modeling person re-id as a Dominant Set}
Recent methods \cite{Chen0LSW18}, \cite{BertasiusTYS17} have proposed different models, which leverage local and group similarity of images in an end-to-end manner. Authors in \cite{Chen0LSW18} define a group similarity which emphasizes the advantages of estimating a similarity of two images, by employing the dependencies among the whole set of images in a given group. In this work, we establish a natural connection between finding a robust probe-gallery similarity and constrained dominant sets. Let us first elaborate the intuitive concept of finding a coherent subset from a given set based on the global similarity of given images. For simplicity, we represent  person-images as vertices of graph $G,$ and their similarity as edge-weight $w_{ij}$. Given vertices $V,$ and $S \subseteq V$  be a non-empty subset of vertices and $i \in S$, average weighted degree of each  $i$ with regard to $S$ is given as $\vspace{-0.4cm}$  $$ \label{eq1} \phi_S(i,j)=a_{ij}-\frac{1}{|S|} \sum_{k \in S} a_{ik}~,
$$ where $\phi_S(i,j)$ measures the (relative) similarity between node $j$ and $i$, with respect to the average similarity between node $i$ and its neighbors in $S$. Note that $\phi_S(i,j)$ can be either positive or negative. Next, to each vertex $i \in S$ we assign a weight defined (recursively) as follows:
\vspace{-0.2cm}
\begin{equation}
\vspace{-0cm}
\label{eqn:wdegree}
w_S(i)=
\begin{cases}
1,&\text{if\quad $|S|=1$},\\
\sum_{j \in S \setminus \{i\}} \phi_{S \setminus \{i\}}(j,i)w_{S \setminus \{i\}}(j),&\text{otherwise}
\end{cases}
\end{equation}
where $w_{ij} (i) = w_{ij} (j) = a_{ij}$  for all $i, j  \in V( i \neq j)$. \\
Intuitively, $w_S(i)$ gives us a measure of the overall similarity between vertex $i$ and the vertices of $S\setminus \{i\}$, with respect to the overall similarity among the vertices in $S\setminus \{i\}$. Hence, a \textbf{positive} $w_S(i)$ indicates that adding $i$ into its neighbors in $S$ will raise the internal coherence of the set, whereas in the presence of a \textbf{negative} value we expect the overall coherence to  decline. In CDS, besides the additional feature, which allows us to incorporate a constraint element in the resulting cluster, all the characters of DS are inherited.
\subsubsection{A Set of a person images as a constrained cluster}
We cast person re-identification as finding a constrained cluster, where, elements of the cluster correspond to a set of same person images and the constraint refers to the probe image used to extract the corresponding cluster. As customary, let us consider  a given mini-batch with $M$ number of person-images, and each mini batch with $k$  person identities (ID), thus, each person-ID has $\Omega = M / k$ images in the given mini-batch. Note that, here, instead of a random sampling we design a custom sampler which samples $k$ number of person IDs in each mini-batch. Let $B = \{I_{p_1}^1, . . . I_{p_1}^{\Omega}, I_{p_2}^1, . . . I_{p_2}^{\Omega}, . . . I_{p_k}^1, . . . I_{pk}^{\Omega} \}$ refers to the set of images in a single mini-batch.  Each time when we consider image $I_{p_1}^1$ as a probe image $P$, images which belong to the same person id, $\{I_{p_1}^2, I_{p_1}^3 . . . I_{p_1}^k \},$ should be assigned a large membership score to be in that cluster. In contrast, the remaining images in the mini-batch should be assigned significantly smaller membership-score to be part of that cluster.
Note that our ultimate goal here is to find a constrained cluster which comprises all the images of the corresponding person given in that specific mini-batch. Thus, each participant in a given mini-batch is assigned a membership-score to be part of a cluster. Furthermore, the characteristics vector, which contains the membership scores of all participants is always a stochastic vector, meaning that  $\sum_{i =1 }^M z_i = 1,$ where $z_i$ denotes the membership score of each image in the cluster.

As can be seen from the toy example in Figure \ref{fig:exampler}, the initial pairwise similarities between the query and gallery images hold  valuable information, which define the relation of nodes in the given graph. However, it is not straightforward to redefine the initial pairwise similarities in a way which exploit the inter-images relationship. Dominant Sets (DS) overcome this problem with defining a  weight of each image $p, g_1, g_2, g_3$ with regard to subset $S\backslash i$ as depicted in Figure$\ref{fig:exampler}, (2-5),$ respectively. As can be observed from Figure \ref{fig:exampler}, adding node $g_3$ to cluster $S$  degrades the coherency of cluster $S = \{p, g_1, g_2, g_3\},$ whereas the relative similarity of the remaining images with respect to set $S = \{p, g_1, g_2\}$ has a positive impact on the coherency of the cluster. 
It is evident that the illustration in Figure \ref{fig:exampler} verifies that the proposed DCDS (Deep Constrained Dominant Set) could easily measure the contribution of each node in the graph and utilize it in an end-to-end learning process. Thereby, unlike a siamese, triplet and quadruplet based contrastive methods, DCDS consider the whole set of images in the mini-batch to measure the similarity of image pairs and enhance the learning process.

\subsection{CDS Based End-to-end Learning}
In this section, we discuss the integration of CDS in end-to-end learning. We adopt a siamese based Resent101, with a novel  verification loss to find probe-gallery similarity, $R$, and dissimilarity, $D$ scores. 
As can be seen from Figure \ref{fig:pipline}, we  have two main branches:  CDS network branch (CDS-Net) and verification network branch (V-Net). 
In the CDS-Net, the elements of pairwise affinity matrix are computed first as a dot product of the global pooling feature of a pair of images. Afterward, the replicator dynamics \cite{Wei95} is applied, which is a discrete time solver of the parametrized quadratic program, Equ. \ref{eqn:parQP2}, whose solution corresponds to the CDS. Thus, assuming that  there are $M$ images in the given mini-batch, the replicator dynamics, Equ. \ref{eqn:Replicator}, is  recursively applied $M$ times taking each image in the mini-batch as a constraint.  Given graph $G = (V, E,w)$ and its corresponding adjacency matrix $A \in R^{M \times M},$ and probe $P \subseteq V.$ First, a proper modification of the affinity matrix $A$ is applied by setting parameter $-\alpha$ to the diagonal corresponding to the subset $V \backslash P$ and zero to the diagonal corresponding to the constraint image $P$. 
Next, the modified adjacency matrix, $B,$ is feed to the Replicator dynamics, by initiating the dynamics with a characteristic vector of uniform distribution $x^{t_0}$, such  that  initially all the images in the mini-batch are assigned equal membership probability to be part of the cluster. Then, to find a constrained cluster  a parametrized quadratic program is defined as:
\begin{equation}
\label{eqn:parQP2}
\begin{array}{ll}
\text{maximize }  &  f_P^\alpha(\x)^i = \x' B \x \quad  where,  B = A - \alpha \hat I_p. \\ 
\text{subject to} &  \mathbf{x} \in \Delta
\end{array}
\end{equation}

The solution, $ \x_i^*,$ of $ f_P^\alpha(\x)^i$  is a characteristics vector which indicates the probability of each gallery image to be included in a cluster, containing the probe image $P^i$. Thus, once we obtain the CDS, $\x_i^* = [z^i_{g_1}, z^i_{g_2} . . . z^i_{g_M}],$  for each probe image, we store each solution $ \x_i^*$, in $Y \in {\rm I\!R}^{M \times M},$ as 
$$
Y = 
\begin{pmatrix} 
 \x_i^* \\ 
\vdots \\
\x_M^*\\ 
\end{pmatrix} =
\begin{pmatrix} 
z_{g_1}^{1 }&z_{g_2}^{1 }& \cdots  & z_{g_M}^{1} \\ 
\vdots && ~~ \ddots ~~ & \vdots &\\
z_{g_1}^{M }&z_{g_2}^{M }& \cdots  & z_{g_M}^{M}\\ 
\end{pmatrix}.
$$ 
Likewise, for each probe, $P^i,$ we store the probe-gallery similarity, R, and dissimilarity, D, obtained from the V-Net (shown in Figure \ref{fig:pipline}) in $S'$ and $D'$ as, $S' = [R^1, R^2, . . . R^M]$ and $D' = [D^1, D^2, . . . D^M].$
Next, we fuse the similarity obtained from the CDS branch with the similarity  from the V-Net as 
\begin{equation}
\label{eqn:fus}
\begin{array}{ll}
F_s = \beta(Y) \otimes (1 - \beta)(S'),  \\ 
F_d = \beta(Y_d) \otimes (1 - \beta)(D'), \quad where, \quad Y_d = \delta - Y
\end{array}
\end{equation}
$\delta$ is empirically set to 0.3. We then vectorize $F_s$ and $F_d$ into ${\rm I\!R}^{(M^2 \times 2)},$ where, the first column stores the dissimilarity score, while the second column stores the similarity score. Afterward, we simply apply cross entropy loss to find the prediction loss.  
The intriguing feature of our model is that it does not need any custom optimization technique, it can be end-to-end optimized through a standard back-propagation algorithm. Note that, Figure \ref{fig:pipline} illustrates the case of a single probe-gallery, whereas Equ. \ref{eqn:fus} shows the solution of $M$ probe images in a given mini-batch.
\begin{figure}[t]

	\begin{center}
		
		\includegraphics[width=1\linewidth ,trim=0cm 0cm 0cm 0cm,clip]{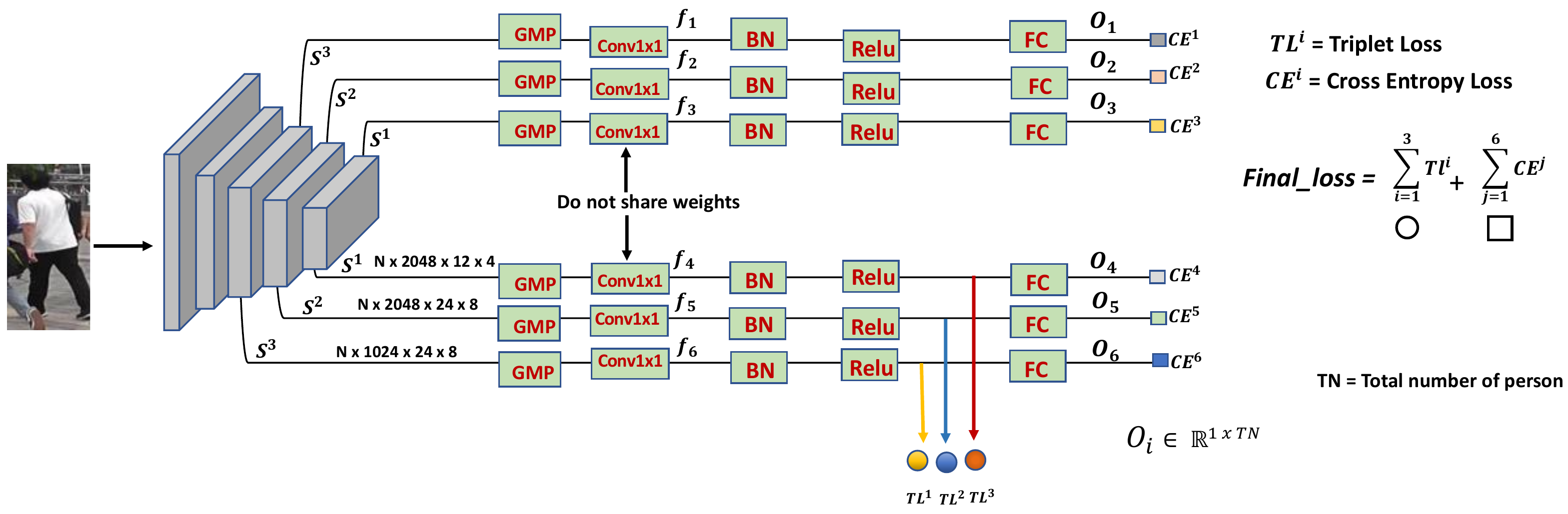}
	\end{center}
	
	\caption{Illustrates the auxiliary net, which consists of two branches which are jointly trained. We first  use  features at  different layers, $S_1, S_2, S_3,$ and then feed these to Global Maxpooling (GMP), Conv, BN, Relu and FC layers for further encoding. 
	We then compute triplet losses employing the features from the lower three streams after Relu, shown  by yellow, blue, and red circles. Next, after the final FC layer, we compute the cross-entropy loss for each of the six different outputs, $O_i,$ from the upper and lower stream  shown by distinct colored-boxes. Note that even if the upper and lower stream apply the same operations, on $S_1, S_2$ and $S_3,$ they do not share the weights; thus the encoding is different. We finally compute the final loss as the sum of the average of the triplet and cross entropy losses.}
	\label{fig:Auxnet}
\end{figure}

\subsection{Auxiliary Net}

\begin{figure}
	
	\begin{center}
		
		\includegraphics[width=1\linewidth ,trim=0cm 0cm 0cm 0cm,clip]{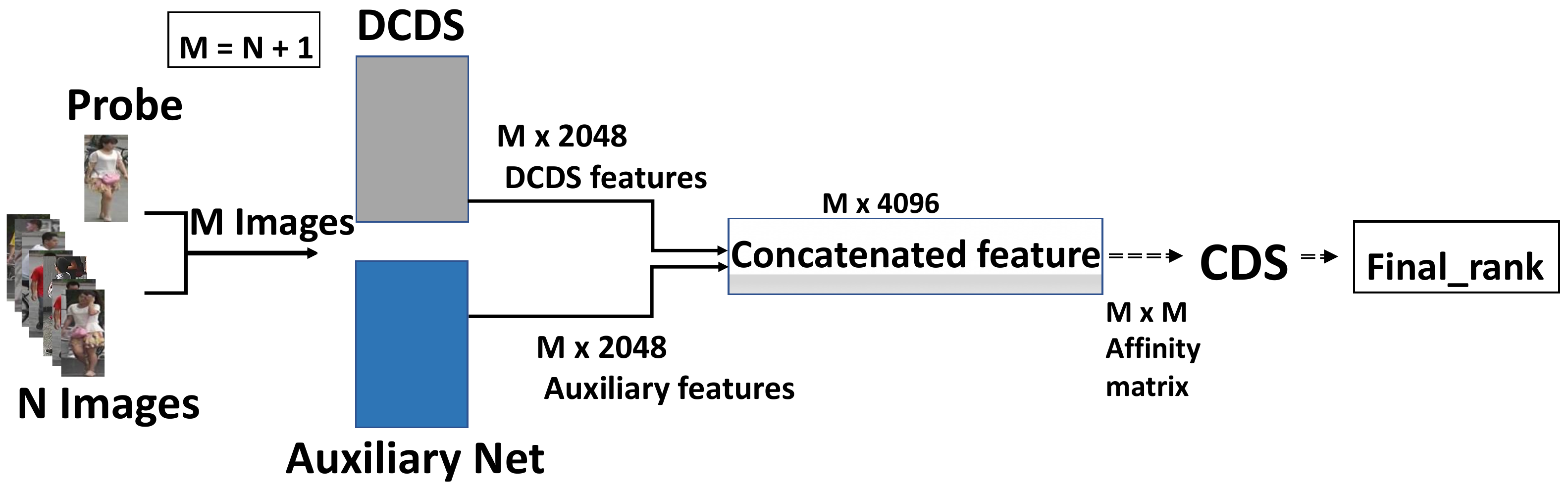}
	\end{center}
	\caption{During testing, given a probe and gallery images, we extract DCDS and auxiliary features and concatenate them to find a single vector. Afterward, we build M x M affinity matrix and run CDS with constraint expansion mechanism to find the final probe-gallery similarity rank.}
	\label{fig:Testphase}
\end{figure}

In this work, we integrate an auxiliary net to further improve the performance of our model. The auxiliary net is trained based on the multi-scale prediction of Resnet50 \cite{HeZRS16}. It is a simple yet effective architecture, whereby we can easily  compute both triplet and cross entropy loss of different layers of Resnet50 \cite{HeZRS16}, hence further enhancing the learning capability. Consequently, we compute the average of both losses to find the final loss. As can be observed from Figure \ref{fig:Auxnet}, we employ three features at different layers from Resnet50 $conv5\_x$ Layer, and then we fed these three features to the subsequent layers, MP, Conv, BN, and FC layers. Next, we compute triplet and cross entropy loss for each feature which comes from the Relu and FC layers, respectively. During testing phase we concatenate the features that come from the DCDS and Auxiliary Net to find 4096 dimensional feature. We then apply CDS to find the final ranking$\_$score, (See Figure \ref{fig:Testphase}).

\begin{figure}[t]

	\begin{center}
		
		\includegraphics[width=1\linewidth ,trim=0cm 5cm 0cm 0cm,clip]{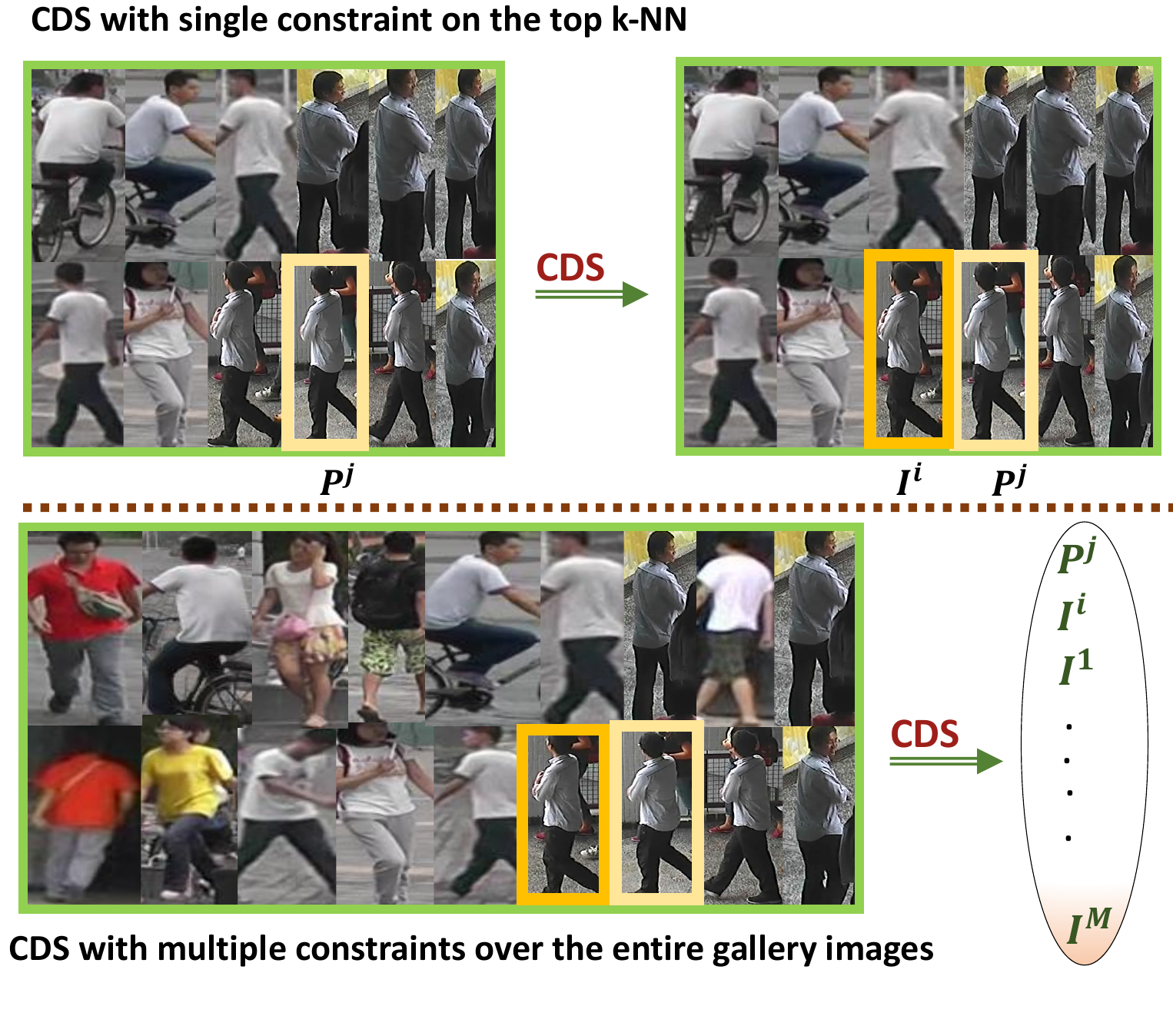}
	\end{center}
	\caption{Given a constraint (probe-image) $P^j,$ we first collect k-NNs to the probe-image, based on the pairwise similarities. Subsequently, we run CDS on the graph of the k-NN. Then, based on the cluster membership score obtained, we choose image $I^{i }$,  with the highest membership score and re-run CDS, considering $P^j$ and $I^i$ as constraints, over the graph of the all set of  images, $I^M,$ in the minibatch. Afterward, we consider the solution as our final rank.}
	\label{fig:Conexpansion}
\end{figure}

\begin{table}[h]
	
	\begin{center}
		
		\begin{tabular}{l|c|c|c} 
			\hline
			Methods & mAP& rank-1 & rank-5\\
			
			\hline\hline
			SGGNN \cite{ShenLYCW18DSGNN} ECCV18& 82.8&92.3& 96.1    \\
			DKPM \cite{ShenXLYW18DKPM} CVPR18&75.3&90.1& 96.7    \\
			DGSRW \cite{ShenLXYCW18DGSRW} CVPR18& 82.5&92.7& 96.9    \\
			GCSL \cite{Chen0LSW18} CVPR18&  81.6&93.5& - \\  
			CPC \cite{WangCW018} CVPR18& 69.48&83.7& -    \\		  
			MLFN \cite{ChangHX18} CVPR18&  74.3&90.0& - \\   
			HA-CNN \cite{LiZG18} CVPR18& 75.7 &91.2& -   \\		 
			PA \cite{SuhWTML18} ECCV18&  74.5& 88.8& 95.6 \\ 
			HSP \cite{KalayehBGKS18} CVPR18&  83.3 & 93.6& 97.5 \\
			{\bf Ours}  & \textbf{85.8}& \textbf{94.81}& \textbf{98.1} \\
			\hline\hline
			$RA_{w/ RR}$ \cite{WangWYZCLHHW18} CVPR18& 86.7&90.9& -    \\	
			$PA _{w/ RR}$ \cite{SuhWTML18} ECCV18&  89.9& 93.4& 96.4 \\ 
			$HSP_{w/ RR}$ \cite{KalayehBGKS18} CVPR18 &  90.9& 94.6& 96.8 \\	
			{\bf Ours$_{w/ RR}$} & \textbf{93.3}& \textbf{95.4}&  \textbf{98.3} \\
			
			\hline
		\end{tabular}
		
	\end{center}
	
	\caption{A comparison of the proposed method with state-of-the-art methods on Market1501 dataset. Upper block, without re-ranking methods. Lower block, with re-ranking method, $w/ RR$, \cite{ZhongZCL17}.}
	\label{table:market}
\end{table}

\begin{table*}[h]
	
	\begin{center}
		
		\begin{tabular}{p{2cm}|p{0.8cm}|c|c|c|c|c|c|c} 
			\hline
			\multirow{2}{*}{Methods}&\multicolumn{2}{c}{Market1501}&&\multicolumn{1}{c}{CUHK03}&&\multicolumn{2}{c}{DukeMTMC-reID}\\
			
			& mAP& rank-1 & rank-5&rank-1&rank-5&mAP& rank-1&rank-5\\
			
			\hline\hline
			
			Baseline SD & 72.2& 86.5&  94.0&87.1&94.3&61.1&77.6&87.3 \\ \hline
			Baseline MD & 74.3& 87.5&  95.3 &87.7&95.2&62.3&79.1&88.8\\ \hline
			DCDS (SD ) & 81.4& 93.3&  97.6 &93.1&98.8&69.1&83.3&89.0\\ \hline
			DCDS (MD) & 82.3& 93.7&  98.0 &93.9&98.9&70.5&84.0&90.3\\ \hline
			Ours (SD + Auxil Net) & 83.0& 93.9&  98.2 &95.4&99.0&74.4&85.6&93.7\\ \hline
			{\bf Ours (MD + Auxil Net)} & {\bf 85.8}& {\bf 94.1}&  {\bf 98.1} &{\bf 95.8} & {\bf 99.1}& {\bf 75.5}& {\bf 86.1} & {\bf 93.2}\\
			
			\hline
		\end{tabular}
	\end{center}
	
	\caption{Ablation studies on the proposed method. SD and MD respectively refer to the method  trained on single and multiple-aggregated datasets. Baseline is the proposed method without CDS branch.}
	\label{table:ABLATION}
\end{table*}

\begin{table}
	
	\begin{center}
		
		\begin{tabular}{l|c|c} 
			\hline
			Methods & rank-1 & rank-5\\
			
			\hline\hline
			SGGNN \cite{ShenLYCW18DSGNN} ECCV18& 95.3 &99.1   \\
			DKPM \cite{ShenXLYW18DKPM} CVPR18& 91.1 & 98.3   \\
			DGSRW \cite{ShenLXYCW18DGSRW} CVPR18& 94.9 & 98.7   \\
			GCSL \cite{Chen0LSW18} CVPR18& 90.2& 98.5\\  
			MLFN \cite{ChangHX18} CVPR18&  89.2&- \\ 
			CPC \cite{WangCW018} CVPR18& 88.1 & -   \\		  
			PA \cite{SuhWTML18} ECCV18 & 88.0& 97.6 \\ 
			HSP \cite{KalayehBGKS18} CVPR18& 94.28& 99.04 \\
			{\bf Ours}  & {\bf 95.8} &  {\bf 99.1} \\
			\hline
		\end{tabular}
	\end{center}
	
	\caption{A comparison of the proposed method with state-of-the-art methods on CUHK03 dataset.}
	\label{table:compCUHK}
\end{table}

\begin{table}[h]
	
	\begin{center}
		
		\begin{tabular}{l|c|c|c} 
			\hline
			Methods & mAP& rank-1 & rank-5\\
			
			\hline\hline
			SGGNN \cite{ShenLYCW18DSGNN} ECCV18& 68.2 &81.1& 88.4  \\
			DKPM \cite{ShenXLYW18DKPM} CVPR18& 63.2 &80.3& 89.5   \\
			DGSRW \cite{ShenLXYCW18DGSRW} CVPR18& 66.4&80.7& 88.5    \\
			GCSL \cite{Chen0LSW18} CVPR18&  69.5&84.9& - \\  
			CPC \cite{WangCW018} CVPR18& 59.49 &76.44& -  \\		  
			MLFN \cite{ChangHX18} CVPR18&  62.8&81.0& - \\   
			RAPR \cite{WangWYZCLHHW18} CVPR18& 80.0&84.4& -    \\	
			PA \cite{SuhWTML18} ECCV18&  64.2& 82.1& 90.2 \\ 
			HSP \cite{KalayehBGKS18} CVPR18 &  73.3& 85.9& 92.9 \\
			Ours & \textbf{75.5} & \textbf{87.5}&  - \\
			\hline\hline
			$PA _{w/ RR}$ \cite{SuhWTML18} ECCV18&  83.9 & 88.3& 93.1\\ 
			$HSP_{w/ RR}$ \cite{KalayehBGKS18} CVPR18 &  84.99& 88.9& 94.27 \\
			{\bf Ours} $_{w/ RR}$ & \textbf{86.1}& 88.5&  - \\
			
			\hline
		\end{tabular}
	\end{center}
	
	\caption{A comparison of the proposed method with state-of-the-art methods on DukeMTMC-reID dataset.Upper block, without re-ranking methods. Lower block, with re-ranking method,$w/ RR$, \cite{ZhongZCL17}.}
	\label{table:compDuke}
\end{table}

\begin{figure}[t]
	
	\begin{center}
		
		\includegraphics[width=1\linewidth ,trim=0cm 0cm 0cm 0cm,clip]{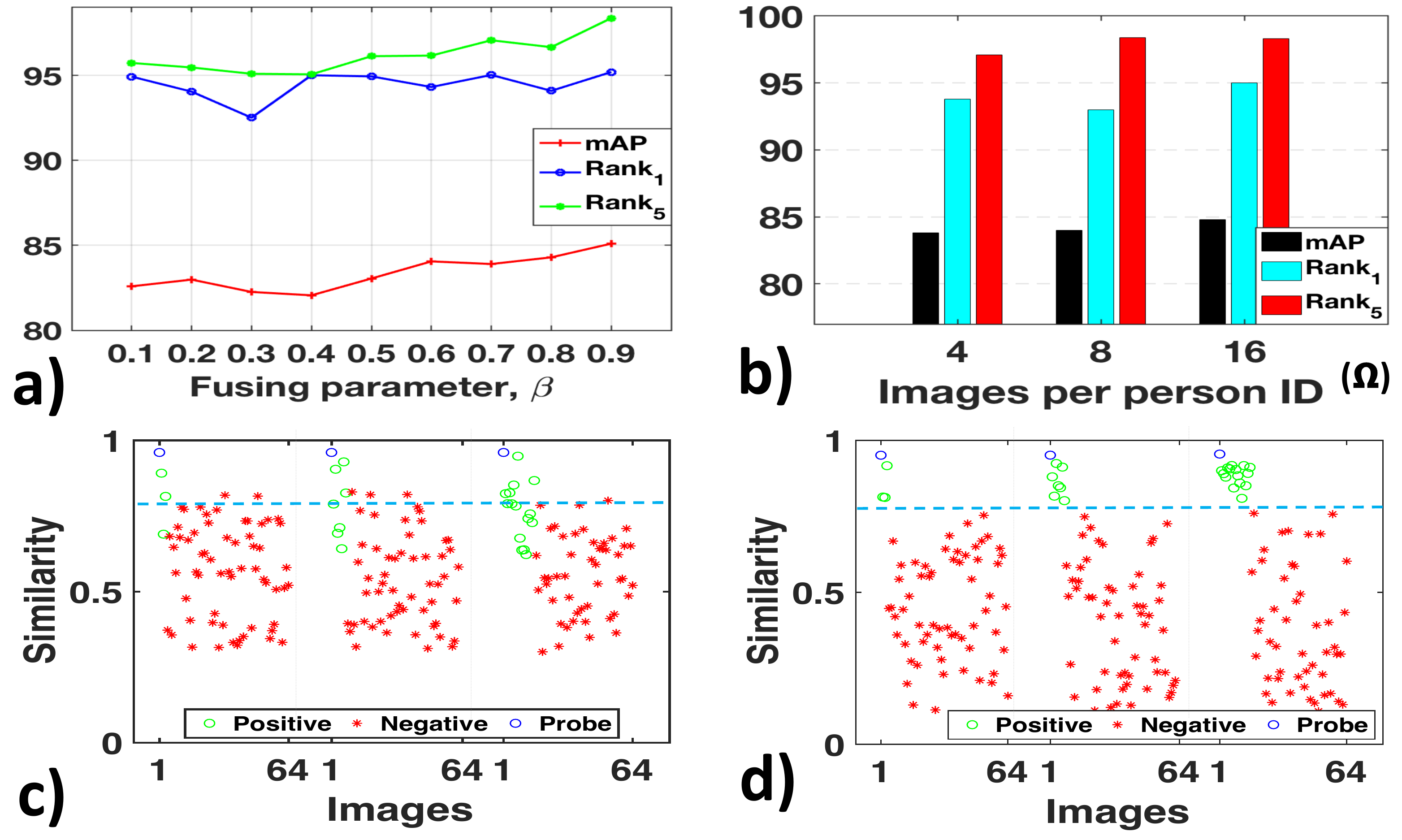}
	\end{center}
	
	\caption{Illustrates different experimental analysis performed on Market1501 dataset. a) shows the impact of fusing parameter $\beta$ in Equ. \ref {eqn:fus}. b) shows the performance of our model with varying the number of images per person in a given batch. c) and d) illustrate the similarity between the probe and gallery images obtained from the baseline and DCDS method, respectively. It can be observed that the baseline method has given larger similarity values for false positive samples (red asterisks above the blue dashed-line) and smaller similarity values for false negative samples (green circles below the blue dashed- line). On the other hand, the proposed DCDS has efficiently assigned the appropriate similarity scores to the true positive and negative samples. 
	}
	\label{fig:paramAnal}
\end{figure}

\begin{figure}
	\begin{center}
		
		\includegraphics[width=1\linewidth ,trim=0cm 0cm 0cm 0cm,clip]{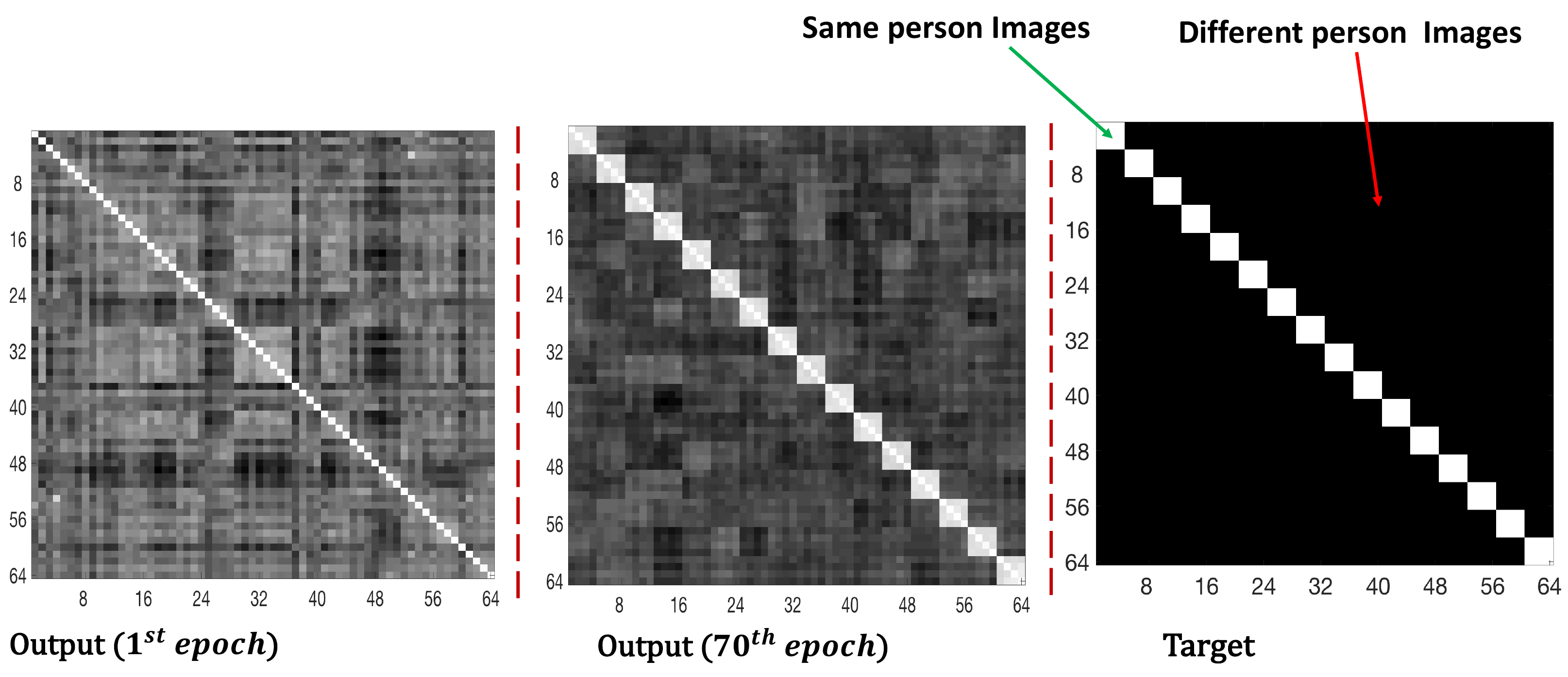}
	\end{center}
	
	\caption{On the right hand side, the target matrix is shown.
		There are total 16 persons in the mini-batch and 4 images per ID ($\Omega$ = 4),  batch size = 64. In the target matrix, the white-blocks represent the similarity between the same person-images in the mini-batch, whereas the black-blocks of the matrix define the dissimilarities between different person images. In the similarity matrix shown left ( after one epoch) and middle (after $70^{th}$ epochs) each row of the output matrix denotes the fused similarity obtained from the CDS-Net and V-Net, per Equ. (6) in the main manuscript. Thus, we optimize our model until we obtain an output with a similar distribution of the target matrix. As can be seen, our model has effectively learned and gives a similarity matrix (shown in the middle) which is closer to the target matrix. }
	\label{fig:metaphore}
	
\end{figure}

\subsection{Constraint Expansion During Testing}
We propose a new scheme (illustrated in Figure \ref{fig:Conexpansion}) to expand the number of constraints in order to guide the similarity propagation during the testing phase. Given an affinity matrix, which is constructed using the features obtained from the concatenated features (shown in Figure \ref{fig:Testphase}), we first collect k-NNs of the probe image. Then, we run CDS on the graph of the NNs. Next, from the resulting constrained cluster, we select the one with the highest membership score, which is used as a constraint in the subsequent step. We then use multiple-constraints and run CDS.

\begin{figure}
	\begin{center}
		
		\includegraphics[width=1\linewidth ,trim=0cm 0cm 0cm 0cm,clip]{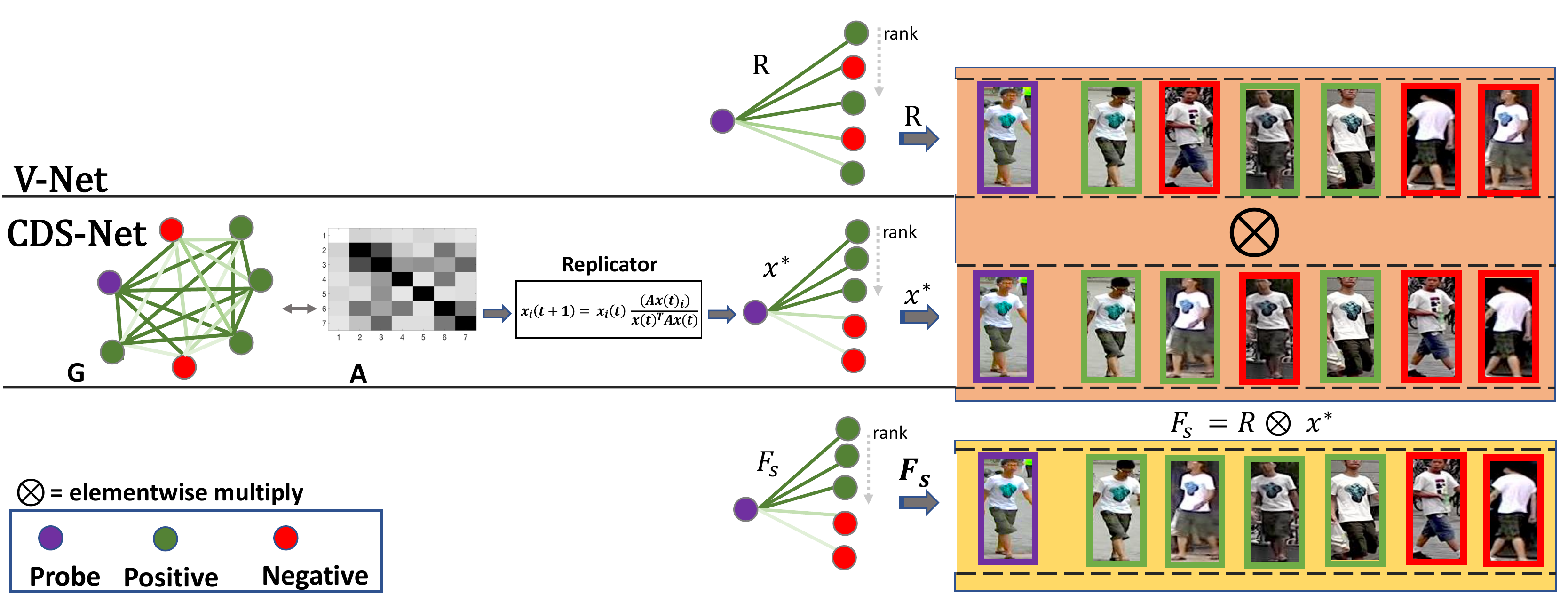}
	\end{center}
	
	\caption{Exemplar results obtained as a result of the similarity fusion between the V-Net and CDS-Net. The Upper-row shows the probe and gallery similarity (R) obtained from the V-Net, where the green circles show persons similar to the probe (shown by purple-circle), while the red circles denote persons different from the probe image. Middle-row shows the workflow in CDS-Net. First, graph G is formed using the similarity obtained from the dot products. We then construct the modified affinity matrix $B$, followed by application of replicator dynamics on $B$ to obtain the probe gallery similarity ($X^*$). Finally, We elementwise multiply $X^*$ and $R$ to find the final probe-gallery similarity ($F_s$), shown in the third row. The intensity of the edges in, $G,$ $R,$ $x^*,$ and $F_s$ define the similarity value, where the bold ones denote larger similarity values, whereas the pale-edges depict smaller similarity values.}
	\label{fig:exemplarResult}
	
\end{figure}

\begin{table}
	
	\begin{center}
		
		\begin{tabular}{l|l|c} 
			\hline
			
			&\multicolumn{2}{l}{\small {\bf Train} on Duke, CUHK03 $\rightarrow$ {\bf Test} on Market1501}\\
			\hline
			Methods & mAP & rank-1\\
			
			\hline\hline
			PUL \cite{FanZYY18}  & 20.5 &45.5   \\
			Ours   & \textbf{24.5} & \textbf{51.3}   \\
			\hline
		\end{tabular}
	\end{center}
	
	\caption{A comparison of the proposed method with PUL \cite{FanZYY18} on Market1501 dataset.}
	\label{table:compCUHK}
	
\end{table}
\begin{figure}
	\begin{center}
		
		\includegraphics[width=1\linewidth ,trim=0cm 0cm 0cm 0cm,clip]{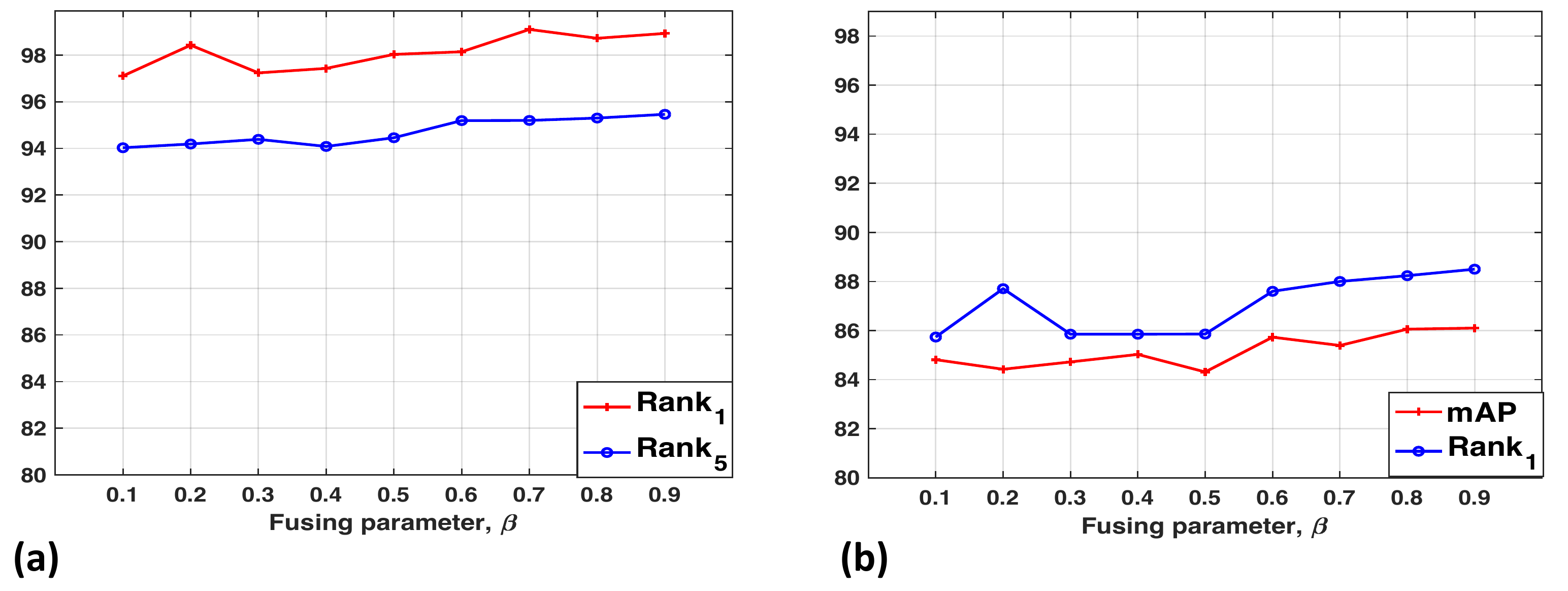}
	\end{center}
	
	\caption{Performance of our model with respect to fusing parameter $\beta$, on (a) CUHK03,  and   (b) DukeMTMC-reID, datasets.}
	\label{fig:fusingpar}
	
\end{figure}

\begin{figure}
	\vspace{-0.3cm}
	
	\begin{center}
		
		\includegraphics[width=1\linewidth ,trim=0cm 0cm 0cm 0cm,clip]{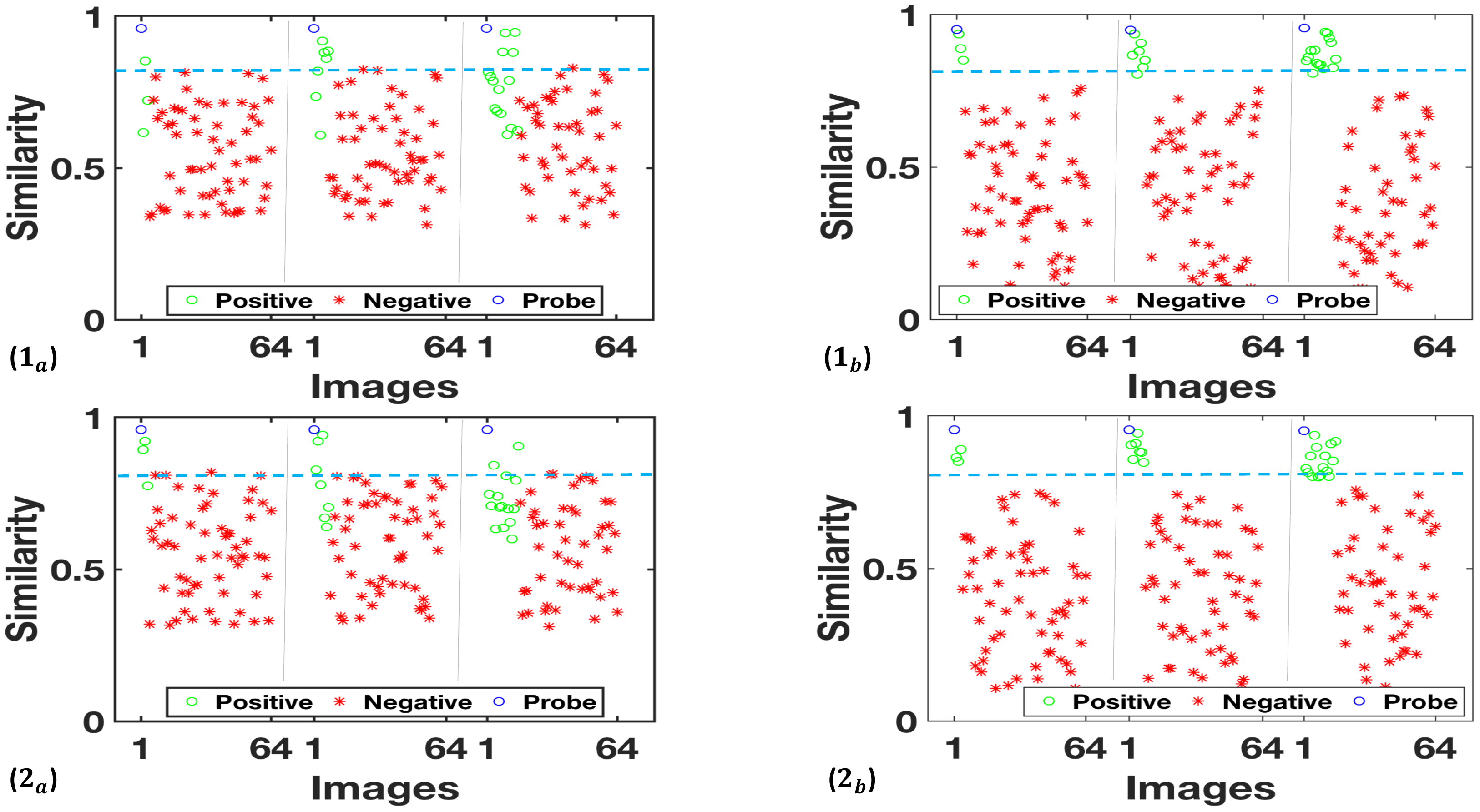}
	\end{center}
	
	\caption{Shows experimental analysis performed on CUHK03 $(1_{a, b}),$ and DukeMTMC-reID $(2_{a, b})$ datasets. $1_a, 2_a$ and $1_b, 2_b$ illustrate the similarity between the probe-gallery images obtained from the baseline and DCDS method, respectively. It can be observed that the baseline method has assigned larger similarity values for false positive samples (red asterisks above the blue dashed-line) and smaller similarity values for false negative samples (green circles below the blue dashed-line). On the other hand, the proposed DCDS has efficiently assigned the appropriate similarity scores to the true positive and negative samples. Note that, for better visibility, we have randomly assigned a large (close to 1) self-similarity value to the probe (blue-circle).}
	\label{fig:SimilDist}
	
\end{figure}

\section{Experiments}
To validate the performance of our method we have conducted several experiments on three publicly available benchmark datasets, namely CUHK03 \cite{LiZXW14}, Market1501 \cite{ZhengSTWWT15}, and DukeMTMC-reID \cite{ZhengZY17}.

\subsection{Datasets and evaluation metrics}
\textbf{Datasets:} CUHK03 \cite{LiZXW14} dataset comprises 14,097 manually and automatically cropped images of 1,467 identities, which are captured by two cameras on campus; in our experiments, we have used manually annotated images. Market1501 dataset \cite{ZhengSTWWT15} contains 32,668 images which are split into 12, 936 and 19,732 images as training and testing set, respectively. Market1501 dataset has totally 1501 identities which are captured by five high-resolution and one low-resolution cameras, the training and testing sets have 751 and 750 identities respectively. To obtain the person bounding boxes, Deformable part Model (DPM) \cite{FelzenszwalbGMR10} is utilized. DukeMTMC-reID  is generated from a tracking dataset called DukeMTMC. DukeMTMC is captured by 8 high-resolution cameras, and person-bounding box is manually cropped; it is organized as 16,522 images of 702 person for training and 18, 363 images of 702 person for testing.\\

In multiple dataset (MD) setup, we first train our model on eight datasets: CUHK03 \cite{LiZXW14}, CUHK01 \cite{LiZW12}, Market1501 \cite{ZhengSTWWT15}, DukeMTMC-reID \cite{ZhengZY17}, Viper \cite{dgray}, MSMT17 \cite{WeiZ0018}, GRID \cite{LoyXG09}, and ILIDS \cite{ZhengGX09}. Next, we fine-tune and evaluate on each of   CUHK03 \cite{LiZXW14}, Market1501 \cite{ZhengSTWWT15}, and DukeMTMC-reID \cite{ZhengZY17} datasets.

\textbf{Evaluation Metrics:} Following the recent person re-id methods, we  use mean average precision (mAP) as suggested in \cite{ZhengSTWWT15}, and Cumulated Matching Characteristics (CMC) curves to evaluate the performance of our model. Furthermore, all the experiments are conducted using the standard single query setting \cite{ZhengSTWWT15}. 


\subsection{Implementation Details}
We implement DCDS based on Resnet101 \cite{HeZRS16} architecture, which is pretrained on imagenet dataset. We adopt the training strategy of Kalayeh \et \cite{KalayehBGKS18}, and aggregate eight different person re-id benchmark dataset to train our model. In total, the merged dataset contains 89,091 images, which comprises 4937 person-ID (detail of the eight datasets is given in the supplementary material). We first train our model using the merged dataset (denoted as multi-dataset (MD)) for 150 epochs and fine-tune it with CUHK03, Market1501, and DukeMTMC-reID dataset. To train our model using the merged dataset, we set image resolution to 450 $\times$ 150. Subsequently, for fine-tuning the model we set image resolution to 384 $\times$ 128. Mini-batch size is set to 64, each mini-batch has 16 person-ID and each person-ID has 4 images. We also experiment only using a single dataset for training and testing, denoted as single-dataset (SD). For data augmentation, we apply random horizontal flipping and random erasing \cite{Zhong}. For optimization we use Adam, we initially set the learning rate to 0.0001, and drop it by 0.1 in every 40 epochs. The fusing parameter in Equ. \ref{eqn:fus}, $\beta$, is set to 0.9.

\subsection{Results on Market1501 Datasets}
As can be seen from Table \ref{table:market}, on Market dataset, our proposed method improves state-of-the-art method \cite{KalayehBGKS18}
by $2.5 \%, 1.21 \%,$ and $0.6 \%$ in mAP, rank-1 and rank-5 scores, respectively. Moreover, comparing to state-of-the-art graph-based DNN method, SGGNN \cite{ShenLYCW18DSGNN}, the improvement margins are $3 \%, 2.5 \%, $ and $2 \%$ in mAP, rank-1, and rank-5 score, respectively. Thus, our framework has significantly demonstrated its benefits over state-of-the-art graph-based DNN models. To further improve the result we have adapted a re-ranking scheme \cite{ZhongZCL17}, and we compare our method with state-of-the art methods which use a re-ranking method as a post-processing. As it can be seen from Table \ref*{table:market}, our method has gain mAP of $2.2\%$ over HSP \cite{KalayehBGKS18}, and 10.5 $\%$ over SGGNN\cite{ShenLYCW18DSGNN}, 10.8 $\%$ over DGSRW. 

\subsection{Results on CUHK03 Datasets}
Table \ref{table:compCUHK} shows the performance of our method on CUHK03 dataset. Since most of the Graph-based DNN models report their result on the standard protocol \cite{LiZXW14}, we have experimented on the standard evaluation protocol, to make fair comparison. As can be observed from Table \ref{table:compCUHK}, our method gain a marginal improvement in the mAP. Using a reranking method \cite{ZhongZCL17}, we have reported a competitive result in all evaluation metrics.

\subsection{Results on DukeMTMC-reID Dataset}
Likewise, in DukeMTMC-reID dataset, the improvements of our proposed method is noticeable.
Our method has surpassed state-of-the-art method \cite{KalayehBGKS18} by $1.7\% / 1.6 \% $ in mAP/rank-1 scores. Moreover, comparing to state-of-the-art graph-based DNN, our method outperforms DGSRW \cite{ShenLXYCW18DGSRW},  SGGNN \cite{ShenLYCW18DSGNN} and GCSL \cite{Chen0LSW18} by $9.1 \%, 7.3 \%,$ and $6\%$ in mAP, respectively. 


\subsection{Ablation Study}
To investigate the impact of each component in our architecture, we have performed an ablation study. Thus, we have reported the contributions of each module in Table \ref{table:ABLATION}. To make a fair comparison with the baseline and graph-based DNN models, the ablations study is conducted in a single-dataset (SD) setup.
\textbf{Improvements over the Baseline.} As our main contribution is the DCDS, we examine its impact over the baseline method. The baseline method refers to the lower branch of our architecture that incorporates the verification network, which has also been utilized in \cite{ShenXLYW18DKPM}, \cite{ShenLXYCW18DGSRW}, \cite{ShenLYCW18DSGNN}. On Market1501 dataset, DCDS provides improvements of $9.2 \%, 6.8 \%$ and $3.6 \%$ in mAP, rank-1, and rank-5 scores, respectively, over the baseline method; whereas in DukeMTMC-reID dataset the proposed DCDS improves the baseline method by $8.0 \%, 5.5 \%$ and $1.7 \%$ in mAP, rank-1, and rank-5 scores, respectively.\\
\textbf{Comparison with graph-based deep models.} We  compare our method with recent graph-based-deep models, which adapt similar baseline method as ours, such as \cite{ShenLXYCW18DGSRW},\cite{ShenLYCW18DSGNN}. As a result, on DukeMTMC-reID dataset our method surpass \cite{ShenLXYCW18DGSRW} by $9.1 \% /6.8 \%,$ and \cite{ShenLYCW18DSGNN} by 17.9 $\%$ / 7.4 $\%$ in mAP / rank-1 scores. In light of this, We can   conclude that incorporating a constrained-clustering  mechanism in end-to-end learning has a significant benefit on finding a robust similarity ranking. In addition, experimental findings demonstrate the superiority of DCDS over existing graph-based DNN models.\\
\textbf{Parameter analysis.} 
Experimental results by varying several parameters are shown in Figure \ref{fig:paramAnal}. Figure \ref{fig:paramAnal}(a) shows the effect of fusing parameter, $\beta,$ Equ. (\ref {eqn:fus}) on the mAP. Thereby, we can observe that the mAP tends to increase with a larger $\beta$ value. This shows that the result gets better when we deviate much from the CDS branch. Figure \ref{fig:paramAnal}(b) shows the impact of the number of images per person-ID ($\Omega$) in a given batch. We have experimented setting $\Omega$ to 4, 8, and 16, as can be seen, we obtain  a marginal improvement when we set $\Omega$ to 16. However, considering the direct relationship between the running time and $\Omega$, the improvement is negligible. c) and d) show probe-gallery similarity obtained from baseline and DCDS method, using three different probe-images, with a batch size of 64, and setting $\Omega$ to 4, 8 and 16. 

In the supplementary material, we provide additional experiments on cross-dataset person-re-identification (re-id) using the proposed deep constrained dominant sets (DCDS) on Market1501 dataset. In section one, we summarize the  datasets we used in our experiments. In section two, we present the experiments we have performed on cross-dataset person re-id. And, in section three, we provide hyper  parameter analysis on DukeMTMC-reID and CUHK03 datasets. Figure \ref{fig:metaphore} illustrates an example of our method training-output (left) and learning objective, target matrix, (right). Figure \ref{fig:exemplarResult} demonstrates the similarity fusing process, between the V-Net and CDS-Net, alongside sample qualitative results.



\textbf{Experiments on cross-datasets evaluation.} Due to the lack of abundant labeled data, cross-dataset person re-id has attracted great interest. Recently, Fan \et\cite{FanZYY18} have developed a progressive clustering-based method to attack cross-dataset person re-id problem. To further validate our proposed DCDS, we apply our method on cross-dataset person re-id problem and compare it with progressive unsupervised learning (PUL)  \cite{FanZYY18}. To this end, we train our model on DukeMTMC-reID and CUHK03 datasets and test it on Market1501 dataset.  We then compare it with PUL \cite{FanZYY18}, which has also been trained on CUHK03 and DukeMTMC-reID datasets. As can be observed from Table \ref{table:compCUHK}, even though our proposed method is not intended for cross-dataset re-id, it has gained a substantial improvements over PUL \cite{FanZYY18}, that was  mainly designed to attack person re-id problem in a cross-dataset setup.

\subsection{Parameter Analysis}
Similar to  the parameter analysis reported in the main manuscript, we report hyper parameter analysis on DukeMTMC-reID and CUHK03 dataset. The performance of our method with respect to the fusing parameters on DukeMTMC-reID and CUHK03 are shown in Figure \ref{fig:fusingpar} (a) and Figure \ref{fig:fusingpar} (b), respectively. Thereby, as can be observed, the results  show similar phenomena as in Market1501, where the mAP increases with a larger $\beta$ value. Figure \ref{fig:SimilDist} shows the similarity distribution given by the baseline and the proposed DCDS using three different probe-images, with a batch size of 64, and setting $\Omega$ to 4, 8 and 16.


\section{Summary}

In this work, we  presented a novel insight to enhance the learning capability of a DNN through the exploitation of a constrained clustering mechanism. To validate our method, we have conducted extensive experiments on several benchmark datasets. Thereby, the proposed method not only improves state-of-the-art person re-id methods but also demonstrates the benefit of incorporating a constrained-clustering mechanism in the end-to-end learning process. Furthermore, the presented work could naturally be extended to other applications which leverage a similarity-based learning. 
As a future work, we would like to investigate dominant sets clustering as a loss function.  
\newpage
\chapter{Conclusion}
\label{chap:intro}

In this thesis, we have proposed several schemes which exploit constrained clustering mechanism to tackle different computer vision problems such as, Image Segmentation and Co-segmentation, Image Retrieval, and Person Re-identification. Thereby, we validate the indispensability of the proposed graph-based algorithms. Moreover, the usage of constrained dominant sets (CDS) in an end-to-end manner demonstrates the advantage of integrating graph-based classical methods into a deep neural network (DNN) model.

In Chapter 2, we have demonstrated the applicability of CDS to problems such as interactive image segmentation and co-segmentation (in both the unsupervised and the interactive flavor). In our perspective, these can be thought of as ``constrained'' segmentation problems involving an external source of information (being it, for example, a user annotation or a collection of related images to segment jointly) which somehow drives the whole segmentation process. The approach is based on some properties of a family of quadratic optimization problems related to dominant sets which show that, by properly selecting a regularization parameter that controls the structure of the underlying function, we are able to ``force'' all solutions to contain the constraint elements. 
The proposed method is flexible and is capable of dealing with various forms of constraints and input modalities, such as 
scribbles and bounding boxes, in the case of interactive segmentation. 
Extensive experiments on benchmark datasets have shown that our approach considerably improves the state-of-the-art results on the problems addressed.

In Chapter 3, we addressed the content-based image retrieval problem. We developed a novel and computationally efficient CBIR method based on a constrained-clustering concept. In particular, we showed an efficient way of estimating a positive impact weight of features in a query-specific manner. Thus it can be readily used for feature combination. Furthermore, the proposed scheme is fully unsupervised, and can easily be able to detect false-positive NNs to the query, through the diffused similarity of the NNs. To demonstrate the validity of our method, we performed extensive experiments on benchmark datasets. Besides the improvements achieved on the state-of-the-art results, our method shows its effectiveness in quantifying the discriminative power of given features. Moreover, its effectiveness on feature-weighting can also be exploited in other computer vision problems, such as person re-identification, object detection, and image segmentation.

On the other hand, in CDSIR, we have developed a locally constrained diffusion
process which, as of existing methods, has no problems such as choosing optimal local neighbor size and initializing the dynamics to extract dense neighbor which constrain the diffusion process. The framework alleviates the issues while
improving the performance. Experimental results on three well known datasets in the field of retrieval demonstrate that the approach compares favorably with state-of-the-art algorithms. 

In Chapter 4, we presented a novel insight to enhance the learning capability of a DNN through the exploitation of a constrained clustering mechanism. To validate our method, we have conducted extensive experiments on several benchmark datasets. Thereby, the proposed method not only improves state-of-the-art person re-id methods but also demonstrates the benefit of incorporating a constrained-clustering mechanism in the end-to-end learning process. Furthermore, the presented work could naturally be extended to other similarity based applications. 

\cleardoublepage
\phantomsection
\addcontentsline{toc}{chapter}{\bibname}
\small
\bibliographystyle{plain}
\bibliography{thesis}

\end{document}